\documentclass[letterpaper,11pt]{article} 

\usepackage{amsmath}
\usepackage{amssymb}
\usepackage[comma,longnamesfirst]{natbib}
\usepackage[dvips]{epsfig}
\usepackage{dcolumn}
\usepackage{enumerate}
\usepackage{hhline}
\usepackage{dsfont}
\usepackage{afterpage}
\usepackage{arydshln}
\usepackage{graphicx}
\usepackage{color}
\usepackage[usenames,dvipsnames]{xcolor}
\usepackage{rotating}
\usepackage[breaklinks]{hyperref}
\usepackage{breakurl} 
\usepackage{xr}
\usepackage[percent]{overpic}
\usepackage[]{caption}
\usepackage{algorithmicx}
\usepackage[]{algpseudocode}
\usepackage{algorithm}
\usepackage{diagbox}
\usepackage{graphicx}
\usepackage{wrapfig}
\usepackage{lscape}
\usepackage{fontenc}
\usepackage{setspace}
\usepackage{bm}
\usepackage{slashbox}
\usepackage{lscape}
\usepackage{breakurl} 
\usepackage{multirow,booktabs}
\usepackage{eurosym}
\usepackage{titlesec}
\usepackage{sectsty}
\usepackage{placeins}
\usepackage{subcaption}
\usepackage[flushleft]{threeparttable}
\usepackage{makecell}
\usepackage[normalem]{ulem}
\epsfverbosetrue
\def\theequation{\thesection.\arabic{equation}}  
\def\abstract{\if@twocolumn
\section*{Abstract}
\else \normalsize 
\begin{center}
{\bf Summary\vspace{-.5em}\vspace{0pt}} 
\end{center}
\quotation 
\fi}
\def\endabstract{\if@twocolumn\else\endquotation\fi}

\makeatletter
\newcommand{\myappendix}[1]{
	\setcounter{section}{1}
        \renewcommand{\thesection}{A\arabic{section}}}

\DeclareMathOperator{\diag}{diag}

\def \calL {\mathcal L}

\def \bvec {\text{\boldmath$b$}}    
\def \cvec {\text{\boldmath$c$}}    
\def \dvec {\text{\boldmath$d$}}    
\def \evec {\text{\boldmath$e$}}

\def \hvec {\text{\boldmath$h$}}

\def \lvec {\text{\boldmath$l$}}    
\def \mvec {\text{\boldmath$m$}}

\def \pvec {\text{\boldmath$p$}}

\def \uvec {\text{\boldmath$u$}}    
    
\def \wvec {\text{\boldmath$w$}}    \def \mW {\text{\boldmath$W$}}
\def \xvec {\text{\boldmath$x$}}    \def \mX {\text{\boldmath$X$}}
\def \yvec {\text{\boldmath$y$}}    
\def \zvec {\text{\boldmath$z$}}    \def \mZ {\text{\boldmath$Z$}}

\def \alphavec        {\text{\boldmath$\alpha$}}
\def \betavec         {\text{\boldmath$\beta$}}

\def \varepsilonvec   {\text{\boldmath$\varepsilon$}}

\def \etavec          {\text{\boldmath$\eta$}}
\def \thetavec        {\text{\boldmath$\theta$}}

\def \iotavec         {\text{\boldmath$\iota$}}

\def \lambdavec       {\text{\boldmath$\lambda$}}
\def \muvec           {\text{\boldmath$\mu$}}

\def \psivec          {\text{\boldmath$\psi$}}

\def \omegavec        {\text{\boldmath$\omega$}}

\usepackage{color}
\usepackage{colordvi}
\newlength{\breite}
\breite\textwidth
\newcounter{aufg}[section]
  {\refstepcounter{aufg}\noindent\textbf{Exercise \arabic{aufg}:}
   \\*[1ex]\noindent}{\vspace{.5cm}}
   
 \newcounter{notes}[section]
  {\refstepcounter{aufg}\noindent\textbf{}
   \\*[1ex]\noindent}{\vspace{.5cm}}
   
\usepackage{amsthm}  

\theoremstyle{definition}

\newtheorem*{beisp*}{Example}
\newtheorem{Proof}{Proof}


\newtheoremstyle{break}
  {}
  {}
  {}
  {}
  {\bfseries}
  {.}
  {\newline}
  {}
  
\theoremstyle{break}

\newcommand{\head}[2]%
 {\hrule \vspace{.15cm} {\sfbold Advanced Statistical Inference, Summer Term 2012, Georg-August-University G\"ottingen}\hfill
{\sfbold Sheet #1}\\
{\sfbold Prof. Dr. Thomas Kneib, Nadja Klein}\hfill {\sfbold #2}

\vspace{.2cm}
\hrule

\vspace{1cm}

}

\newcounter{auf}
{\refstepcounter{auf}
\begin{center}
\fcolorbox[gray]{0}{.95}{
\makebox[\breite]{
\textbf{Exercise \arabic{auf}}
}}\\*[1ex]\noindent
\end{center}
}{\vspace{.5cm}}

\newcounter{loes}[section]
{\stepcounter{loes}
\begin{center}
\fcolorbox[gray]{0}{.95}{
\makebox[\breite]{
\textbf{L"osung \arabic{loes}}
}}\\*[1ex]\noindent
\end{center}
}{}

{\begin{center}
\fcolorbox[gray]{0}{.95}{
\makebox[\breite]{
\textbf{Zu Aufgabe #1}
}}\\*[1ex]\noindent
\end{center}\vspace{1cm}
}{\vspace{1cm}}

\newcounter{ka}
{\refstepcounter{ka}
\begin{center}
\framebox[\textwidth]{
\textbf{Aufgabe \arabic{ka}} \hfill #1 Punkte
}\\*[1ex]\noindent
\end{center}
}{\vspace{1cm}}

\newcounter{lka}
{\refstepcounter{lka}
\begin{center}
\framebox[\textwidth]{
\textbf{L\"osung \arabic{lka}} \hfill #1 Punkte
}\\*[1ex]\noindent
\end{center}
}{\vspace{1cm}}

\usepackage[margin=1in]{geometry}
\titlespacing*\section{0pt}{0pt plus 4pt minus 2pt}{0pt plus 2pt minus 2pt}
\titlespacing*\subsection{0pt}{0pt plus 4pt minus 2pt}{0pt plus 2pt minus 2pt}
\titlespacing*\subsubsection{0pt}{0pt plus 4pt minus 2pt}{0pt plus 2pt minus 2pt}

\definecolor{myblue}{RGB}{0,73,114}
\allsectionsfont{\sffamily\color{myblue}}

\newcounter{myremark}

\newcounter{mynotation}

\usepackage{paralist}

\renewenvironment{itemize}[1]{\begin{compactitem}#1}{\end{compactitem}}
\renewenvironment{enumerate}[1]{\begin{compactenum}#1}{\end{compactenum}}

\newcommand{\mycomment}[1]{}

\newtheorem{theorem}{Theorem}
\newtheorem{corollary}{Corollary}[theorem]

\makeatletter
\def\@seccntformat#1{\@ifundefined{#1@cntformat}%
	{\csname the#1\endcsname\quad}  
	{\csname #1@cntformat\endcsname}
}
\let\oldappendix\appendix 
\renewcommand\appendix{%
	\oldappendix
	\newcommand{\section@cntformat}{\appendixname~\thesection\quad}
}
\makeatother

\usepackage{titlesec}

\usepackage{scalerel,stackengine}
\stackMath
\newcommand\reallywidehat[1]{%
\savestack{\tmpbox}{\stretchto{%
  \scaleto{%
    \scalerel*[\widthof{\ensuremath{#1}}]{\kern-.6pt\bigwedge\kern-.6pt}%
    {\rule[-\textheight/2]{1ex}{\textheight}}
  }{\textheight}%
}{0.5ex}}%
\stackon[1pt]{#1}{\tmpbox}%
}

\begin{document}
\setlength{\abovedisplayskip}{0.15cm}
\setlength{\belowdisplayskip}{0.15cm}
\pagestyle{empty}
\begin{titlepage}

\title{\bfseries\sffamily\color{myblue}  
	 Natural Gradient Hybrid Variational Inference with Application to Deep Mixed Models}
\author{Weiben Zhang, Michael Smith, Worapree Maneesoonthorn\\ \& Rub\'en Loaiza-Maya}
\date{\today}
\maketitle
\noindent
{\small Weiben Zhang is a PhD student and Michael Smith is Professor of Management (Econometrics) at the Melbourne Business School, University of Melbourne, Australia. Worapree Maneesoonthorn is Associate Professor and Rub\'en Loaiza-Maya is Senior Lecturer at the Department of 
	Econometrics and Business Statistics, Monash University, Australia. Worapree Maneesoonthorn gratefully acknowledges support by the Australian Research Council through grant DP200101414. Rub\'en Loaiza-Maya gratefully acknowledges support by the Australian Research Council through grant DE230100029. Weiben Zhang gratefully acknowledges support from the University of Melbourne through the Faculty of Business and Economics Graduate Research Scholarship. The authors would like to thank two referees who provided comments that improved the paper. Correspondence should be directed to Michael Smith at {\tt mikes70au@gmail.com}.
}

\newpage
\begin{center}
\mbox{}\vspace{2cm}\\
{\LARGE \title{\bfseries\sffamily\color{myblue} Natural Gradient Hybrid Variational Inference with Application to Deep Mixed Models}
}\\
\vspace{1cm}
{\Large Abstract}
\end{center}
\vspace{-1pt}
\onehalfspacing
\noindent
Stochastic models with global parameters and latent variables are common, and for which variational inference (VI) is popular. However, existing methods are often either slow or inaccurate in high dimensions. We suggest a fast and accurate VI method for this case that employs a well-defined natural gradient variational optimization that targets the joint posterior of the global parameters and latent variables. It is a hybrid method, where at each step the global parameters are updated using the natural gradient and the latent variables are generated from their conditional posterior. A fast to compute expression for the Tikhonov damped Fisher information matrix is used, along with the re-parameterization trick, to provide a stable natural gradient. We apply the approach to deep mixed models, which are an emerging class of Bayesian neural networks with random output layer coefficients to allow for heterogeneity. A range of simulations show that using the natural gradient is substantially more efficient than using the ordinary gradient, and that the approach is faster and more accurate than two cutting-edge natural gradient VI methods. In a financial application we show that accounting for industry level heterogeneity using the deep mixed model improves the accuracy of asset pricing models. {\em MATLAB code to implement the method can be found at:} {\tt https://github.com/WeibenZhang07/NG-HVI}.
\vspace{20pt}
 
\noindent
{\bf Keywords}:  Asset Pricing; Bayesian Neural Networks; Natural Gradient Optimization;  Random Coefficients; Re-parameterization trick; Variational Bayes.

\end{titlepage}

\newpage
\pagestyle{plain}
\setcounter{equation}{0}
\renewcommand{\theequation}{\arabic{equation}}

\section{Introduction}\label{sec:intro}
Black box variational inference (VI) methods~\citep{ranganath2014black} 
that employ a generic
approximating family for the Bayesian posterior distribution are popular.
The approximation is usually learned
by minimizing the
Kullback-Leibler divergence between the two, with 
stochastic gradient ascent \footnote{While it is more common to refer to 
	optimization algorithms as ``descent'', throughout this paper 
	we use the term ``ascent'' instead because our variational optimization is written as a maximization problem.}
	 (SGA) the most common
choice of algorithm to solve the optimization
problem~\citep{bottou10,hoffmanStochasticVariationalInference2013}. Here, the variational approximation (VA)
is updated in the direction of the (noisy) ordinary gradient of the objective function. 
However, the natural gradient gives the steepest direction for this  optimization problem~\citep{amariNaturalGradientWorks1998},
and stochastic natural gradient ascent (SNGA) is an alternative algorithm that
can converge in fewer
steps and avoid plateaus in the objective function~\citep{rattray1998,martensNewInsightsPerspectives2020}. But for large scale problems computing the natural gradient is usually 
impractical for approximations outside the exponential family. 
In this paper, we show that the natural gradient can be computed quickly
for the VA family proposed by~\cite{loaiza-mayaFastAccurateVariational2022},
thereby extending the class of VAs to which SNGA can be applied.
Doing so can greatly reduce the time required to learn this approximation
 compared to SGA.

\cite{loaiza-mayaFastAccurateVariational2022} suggested a VI method
where the target vector $\psivec$ is partitioned
into two components $\psivec=(\thetavec^\top,\zvec^\top)^\top$.
The method samples $\zvec$ from its conditional posterior, while using (noisy)
ordinary gradients to update the marginal VA for
$\thetavec$.
The authors show these two steps together form a
well-defined SGA algorithm
for solving the variational 
optimization for the joint posterior of $\psivec$, and 
call their method ``hybrid VI'' because it combines a generation and a gradient update step. 
The approach is particularly
useful when $\thetavec$ are the global parameters and $\zvec$ is a high-dimensional vector of latent variables, which are also sometimes called local
parameters~\citep{hoffmanStochasticVariationalInference2013}.
However, the method can
suffer from slow convergence in common with other first
order stochastic optimization methods for a number of problems, 
such as training some neural networks~\citep{zhangNoisyNaturalGradient2018}. 
In this paper we improve the efficiency of 
hybrid VI by using SNGA optimization. 
The objective is to reduce the number of draws of $\zvec$---typically the slowest step of the algorithm---by 
providing faster convergence of the 
variational optimization. We show that the combination of natural gradient based optimization and hybrid VI provides a 
particularly attractive VI method for complex high-dimensional target posteriors. 

The natural gradient is equal to the ordinary gradient pre-multiplied by the inverse of the Fisher information matrix (FIM)~\citep{amariNaturalGradientWorks1998}, and the computation, storage and factorization of the latter can be costly.\footnote{While the inverse is not usually computed and stored directly, systems of equations are solved that still require derivation of a factorization of the inverse FIM.} We show that
in hybrid VI the FIM of the VA of $\psivec$
is equal to that of the FIM of the marginal VA of $\thetavec$. 
When 
the dimension of $\thetavec$ is low relative to that of $\zvec$---as is often
the case when $\zvec$ is a vector of latent variables---then 
the additional cost of computing the natural gradient of the VA for $\psivec$ is minor compared to the ordinary gradient. This additional cost is more than 
off-set by the savings gained from the smaller number of draws of $\zvec$ required, thus improving the overall efficiency
of the hybrid VI algorithm. 
For the marginal VA of $\thetavec$ we
use a Gaussian distribution with a factor covariance matrix~\citep{miller2017,mishkin2018slang,ongGaussianVariationalApproximation2018,tranBayesianDeepNet2020}. Fast to compute analytical re-parameterized natural gradient updates that use
the Tikhonov damped Fisher information matrix are derived. 
Tikhonov damping,
combined with adaptive learning rates, is a way to address the
numerical difficulties often encountered in practice with natural gradient
updates when training complex models~\citep{osawa2019large}.

There is growing interest in using NGA in VI. \cite{martensNewInsightsPerspectives2020} gives an overview
and discussion of NGA as
a second order optimization method, while
 \cite{martensOptimizingNeuralNetworks2015,khanFastSimpleNaturalGradient2018}
 and \cite{zhangFastConvergenceNatural2019} show that  natural gradient based VI methods can improve the efficiency of training neural networks,
 which we also find here.
\cite{khanConjugateComputationVariationalInference2017,khanFastSimpleNaturalGradient2018}
consider the case
where the VA is from the exponential family, 
and
\cite{linFastSimpleNaturalGradient2019} where the VA 
is a mixture of exponential distributions.
\cite{martensOptimizingNeuralNetworks2015} and~\cite{tranBayesianDeepNet2020} approximate the FIM by block diagonal matrices to reduce computation cost. \cite{tanAnalyticNaturalGradient2022} considers efficient application of SNGA to learn high-dimensional Gaussian VAs based on Cholesky factorization, whereas~\cite{linTractableStructuredNaturalgradient2021} 
suggests transforming the variational parameters to simplify 
 computation of the natural gradient for some fixed form VAs.

We illustrate  our VI approach by using it to estimate several stochastic models. 
Our main focus is on deep mixed models (DMM), which are a class of probabilistic 
Bayesian neural networks. Mixed models (also called random coefficient or multi-level models) are widely used to capture heterogeneity in statistical modeling~\citep{mcculloch2004}, and DMMs extend this approach to deep learning~~\citep{,wikle2019,tranBayesianDeepNet2020,simchoni2021}.
Following~\cite{tranBayesianDeepNet2020}, in our DMM  
the output layer coefficients vary by a group variable and follow a 
 Gaussian distribution, allowing  for heterogeneity. 
 To apply hybrid VI, the vector $\zvec$ contains the output layer group level coefficient values. 
 Using simulation studies, we establish that SNGA is much faster and 
 more reliable than SGA for learning the VA in hybrid VI. We also show 
 for our examples that the natural gradient 
 hybrid VI is faster and more accurate than the alternative 
 natural gradient based VI methods
 of~\cite{tranBayesianDeepNet2020} and~\cite{tanAnalyticNaturalGradient2022}.
 

Recent studies suggests that deep models have strong potential in 
financial 
modeling~\citep{guEmpiricalAssetPricing2020,guAutoencoderAssetPricing2021,fangMachineLearningBased2021}.  We use our approach to estimate three factor~\citep{famaCommonRiskFactors1993} and five factor~\citep{famaFivefactorAssetPricing2015} 
financial asset pricing models. We use monthly returns on 2583 stocks between
January 2005 and December 2014 to train a DMM with feed forward neural network architecture that accounts for heterogeneity 
in 548 industry groups. 
Accuracy of the learned model is assessed using posterior predictive distributions computed
for both the training data, and also for a validation period from January 2015 to December 2019. The results suggest that the DMM improves probabilistic 
predictive accuracy
compared to existing (non-deep) linear mixed modeling, and (non-mixed) feed forward neural networks.

The rest of the paper is organized as follows. Section~\ref{sec:hvi} provides
a brief introduction to the hybrid VI approach, and  Section~\ref{sec:nghvi} extends
hybrid VI to employ the natural gradient. 
Section~\ref{sec:hvidmm} outlines DMMs and in simulation 
studies shows how 
natural gradient hybrid VI leads to an increase in computational efficiency and accuracy for training a DMM,  
compared to both ordinary gradient hybrid VI and the benchmark natural gradient methods
of~\cite{tranBayesianDeepNet2020} and~\cite{tanAnalyticNaturalGradient2022}. 
Section~\ref{sec:finance} contains the financial asset pricing study, while Section~\ref{sec:disc} discusses
the contribution of our study and directions for future work.
\section{Hybrid Variational Inference}\label{sec:hvi}
\subsection{Variational inference}
We consider a stochastic model with data $\yvec$ and unknowns $\psivec$. 
Bayesian inference for $\psivec$ employs its posterior density $p(\psivec|\yvec) \propto p(\yvec | \psivec) p(\psivec)\equiv g(\psivec)$, where $p(\yvec|\psivec)$ is the likelihood and $p(\psivec)$ is the prior. 
The posterior is often difficult to evaluate, and in
VI it is approximated using a density $q(\psivec)\in {\cal Q}$, with ${\cal Q}$  a
family of flexible but tractable densities. The density $q$ is called the ``variational approximation'' (VA) and obtained by minimizing a distance metric between $p(\psivec|\yvec)$ and $q(\psivec)$.
The Kullback-Leibler divergence (KLD) is the most popular choice, and it is easily shown that minimizing
the KLD corresponds to maximizing the Evidence Lower Bound (ELBO)  
\begin{align*}
	\calL = E_{q}\left[\log g(\psivec) - \log q(\psivec)\right] 
\end{align*}
over $q\in {\cal Q}$, where the expectation above is with respect to $\psivec\sim q$. This optimization 
problem is called  ``variational optimization'' and the method used for its solution determines
the speed of the VI method; for overviews of VI see \cite{ormerodExplainingVariationalApproximations2010},
\cite{bleiVariationalInferenceReview2017}, and~\cite{zhangAdvancesVariationalInference2019}.

\subsection{Variational inference for latent variable models}
Stochastic models that have both unknown
parameters $\thetavec$ and latent 
variables $\zvec$ are  popular.
Examples include mixed models 
where $\zvec$ are random coefficients \citep{tranBayesianDeepNet2020}, tobit models where $\zvec$ are uncensored data~\citep{danaherAdvertisingEffectivenessMultiple2020},
state space models where $\zvec$ are latent states~\citep{wangStructuredVariationalInference2022},
and topic models where $\zvec$ are topics~\citep{blei2003}. There are a number
of ways to apply 
VI to this case. The most popular strategy is to set $\psivec=(\thetavec^\top,\zvec^\top)^\top$ and approximate
the ``augmented posterior'' $p(\psivec|\yvec)$; for recent examples, see~\cite{hoffmanStructuredStochasticVariational2015,loaiza-mayaVariationalBayesEstimation2019}
and~\cite{tanUseModelReparametrization2021}.
However, because $\mbox{dim}(\zvec)$ is often very large, approximating the augmented posterior can introduce
substantial cumulative error. One alternative is to integrate out $\zvec$ using
numerical or Monte Carlo methods, such as importance sampling~\citep{gunawanFastInferenceIntractable2017,tranBayesianDeepNet2020}. However, this can prove slow or even computationally infeasible for large models. An alternative that is both more
accurate and scalable was suggested by~\cite{loaiza-mayaFastAccurateVariational2022}, which we
briefly outline below.

\subsection{Hybrid variational inference}
For latent variable models consider the VA
\begin{align}
	q_{\lambda}(\psivec) = p(\zvec|\thetavec,\yvec) q^0_{\lambda}(\thetavec) \,,
	\label{eq:hybrid_q}
\end{align}
where $q^0_{\lambda}(\thetavec)$ is the density of a fixed form VA with 
parameters $\lambdavec$. 
Then it is possible to
show that the ELBO for the augmented posterior of $\psivec=(\thetavec^\top,\zvec^\top)^\top$ is equal to the ELBO of the marginal
posterior of $\thetavec$; that is,
\begin{equation}
{\cal L}(\lambdavec)\equiv E_{q_\lambda}\left[\log g(\psivec) - \log q_\lambda(\psivec)\right]=
 E_{q^0_\lambda}\left[\log \left(p(\yvec|\thetavec)p(\thetavec)\right) - \log q^0_\lambda(\thetavec)\right]\equiv
{\cal L}^0(\lambdavec)\,.\label{eq:elboequal}
\end{equation}
Here, both ELBO's are denoted as functions of $\lambdavec$ because this vector parameterizes both $q^0_\lambda$ and 
$q_\lambda\in {\cal Q}$. The equality at~\eqref{eq:elboequal} means that solving
the variational optimization for $\psivec$ with respect to $\lambdavec$ also solves the variational optimization problem
 for $\thetavec$ with $\zvec$ integrated out {\em exactly}. We stress this point because it is the source
 of the increased accuracy of this VI method.

\cite{loaiza-mayaFastAccurateVariational2022} use a stochastic gradient ascent (SGA) algorithm~\citep{bottou10}
to solve the variational
optimization. The key input to SGA is an unbiased estimate of the 
ordinary gradient $\nabla_\lambda {\cal L}(\lambdavec)$	, and these 
authors show
how the re-parameterization trick~\citep{kingmaAutoEncodingVariationalBayes2014,RezMohWie2014} 
combined with the VA at~\eqref{eq:hybrid_q} gives an efficient
gradient estimate. The re-parameterization is of the model parameters
$\thetavec=h(\varepsilonvec^0,\lambdavec)\sim q^0_\lambda$
in terms of a random vector $\varepsilonvec^0\sim f_{\varepsilonvec^0}$
that is invariant to 
$\lambdavec$, and
a deterministic function $h$. The latent variables $\zvec$ are not 
re-parameterized. 
If the joint density of $\varepsilonvec=\left((\varepsilonvec^0)^\top,\zvec^\top\right)^\top$ is denoted as
\[
f_\varepsilonvec(\varepsilonvec)=f_\varepsilonvec(\varepsilonvec^0,\zvec)=f_{\varepsilonvec^0}(\varepsilonvec^0)p(\zvec|h(\varepsilonvec^0,\lambdavec),\yvec)\,, \]
then it is possible to show that
the gradient can be 
written as an expectation with respect to $\varepsilonvec\sim f_\varepsilon$ as
\begin{equation}
	\nabla_\lambda\mathcal{L}(\bm{\lambda}) = E_{f_\varepsilon}\left[
	\frac{\partial h(\varepsilonvec,\lambdavec)^\top}{\partial \lambdavec}
	\left(
	\nabla_\theta \log g(\thetavec,\zvec) - \nabla_\theta \log q_\lambda^0(\thetavec) 
	\right)
	\right]\,.
	\label{eq:repargrad}
\end{equation}	
A key observation is that to evaluate~\eqref{eq:repargrad} only $ \nabla_\theta \log q_\lambda^0(\thetavec) $ is required, rather 
the higher dimensional $\nabla_\psivec \log q_\lambda(\psivec)$ which would be necessary for other choices of the VA at~\eqref{eq:hybrid_q}. 
An unbiased approximation to the expectation above is evaluated by plugging in a single draw from $f_\varepsilon$ obtained by first drawing from $f_{\varepsilon^0}$, and then 
from $p(\zvec|\thetavec,\yvec)$ with $\thetavec=h(\varepsilonvec^0,\lambdavec)$. 
For many stochastic models, generation from
$p(\zvec|\thetavec,\yvec)$ can be done either exactly or approximately using MCMC or other methods.
This provides a framework where MCMC can be used within a well-defined SGA variational optimization, so that these authors
refer to the method as hybrid VI. In the next section we extend the approach
of these authors
to use the natural gradient to both increase the effectiveness and further reduce the computational burden of hybrid VI.
\section{Natural Gradient Hybrid Variational Inference}\label{sec:nghvi}
\subsection{Natural gradient ascent}
Given a starting value $\lambdavec^{(0)}$, SGA recursively updates the 
variational parameters as
\begin{equation}\label{Eq:Update}
	\bm{\lambda}^{(t+1)} = \bm{\lambda}^{(t)}+\bm{\rho}^{(t)}\circ\left.\widehat{\nabla_\lambda\mathcal{L}(\bm{\lambda})}\right|_{\lambdavec=\lambdavec^{(t)}},\; t=0,1,\ldots
\end{equation}
until reaching convergence. Here, $\bm{\rho}^{(t)}$ are adaptive learning rates, `$\circ$' is the Hadamard (element-wise) product,
and $\widehat{\nabla_\lambda\mathcal{L}(\bm{\lambda})}$ is an unbiased 
estimator of the gradient that is evaluated at $\lambdavec=\lambdavec^{(t)}$. While convergence of SGA follows from the conditions in~\cite{robbinsStochasticApproximationMethod1951}, the algorithm does not exploit the information 
geometry of $q_\lambda(\bm\psi)$ that can increase the speed of convergence in variational optimization~\citep{khanFastSimpleNaturalGradient2018}. In contrast, 
using 
the natural gradient does so~\citep{amariNaturalGradientWorks1998,honkela2010}.
The main idea is that $\widehat{\nabla_\lambda\mathcal{L}(\bm{\lambda})}$ is replaced in~\eqref{Eq:Update} by the unbiased estimate\footnote{The notation
	$\widetilde{\nabla}_\lambda\mathcal{L}(\bm{\lambda})$ is also often 
	used for the exact natural gradient, although here we do not employ
	additional notation to distinguish between it and its unbiased approximation.}
	 of the natural gradient 
\begin{equation}
	\widetilde{\nabla}_\lambda\mathcal{L}(\bm{\lambda}) \equiv F(\bm\lambda)^{-1}\widehat{\nabla_\lambda\mathcal{L}(\bm{\lambda})},
\end{equation}
where the Fisher information matrix (FIM)
\[
F(\bm\lambda) = E_{q_\lambda}\left[\nabla_\lambda\log q_\lambda(\bm\psi)\nabla_\lambda\log q_\lambda(\bm\psi)^\top\right]
=-E_{q_\lambda}\left[\frac{\partial^2 \log q_\lambda(\psivec)}{\partial \lambdavec \partial \lambdavec^\top}\right]\,,
\] 
points $\widehat{\nabla_\lambda\mathcal{L}(\bm{\lambda})}$ towards the steepest direction of the objective function. This approach, known as stochastic natural gradient ascent (SNGA), typically requires far fewer iterations to reach convergence than SGA; see~\cite{martensNewInsightsPerspectives2020} for a recent overview of natural gradient optimization.

The main requirement of NGA is that a tractable expression for $F(\bm\lambda)$ is available. Its
derivation for the VA at~\eqref{eq:hybrid_q} may appear challenging, but Theorem~\ref{thm:FIM} shows that $F(\lambdavec)$ for this
VA is tractable whenever the FIM for $q^0_\lambda$ is also.
\begin{theorem}[{\em Fisher information matrix for hybrid VI}]\label{thm:FIM}
	Let $q_{\lambda}(\psivec) = p(\zvec|\thetavec,\yvec) q^0_{\lambdavec}(\thetavec)$ and denote the Fisher information matrix of the marginal approximation $q^0_\lambda(\bm{\theta})$ as $F^0(\bm\lambda) = E_{q_\lambda^0}\left[\nabla_\lambda\log q_\lambda^0(\bm\theta)\nabla_\lambda\log q_\lambda^0(\bm\theta)^\top\right]$. Then it holds that 
\begin{equation*}
	F(\bm\lambda)  =  F^0(\bm\lambda)
\end{equation*}
\end{theorem}
\noindent {\em Proof}: See Appendix~\ref{App:NatGrad}

The corollary below follows immediately from Theorem~\ref{thm:FIM}. 
\begin{corollary}[{\em Natural gradient for hybrid VI}]\label{cor:ng}
If $\widehat{\nabla_\lambda\mathcal{L}(\bm{\lambda})}$ is an unbiased estimate
of the gradient of the evidence lower bound, then an unbiased estimate
of the natural gradient is given by 
	\begin{equation}\label{Eq:natgrad}
		\widetilde{\nabla}_\lambda\mathcal{L}(\bm{\lambda})  =   F^0(\bm\lambda)^{-1}\widehat{\nabla_\lambda\mathcal{L}(\bm{\lambda})}\,.
	\end{equation}
\end{corollary}

Theorem~\ref{thm:FIM} and Corollary~\ref{cor:ng} have three important implications 
for the use of SNGA to 
solve the variational optimization problem with
 the VA at~\eqref{eq:hybrid_q}. First, the natural gradient can be constructed for a wide range of latent variable models because it does 
not require evaluation of the density 
$p(\zvec|\thetavec,\yvec)$ or its derivative, but only the ability 
to generate from it. Second,
evaluation of the natural gradient is unaffected
by the dimension of $\zvec$.
In contrast,
other VI methods for latent variables can scale poorly; for example, the 
computational complexity of the FIM increases with the dimension of $\bm{z}$ for the VA in~\cite{tanAnalyticNaturalGradient2022}.
A third implication is that in Corollary~\ref{cor:ng} the
computation of the FIM and $\widehat{\nabla_\lambda\mathcal{L}(\bm{\lambda})}$ can 
be done separately.
This allows variance reduction techniques when estimating the latter, 
such as the 
use of control variates~\citep{paisley2012variational,ranganath2014black,trannottkohn2017} or 
the re-parametrization trick as adopted here. 

The natural gradient can provide an unreliable update in regions where the ELBO function is flat, as the FIM becomes singular in these regions. 
	As discussed in \cite{linTractableStructuredNaturalgradient2021}, this is in fact the case for Gaussian variational approximations with factor covariance structures, such as that of~\cite{ongGaussianVariationalApproximation2018} considered in this paper. Damping of the FIM has been proposed as an approach to tackle this type of problems inside variational inference \citep{zhangNoisyNaturalGradient2018}. 
	Here, we follow suit and employ the damping of the form 
\begin{equation}
	\widetilde{F}^0(\bm\lambda) = F^0(\bm\lambda)+\delta\text{diag}\left(F^0(\bm\lambda)\right)\,,
\end{equation}
where $\diag(F^0)$ is a diagonal matrix comprising the leading diagonal elements of $F^0$, and $\delta>0$ is a damping factor. \cite{martensTrainingDeepRecurrent2012} suggest that unlike damping that uses the identity matrix, this type of damping takes into consideration the scale in which the variational parameters lie and would preserve the self-rescaling property of the natural gradient update. We then construct an update by using the normalised damped natural gradient $\widetilde{\nabla}_{\lambda}\mathcal{L}(\lambdavec)/||\widetilde{\nabla}_{\lambda}\mathcal{L}(\lambdavec)||$ within the ADADELTA adaptive learning rate method. As shown in \cite{tanAnalyticNaturalGradient2022} the combination of a normalised natural gradient and an adaptive learning rate method leads to a stable updating algorithm. 
 Algorithm~\ref{alg:nghvi} outlines our proposed natural gradient hybrid VI approach, which we label ``NG-HVI''.   Following \cite{tanAnalyticNaturalGradient2022}, we use Step~(d)  in Algorithm~\ref{alg:nghvi} to add momentum in the evaluation of the natural gradient.


\begin{algorithm}[ht!]
	\begin{algorithmic}
		\State Initiate $\lambdavec^{(0)}$,  $\bar{\mvec}_0 = 0$ and set $t = 0$
		\Repeat
		\State (a) Generate $\thetavec^{(t)} \sim q_{\lambda^{(t)}}^0(\thetavec)$ using its re-parametrized representation
		\State (b)  Generate ${\zvec}^{(t)} \sim p(\zvec| {\thetavec}^{(t)},\yvec)$
		\State (c) Compute the (Tikhonov damped and stochastic) natural gradient \[\widetilde{\nabla}_{\lambda}\mathcal{L}(\lambdavec^{(t)}) = 	\widetilde{F}^0(\bm\lambda^{(t)})^{-1}\left.\widehat{\nabla_\lambda\mathcal{L}(\bm{\lambda})}\right|_{\lambdavec=\lambdavec^{(t)}}\]
		\State (d) $\bar{\mvec}_{t} = a_m\bar{\mvec}_{t-1} + (1-a_m)\widetilde{\nabla}_{\lambda}\mathcal{L}(\lambdavec^{(t)})/||\widetilde{\nabla}_{\lambda}\mathcal{L}(\lambdavec^{(t)})||$.
		\State (e) Compute step size $\boldsymbol{\rho}^{(t)}$ using an adaptive method (e.g. an ADA method)
		\State (f) Set $\lambdavec^{(t+1)} = \lambdavec^{(t)} + \boldsymbol{\rho}^{(t)} \circ \bar{\mvec}_{t}$
		\State (g) Set $t = t+1$
		\Until{either a stopping rule is satisfied or a fixed number of steps is taken}
	\end{algorithmic}
	\caption{Hybrid VI with Stochastic Natural Gradient Ascent (NG-HVI)}\label{fit_nga}
	\label{alg:nghvi}
\end{algorithm}
\subsection{Fixed form approximation}\label{sec:ffapprox}
For the marginal VA of $\thetavec$, we use Gaussian VAs  $q_\lambda^0(\thetavec)=\phi_m(\thetavec;\muvec,\Sigma)$ with a factor model
covariance matrix, also called ``low rank plus diagonal'', as
suggested by~\cite{miller2017,ongGaussianVariationalApproximation2018}
for the covariance matrix and~\cite{mishkin2018slang} for the precision matrix.
If $m=\mbox{dim}(\thetavec)$, then
$\Sigma=BB^\top+D^2$, with $D$ a diagonal matrix and
$B$ an $(m\times p)$ matrix where $p<<m$, so that 
the number of 
variational parameters scales linearly with $m$. 
We set the upper triangular elements of $B$ to zero, while leaving the leading diagonal elements unconstrained. Although the factor decomposition can be made unique by restricting the diagonal elements of $B$ and $D$, we do not do so because the unrestricted parameterization can expedite the optimization process as noted by~\cite{ongGaussianVariationalApproximation2018} and others when using ordinary gradients.
The variational parameters are $\lambdavec=(\muvec^\top,\mbox{vech}(B)^\top,\dvec^\top)^\top$, 
where ``vech'' is the half-vectorization operator applied to a rectangular matrix, and
$\dvec$ is a vector containing the non-zero entries of $D$. 

An advantage
of this factorization is that it has a convenient generative representation
given by
\[\thetavec=\muvec+B\varepsilonvec_1^0 + \dvec\circ\varepsilonvec_2^0\,,\]
that defines the transformation $h$ with $\varepsilonvec^0=((\varepsilonvec_1^0)^\top,(\varepsilonvec_2^0)^\top)^\top\sim N(\bm{0},I_{m+p})$. 
A second advantage is that the derivatives of $q^0_\lambda$ 
required to evaluate the re-parameterized gradient at~\eqref{eq:repargrad}
are given in closed form in~\cite{ongGaussianVariationalApproximation2018} 
and are fast to compute. A third advantage is that the damped natural
gradient $\widetilde{F}^0(\bm\lambda)^{-1}\widehat{\nabla_\lambda\mathcal{L}(\bm{\lambda})}$ can also be computed efficiently 
as follows. 

\cite{ongLikelihoodfreeInferenceHigh2018}
show that the FIM is sparse with
\begin{equation}
F^0(\lambdavec)=
\left[
\begin{array}{ccc}
F_{11}(\lambdavec) &\bm{0} &\bm{0} \\
\bm{0} &F_{22}(\lambdavec) &F_{32}(\lambdavec)^\top \\
\bm{0} &F_{32}(\lambdavec) &F_{33}(\lambdavec) \\
\end{array} 
\right]\,,\label{eq:FIMpattern}
\end{equation}
where the block matrices in $F^0$ follow the partition of $\lambdavec$, and  $\bm{0}$ denotes a conformable matrix of zeros. 
These authors derive closed form expressions for $F_{11},F_{22},F_{32}$
and $F_{33}$. The damped FIM is equal to~\eqref{eq:FIMpattern} 
but with the leading diagonal blocks replaced by 
$\widetilde{F}_{jj}(\lambdavec) = F_{jj}(\lambdavec) + \delta \mbox{diag}\left(F_{jj}(\lambdavec)\right)$ for $j=1,2,3$.
The first elements of the damped natural gradient can be computed as
$\widetilde{F}_{11}(\lambdavec)^{-1}\widehat{\nabla_\mu{\cal L}(\lambdavec)}$, using the closed form expression for $\widetilde{F}_{11}(\lambdavec)^{-1}$ given in Appendix~\ref{app:fim}. 
The remaining
elements 
\begin{equation}
\left[\begin{array}{cc}
	\widetilde{F}_{22}(\lambdavec) &F_{32}(\lambdavec)^\top \\
	F_{23}(\lambdavec) &\widetilde{F}_{33}(\lambdavec) 
\end{array}
\right]^{-1}
\left(\begin{array}{c}
	\widehat{\nabla_{\rm{vech}(B)}{\cal L}(\lambdavec)}\\
	\widehat{\nabla_{d}{\cal L}(\lambdavec)} 
\end{array}
\right)\,,\label{eq:soe}
\end{equation}
are obtained using a conjugate gradient solver.
Full details on the efficient calculation of the damped natural gradient
are given in Part~B of the Web Appendix.

To speed computation of the (un-damped) natural gradient, \cite{tranBayesianDeepNet2020} approximate the FIM as
	block diagonal, in which case there is no need to solve~\eqref{eq:soe} because $\widetilde F_{jj}(\lambdavec)^{-1}$ for $j=2,3$ can also be computed in closed form.
However, this loses information about the curvature of $q_\lambda$.  In addition, these authors set $p=1$ to speed computations,
although a higher number of factors is sometimes necessary to improve accuracy of the VA. In contrast, we do not adopt these simplifications and found that
employing the damped FIM stabilizes the natural gradient update step, allowing us to apply our algorithm
in reasonable time using natural gradients with dimension up to 13,895; see Table~\ref{tab:egsummary} for a summary of the size and characteristics of our examples.
Finally, we note that other VAs may also be considered for $q^0$, although restricting the choice to approximations where 
$F^0$ is tractable (as with the choice here) is necessary for a fast NG-HVI method in practice.

\subsection{Example~1: Linear regression with random effect}\label{sec:LMM}
To demonstrate the significant improvements NG-HVI can provide in even 
very simple models, 
we consider a linear regression with a random effect. This has five fixed effect covariates and intercept with coefficients $\betavec$, an additive random effect $\alpha_k\sim N(0,\sigma^2_\alpha)$ for groups $k=1,\ldots,K$, and errors $\epsilon_i\sim N(0,\sigma^2_\epsilon)$. We generate datasets of five
thousand observations from this data generating process (DGP) using different values for the number of groups $K$ and ratios of random effect to noise variance $\sigma^2_\alpha/\sigma^2_\epsilon$, as listed in Table~\ref{summary_elbo}.

To compute VI using NG-HVI we set $\thetavec=(\betavec^\top , \log(\sigma^2_\alpha), \log(\sigma^2_\epsilon))^\top$ and 
$\zvec=(\alpha_1, \dots, \alpha_K)^\top$, and observe
that it is straightforward to draw from the conditional
posterior $p(\zvec|\thetavec,\yvec)$ of this model; see
Part~C of the Web Appendix.
%
We use two benchmark methods. The first is hybrid VI using the same VA at~\eqref{eq:hybrid_q}, but
learned using SGA and labeled ``SG-HVI''. 
The second is
the data augmentation VI method suggested by \cite{tanAnalyticNaturalGradient2022}, and labeled ``DAVI''. This 
uses an $(K+8)$ dimensional Gaussian
approximation for $\psivec$,
assuming a sparse covariance matrix and utilizing the re-parameterization proposed by~\cite{tanUseModelReparametrization2021} along with SNGA to learn the approximation. 

\begin{figure}[thb!]
	\begin{subfigure}{0.5\linewidth}
		\subcaption{Trace of ELBO against step}
		\includegraphics[width=1\linewidth]{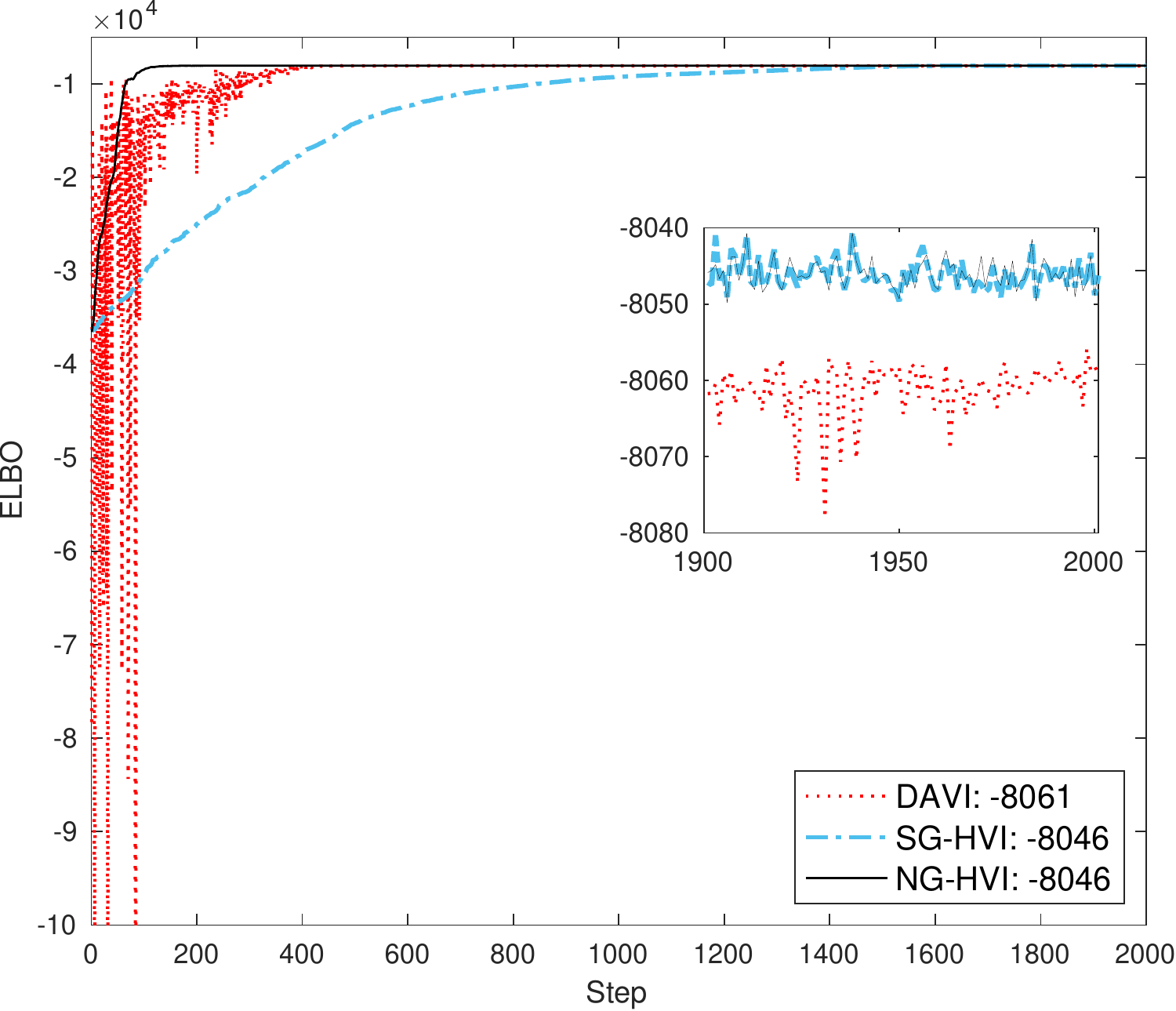}		
	\end{subfigure}
	\begin{subfigure}{0.5\linewidth}
		\subcaption{Trace of ELBO against clock-time}
		\includegraphics[width=1\linewidth]{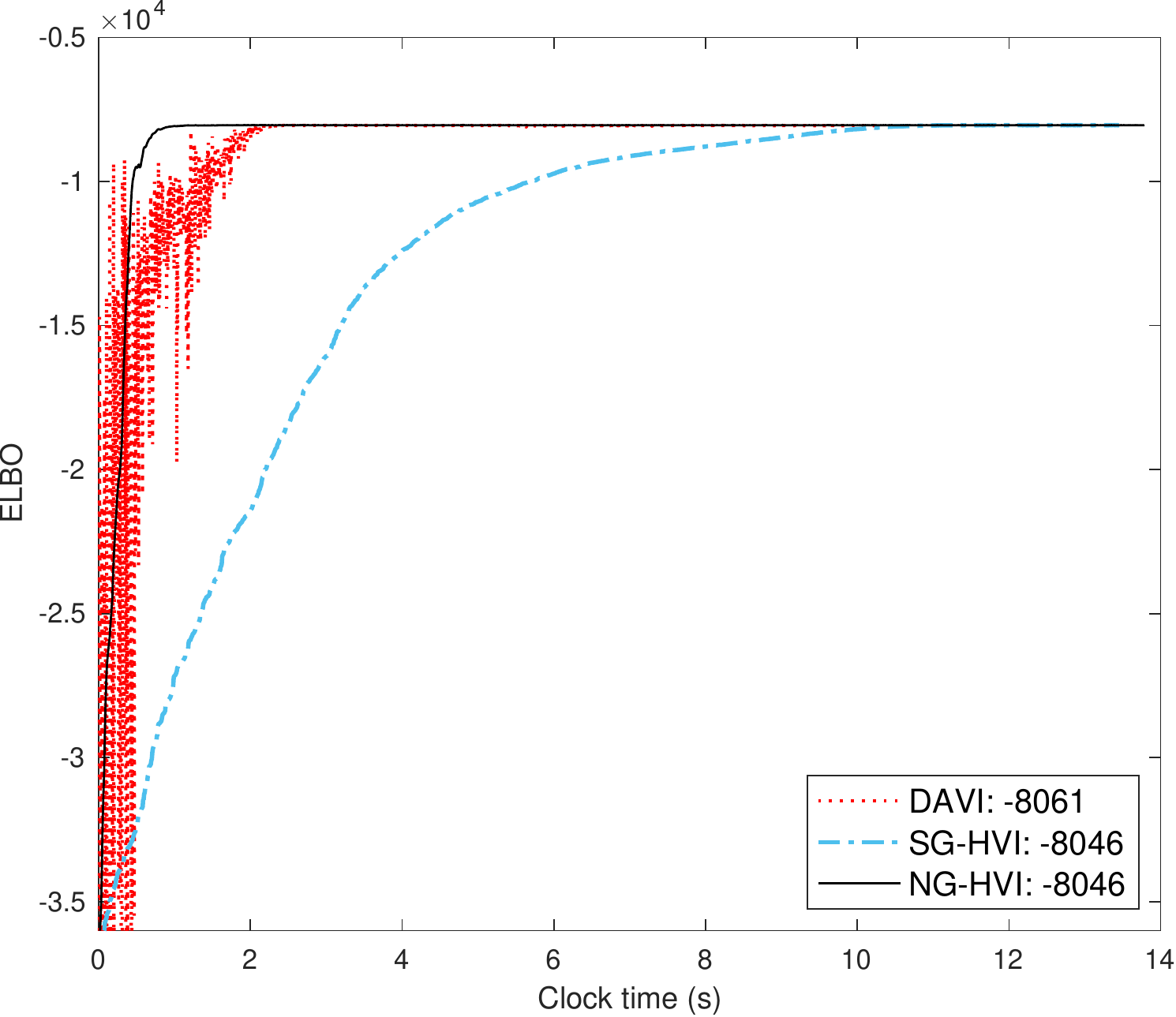}
		
	\end{subfigure}
	\caption{Plots of the noisy ELBO function for the linear random effects regression with $\sigma^2_\epsilon = \sigma^2_\alpha = 1$ and $K = 1000$ for Example~1. Panel~(a) plots against optimization step number, and panel~(b) plots against wall clock time (seconds). The results for DAVI are plotted as a dotted red line, SG-HVI as a dash-dot blue line, and NG-HVI as a solid black line. The average of the noisy ELBO function values over the last 100 steps are also reported.}
	\label{ELBO_LMM_tan}
\end{figure}

To compare methods we compute the noisy ELBO, which at step $t$ of the optimization is
\begin{equation}
	\widehat{\calL(\lambdavec^{(t)}) }= \log p(\yvec|\zvec^{(t)},\thetavec^{(t)}) + \log p(\zvec^{(t)}|\thetavec^{(t)})  + \log p(\thetavec^{(t)})- \log q_{\lambda^{(t)}}(\psivec^{(t)})\,,
	\label{eq:noisyL}
\end{equation}
where $\lambdavec^{(t)}$ are the variational parameters at step $t$, and $\psivec^{(t)}\sim q_{\lambda^{(t)}}$ is a random draw conditional on $\lambdavec^{(t)}$. For the
hybrid VI methods proposed in this paper, \eqref{eq:noisyL} further simplifies to 
	$\widehat{\calL(\lambdavec^{(t)}) }= \log p(\yvec|\zvec^{(t)},\thetavec^{(t)}) + \log p(\thetavec^{(t)})- \log q^0_{\lambda^{(t)}}(\thetavec^{(t)})$.
Figure~\ref{ELBO_LMM_tan} plots  $\widehat{\calL(\lambdavec^{(t)})}$ against (a)~step number, and (b)~clock time, for data simulated with $\sigma^2_\epsilon = \sigma^2_\alpha = 1$, $K = 1000$ groups and 5 data points in each group. 
We make four observations. 
First, the natural gradient methods (DAVI and NG-HVI) both converge at a much faster rate than SG-HVI.
Second, NG-HVI converges faster than DAVI and to a larger value. The latter
is because the VA at~\eqref{eq:hybrid_q} is more accurate than that of DAVI. Third, the maximum ELBO value is the same
for both SG-HVI and NG-HVI because they employ the same VA. 
Last, when compared to the exact posterior computed using MCMC, the
variational posteriors from NG-HVI are
more accurate than those from DAVI. This can be seen in Figure~A1 in Part~C of the Web Appendix, which 
plots the exact and variational marginal posteriors of $\thetavec$.


Datasets were generated using
different values of $K$ and ratios $\sigma^2_\alpha/\sigma^2_\epsilon$, and
the accuracy of DAVI and NG-HVI measured by the 
average of $\widehat{\calL(\lambdavec^{(t)}) }$ over the last
100 steps of the NGA algorithms.
Table~\ref{summary_elbo} reports these values for DAVI, along
with the difference with NG-HVI. The latter is computed for $p \in \{0,1,2,3\}$ factors for the Gaussian factor approximation
$q^0$ to illustrate the robustness of the hybrid VI method. The case
$p=0$ corresponds to a mean field Gaussian approximation
for the marginal VA in $\thetavec$, although we stress this is not a mean field
approximation for $\psivec$.
NG-HVI is more accurate than DAVI, with the improvement being greatest for $K=1000$. 
The choice of $p$ has little impact on the NG-HVI results, with only 
slightly higher ELBO values for larger $p$. 
Further details on this example can be found in Part~C of the Web Appendix. 

\begin{table}[thb]
	\centering
	\captionsetup{skip=0pt}
	\begin{threeparttable}
		\caption{Improvement in ELBO of NG-HVI over DAVI  for different DGP settings and number of factors $p$ in Example~1. }
		\begin{tabular}[ht!]{llcccccc}
			\toprule
			\multicolumn{2}{c}{DGP Setting}&	\multirow{2}{*}{$ELBO_{DAVI}$}  &&\multicolumn{4}{c}{$ELBO_{HVI}-ELBO_{DAVI}$}\\
			\cmidrule{5-8}
			$K$ & {$\sigma^2_\alpha / \sigma^2_\epsilon$}&&&$p=0$ & $p =1 $ &  $p=2$ &  $p=3$\\
			\hline
			\multirow{4}{*}{100}&$0.01$ &-7327.1   &&  6.2   &  7.0   &  7.2  &  7.7\\

			&$1$ &-7470.2  &&   3.0 &   3.8  &  3.8 &   4.4 \\
			
			&$10$ &-7583.8  &&   0.7 &   1.2   & 1.9  &  2.1\\
			\hline
			\multirow{4}{*}{500}&$0.01$ &  -7235.4   &&  8.6 &   8.9  &  9.6  &  9.1\\
			
			&$1$ &-7789.4   &&  7.9  &  8.2  &  9.1  &  8.9 \\
			
			&$10$ &-8342.3  &&   3.8  &  4.3  &   5.0   &  4.7\\
			\hline
			\multirow{4}{*}{1000}&$0.01$ & -7252.4  &&   56.3  &  57.1   & 57.1  &  57.3\\
			&$1$ &-8101.0   &&   54.0  &   54.8   & 54.6  &  54.8 \\
			
			&$10$ &-9166.6   & &   28.4  &   28.8  &   29.2   &  29.4\\
			\bottomrule
		\end{tabular}
		Note: Positive numbers in the last four columns indicate improved accuracy of NG-HVI relative to DAVI.  Results are based on 10000 steps to ensure convergence.
		\label{summary_elbo}
	\end{threeparttable}
\end{table}

\section{Hybrid Variational Inference for Deep Mixed Models}\label{sec:hvidmm}
\subsection{Deep mixed models}
Bayesian neural networks are probabilistic neural networks estimated using 
Bayesian inference; see~\cite{jospinHandsOnBayesianNeural2022} for an 
introduction. Following~\cite{tranBayesianDeepNet2020},
we consider a Bayesian neural network with random output layer coefficients to allow for group-level heterogeneity, which these authors call 
a deep mixed model (DMM). 

For a feed forward
network with $L$ layers, the DMM we consider draws observation 
$i$, belonging to group $k$, from
\begin{eqnarray}
	y_{i} &\sim & {\cal D}\left(f_{L+1}\left((\betavec + \alphavec_k)^\top \hvec^{(L)}_i  \right)\right)\;\mbox{ for }i=1,\dots,n\nonumber\\
	\hvec^{(l)}_i & = &f_l\left(W_l \hvec^{(l-1)}_i\right)\;\mbox{ for }l=1,\ldots,L\,,  \label{eq:dmm}
\end{eqnarray}
where $f_l$ is an activation function, $\hvec_i^{(l)}$ for $l\geq 1$ is
the vector of $l^{th}$ layer values for observation $i$ with the first element a
constant off-set, 
$\hvec^{(0)}_i = \xvec_i$ is the input vector, and $W_l$ is the weight matrix at layer $l$. 
The output value $y_i$ has probability 
distribution ${\cal D}$ with parameters that are a function of 
$h^{(L+1)}_i=f_{L+1}\left((\betavec + \alphavec_k)^\top\hvec^{(L)}_i  \right)$, where
we assume a linear activation function $f_{L+1}$ in our examples.
The output layer coefficients are 
 decomposed into a fixed effect term $\betavec$ and a random effects term $\alphavec_k\sim N(\bm{0},\Omega_\alpha)$ that varies
over group  $k=1,\ldots,K$. 
The random effects allow translation of the nodes from layer $L$ into the output layer to be 
heterogeneous over the $K$  groups. 
The introduction of random coefficients into the output layer 
of other deep neural networks is similar.

\subsection{NG-HVI for DMM} \label{sec:nghvidmm}
The unknowns in the DMM model at~\eqref{eq:dmm} include
weights $\wvec=(\mbox{vec}(W_1)^\top,\ldots,\mbox{vec}(W_L)^\top)^\top$, 
fixed effects $\betavec$, random effects values
$\alphavec_1,\ldots,\alphavec_K$, covariance matrix $\Omega_\alpha$
and any other parameters used to specify the distribution ${\cal D}$. 
Bayesian estimation involves selecting priors for all unknowns
and then evaluating their joint posterior distribution. 
The selection 
of appropriate priors for weights in Bayesian neural networks is an area of current research; see~\cite{tranJMLR22} for a discussion.  Here, we adopt the simple
priors $\wvec \sim N(\bm{0},\sigma_w^2 I)$ and $\betavec\sim N(\bm{0},\sigma^2_\beta I)$ with $\sigma^2_w=\sigma^2_\beta=100$.  
If $\mbox{dim}(\hvec_i^{(l)}) =m_l$, then  
following~\cite{tanAnalyticNaturalGradient2022} a prior 
$\Omega_\alpha^{-1}\sim \mbox{Wishart}(0.01I,\nu)$ is adopted for the
precision matrix, so that the prior for
$\Omega_\alpha$ is an inverse Wishart, with $\nu=m_L+1$.  We stress that other priors for $\wvec,\betavec,\Omega_\alpha$ can also be  
adopted just as readily when using our black box VI method.

Because $K$ is often large in practice, computation
of the posterior can be challenging. However, 
setting $\zvec=(\alphavec_1^\top,\ldots,\alphavec_K^\top)^\top$ and $\thetavec$ to
a vector comprising all the other unknowns,
the proposed NG-HVI method is well-suited to train such DMMs. To implement step~(c) of Algorithm~\ref{alg:nghvi}, evaluation of the gradient $\nabla_\theta \log g(\thetavec,\zvec)$ is required.  
The gradient with respect to $\wvec$ can be evaluated using back-propagation~\citep{rumelhart_backprop_1986}. 
We follow~\cite{tanUseModelReparametrization2021} and re-parameterize
$\Omega_{\alpha}^{-1}=LL^\top$ using its Cholesky factor $L=\{l_{ij}\}$ and 
set $\lvec$ to a vector of the elements $\{l_{ij}; j<i\}\cup \{\log(l_{ii});i=1,\ldots,m_L\}$. These elements are
on the real line, and both the prior for $\lvec$ and the gradient
$\nabla_l  \log g(\thetavec,\zvec)$ are available in closed form; see 
Appendix~\ref{app:omega}.
The gradients with
respect to the other elements of $\thetavec$ depend on the choice for ${\cal D}$, but are usually available 
either in closed
form or using numerical differentiation.  
Another requirement
to implement the algorithm is for generation from $p(\zvec|\thetavec,\yvec)$ at step~(b) to be feasible, which is also contingent on choice ${\cal D}$.
  
The implementation of NG-HVI  is outlined below for two DMMs with Gaussian and
Bernoulli distributions ${\cal D}$ for the output variable. The performance of the method
is assessed using data simulated from 
two examples, where the first DGP matches the DMM fit, while the second DGP does not match the DMM fit (so that there is model mis-specification). 
SG-HVI and two existing natural gradient VI methods act as benchmarks.
Performance is evaluated using the posterior predictive distribution, 
computed as outlined in Part~D of the Web Appendix.

\subsection{Example 2: Gaussian DMM}\label{sec:gmmnormal}
In this section, we consider two DMMs with the Gaussian output distribution
\begin{align}
	y_{i} \sim N\left((\betavec + \alphavec_k)^\top \hvec^{(2)}_{i}, \sigma^2_\epsilon \right).
	\label{eq:GaussDMM}
\end{align}
We demonstrate the strong performance of NG-HVI in a small Gaussian DMM in Example 2(a), where it is possible
	to carefully compare with DAVI, and in a larger Gaussian DMM in Example~2(b).
\subsubsection{Example 2(a): Smaller model}\label{subsec:gmmnormal1}
We generate data from the DMM at~\eqref{eq:dmm} with $L=2$ hidden layers, each with 5 neurons and offset (so $m_1=m_2=6$) and the ReLU activation function. 
There are $K=1000$ random effect groups, with 6 observations drawn from 
each to form training data, and a further 2 observations drawn from each group to form test data.  
The input vector $\xvec_i$ also consists of $m_0=6$ values, with 
unity the first element and the remaining $5$ values generated from a correlated multivariate Gaussian distribution.
	We fix $\sigma^2_\epsilon=20$ and $\Omega_\alpha$ to a $6 \times 6$ diagonal matrix. 
Further details on this DGP are given in Part~D of the Web Appendix.

This DMM is estimated using SG-HVI, NG-HVI and DAVI. 
 All three algorithms have a marginal Gaussian VA for $\thetavec$,
 where SG-HVI and NG-HVI employ a factor covariance structure with $p=3$, while DAVI uses an unrestricted covariance. 
 We did not study how $p$ affects the accuracy of the VA, although~\cite{ongGaussianVariationalApproximation2018} found in their examples that low values were sufficient.
To implement the HVI estimators, at step~(b) of Algorithm~\ref{alg:nghvi}
the density $p(\zvec|\thetavec,\yvec)=\prod_{k=1}^K p(\alphavec_k|\thetavec,\yvec)$ is a product of Gaussians, from which it
is fast and simple to draw. The prior for $\sigma^2_\epsilon$ is a standard
Inverse Gamma distribution with shape and scale parameters equal to 1.01. 
We also implemented the natural gradient algorithm of~\cite{tranBayesianDeepNet2020}, labeled ``NAGVAC'', using the code provided by the authors, but found that in these examples the algorithm with default setup did not work well, giving  poor results, so we do not include them here.

\begin{figure}[ht!]
	\centering
	\begin{subfigure}[t]{0.49\linewidth}
		\subcaption{Convergence of noisy ELBO.}		
		\vspace{3pt}
		\includegraphics[width=1\linewidth]{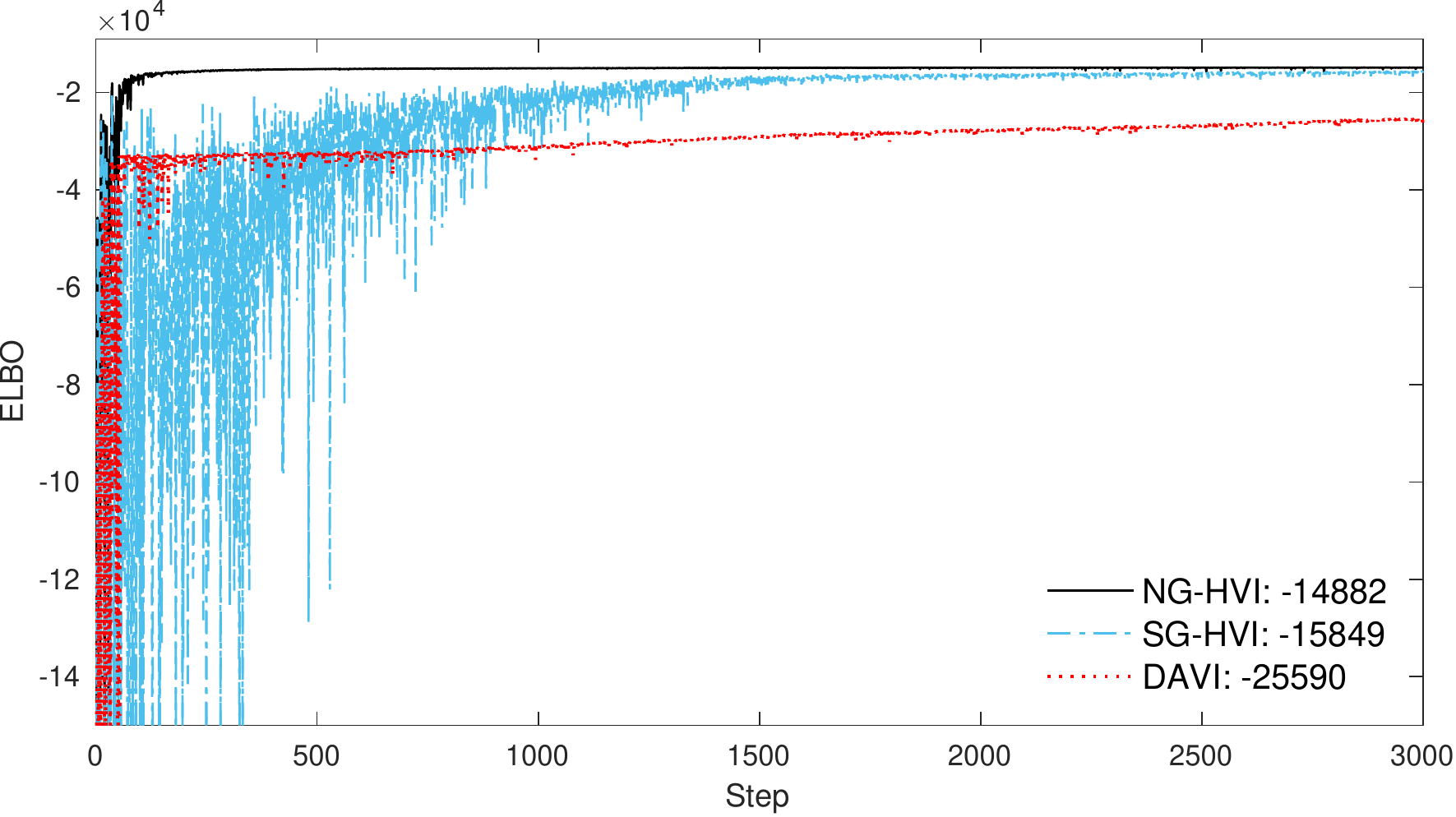}
	\end{subfigure}
	\begin{subfigure}[t]{0.5\linewidth}
		\subcaption{Predictive Robustness}
		\vspace*{3.1mm}
		\includegraphics[width=1\linewidth]{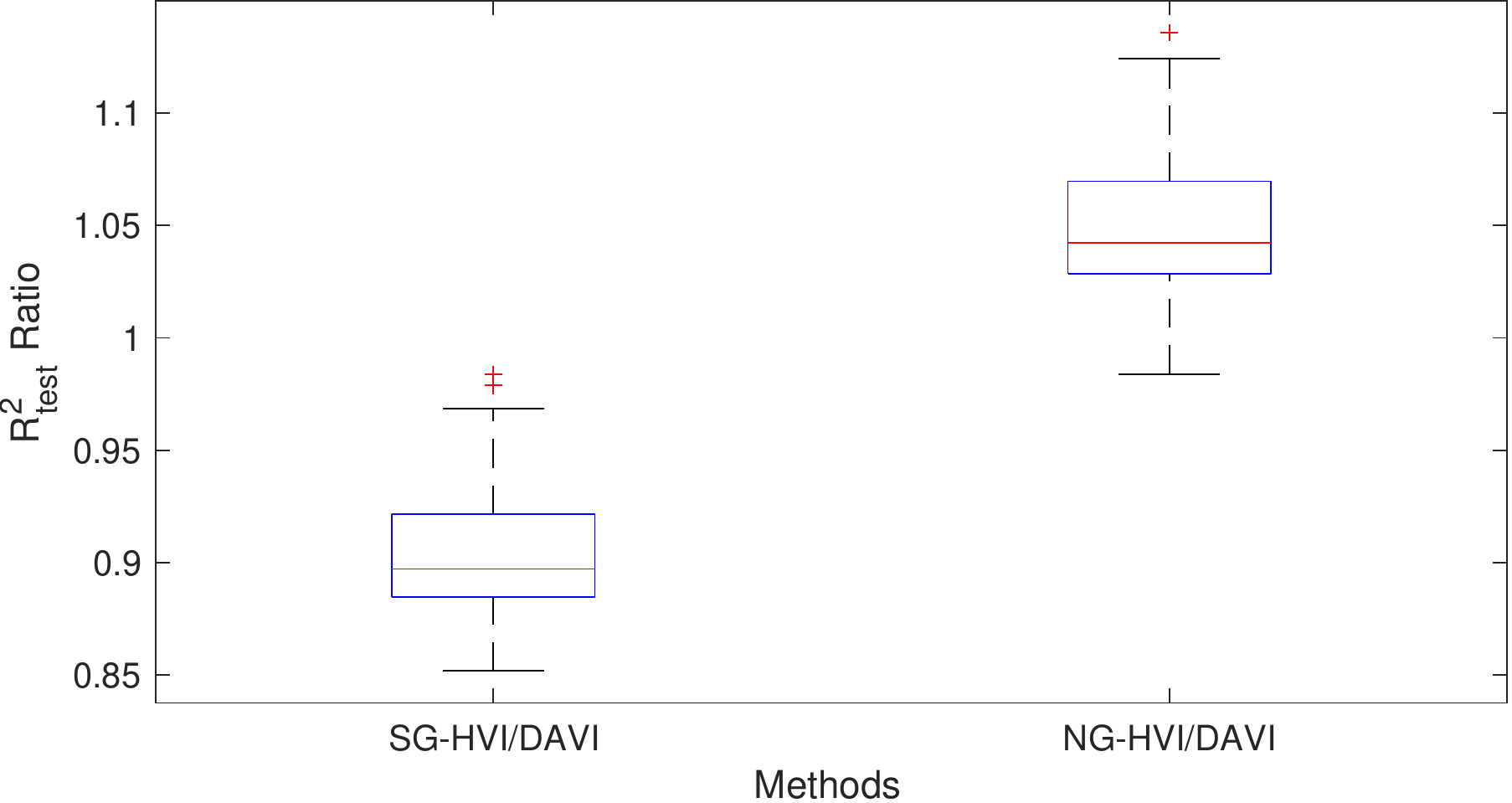}
	\end{subfigure}
	\caption{Simulation results from the Gaussian DMM in Example 2(a). Panel~(a) depicts convergence of the noisy ELBO against optimization step number. 
	Panel~(b) depicts boxplots of the out-of-sample predictive $R^2$ (displayed as a ratio of those from the HVI methods over those from DAVI) resulting from 100 repeated simulated datasets.}
	\label{fig:eg2}
\end{figure}

Figure~\ref{fig:eg2}(a) plots the noisy ELBO~\eqref{eq:noisyL} for DAVI and
the two HVI methods. 
The two natural gradient methods (DAVI and NG-HVI) converge in fewer steps than the first order method SG-HVI. The NG-HVI method reaches a higher maximum ELBO value after 3000 iterations. 
The top half of Table~\ref{gaussiandmm} reports the
predictive accuracy from the fitted DMMs, summarized
using the $R^2$ coefficient for both the training and test data, and
NG-HVI dominates.
We also optimised the model using SG-HVI with 10000 steps, 
and found that by using more steps SG-HVI can achieve similar results as NG-HVI.

One hundred replicate datasets were generated from the DGP,
and the same three methods were used to fit these datasets, with each method optimized over 3000 steps.
Figure~\ref{fig:eg2}(b) gives boxplots of the $R^2$ values
for the test data predictions, where results for the HVI methods are displayed relative to those of DAVI. Ratios greater
	than one indicate greater predictive accuracy than DAVI and we conclude that the dominance of NG-HVI 
observed using the single dataset is a robust result.

\begin{table}[h]
	\centering
	\captionsetup{skip=0pt}
	\begin{threeparttable}
		\caption{Predictive Accuracy for the Gaussian DMM in Example~2}
		\label{gaussiandmm}
		\begin{tabular}{lcccc}
			\toprule
			& \multicolumn{4}{c}{Estimation Method}\\ 
			&DAVI  &   SG-HVI 	&  NG-HVI &  SG-HVI\\ \hline
		   \multicolumn{3}{l}{Example 2(a): Smaller Model} && \\ \cline{1-3}
			$R^2_{train}$  & 0.7896  &   0.7527 &    0.8097   &  0.8099    \\
			$R^2_{test}$   & 0.6045  &   0.5425  &   0.6334   &  0.6332     \\
			Total steps	   & 3000	&    3000     &   3000 &   10000    \\
			Time to fit (min) & 25.7 &   2.5    &  2.6 &    8.3     \\
		   \multicolumn{3}{l}{Example 2(b): Larger Model} && \\ \cline{1-3}
			${R^2_{train}}$  & -- &    0.6997  &   0.7574  &   0.7382  \\
			${R^2_{test}}$   &-- &    0.6414  &   0.6881   &   0.6740 \\
			Total steps	   & --	&    3000     &   3000&  10000   \\
			Time to fit (min) & -- &   9.4 &  133.2 &   31.3  \\
			\bottomrule
		\end{tabular}
		Note: Computation times are for a 2022 HP desktop with Intel i9 processor. Time per step is for variational calibration. For Example 2(b), DAVI takes more than one day on a standard desktop for $3000$ steps, making it impractical for model of such scale, and thus it is excluded. SG-HVI is employed twice, with either $3000$ or $10,000$ steps.\\
	\end{threeparttable}
\end{table}

\subsubsection{Example 2(b): Larger model}\label{subsec:gmmnormal2}
We now extend the previous Gaussian DMM to have 64 inputs and 2 hidden layers, with 32 and 16 neurons, respectively. 
	The number of neurons in each layer follows the geometric pyramid rule as suggested by~\cite{guEmpiricalAssetPricing2020}.
	The training data comprises $K = 1000$ random coefficient groups, each with $30$ observations, while the testing data has a further $10$ observations in each group. We fix $\sigma^2_\epsilon = 2.25$ and set $\Omega_\alpha$ to a $17 \times 17$ diagonal matrix.
	Further details of the DGP are available in the Supplementary Information.

\begin{figure}[ht!]
\centering
 \includegraphics[width=0.8\linewidth]{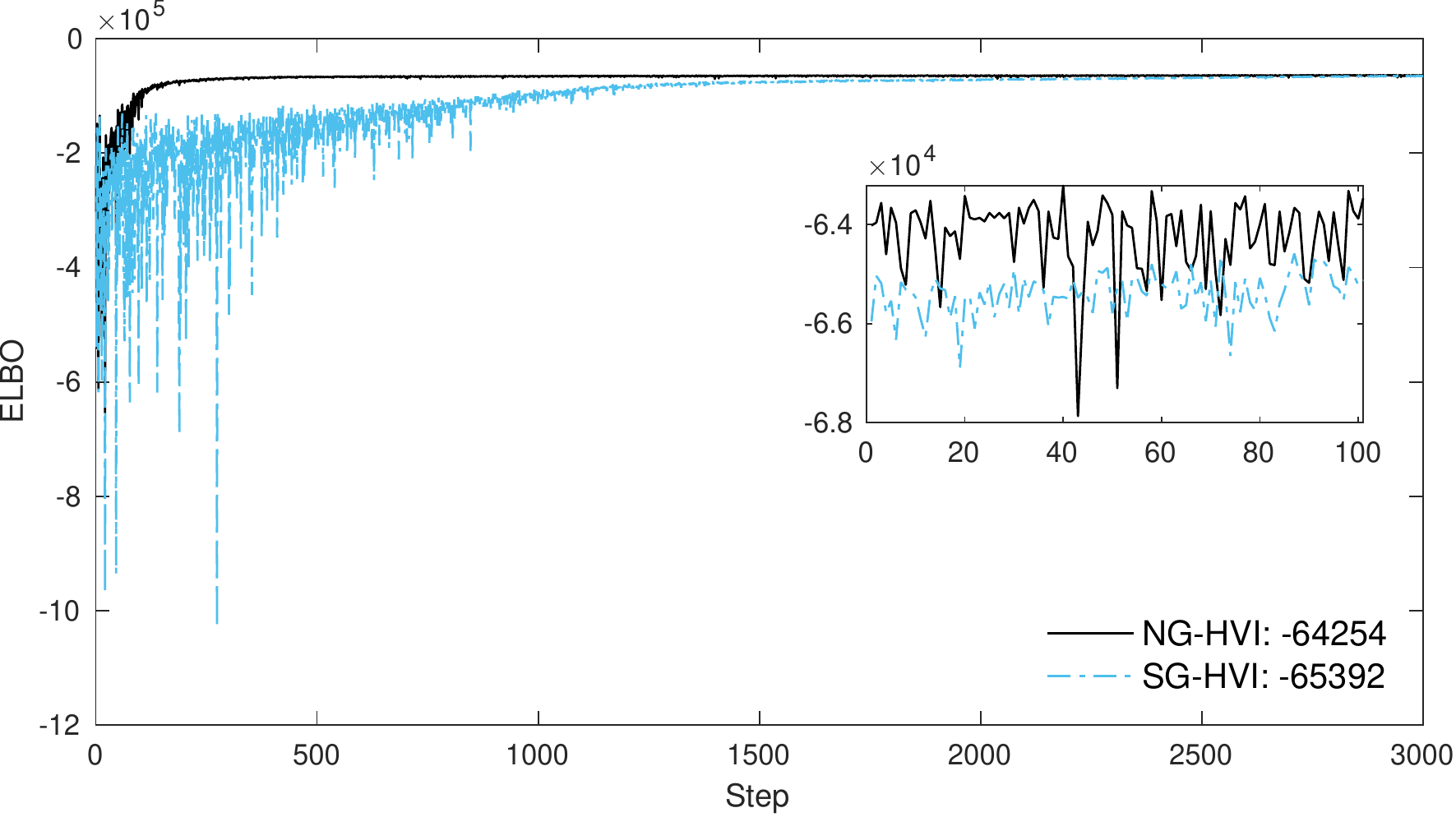}
 \caption{Plots of the noisy ELBO values against optimization step number for the larger Gaussian DMM in Example~2(b).}
 \label{fig:eg2b}
\end{figure}

This model has $\mbox{dim}(\thetavec)=2779$ global parameters and $\mbox{dim}(\zvec)=17,000$ latent variables, which is too
large to estimate using DAVI. When using the NG-HVI method with the Gaussian approximation with a $p=3$ factor covariance
matrix for $q^0(\thetavec)$, the number of variational parameters is $\mbox{dim}(\lambdavec)=13,895$, leading a large FIM at~\eqref{eq:FIMpattern}. The bottom half of Table \ref{gaussiandmm} reports the results from applying NG-HVI and SG-HVI to estimate
the model. Our natural gradient approach takes just over two hours to fit the model on a standard desktop. In contrast, while SG-HVI is faster per
step, even in 10,000 steps the fitted model is substantially less accurate. To illustrate why, Figure~\ref{fig:eg2b} compares the noisy ELBO of NG-HVI against SG-HVI, showing that the latter exhibits much slower convergence to the optimum.
	
\subsection{Example 3: Bernoulli DMM}\label{sec:nghvidgmm}
In this example we compare the performance of the two HVI methods and NAGVAC. We replicate the example given in Section~6.1.4 of~\cite{tranBayesianDeepNet2020}, 
where data is generated from a Bernoulli distribution with a parameter
that is a smooth nonlinear function of five covariates and a single scalar random effect; see Part~D of the Web Appendix
for the full specification of this DGP. The DGP deviates from the original
only by the use of a logistic link function,
rather than a probit link function, so as to advantage
the NAGVAC estimator which employs a logistic link. 
There are $K=1000$ random effect groups, and from each we draw
14 observations to form training data, a further 3 observations to form 
test data, and an additional 3 observations as validation data to implement the NAGVAC stopping rule.

\begin{table}[thb!]
	\captionsetup{skip=0pt}
	\centering
	\begin{threeparttable}
		\caption{Predictive metrics for the Bernoulli DMM in Example~3}
		\label{resultstrandata}
		\begin{tabular}{lcccccccccc}
			\toprule
			&&{Naive} &&{NAGVAC}  &&{NAGVAC} && {SG-HVI}  &&{NG-HVI}\\
			&& &&{\em (stopping rule)} &&{\em (no rule)} && && \\
			\hline
			$PCE_{train}$	&&0.4992 &&   0.4489  &&      0.1541    &&     0.1391  &&   0.1220\\

			$PCE_{test}$	&&0.6958 &&   0.4502  &&      0.1545    &&      0.1460 &&   0.1241	 \\

			$F1_{train}$	&&0.8693 &&   0.8705 &&       0.9567   &&      0.9624 &&   0.9673 \\

			$F1_{test}$		&&0.8605 &&   0.8696  &&      0.9587  &&       0.9608  &&  0.9648\\

			Time per step (secs)&& -- 	&& 33.59   && 33.59	 && {0.48} && {0.50} \\
			
			Total steps		&& -- 	&& 42	   && 3000		&&{3000} 	 &&  {3000} \\
			
			Total time(min) && --	&& 23.51    && 1679.45	&&  {24.23}&&  {24.88}\\
			\bottomrule
		\end{tabular}
		Note: Low PCE values and high F1 values correspond to greater predictive accuracy. For the VI methods the learned Bernoulli DMMs use the variational posterior mean of the parameters. The NAGVAC method  is implemented with and without the stopping rule of~\cite{tranBayesianDeepNet2020}. 
		Computation times are for a 2022 HP desktop with Intel i9 processor. 
	\end{threeparttable}
\end{table}

We use NG-HVI to learn the DMM at~\eqref{eq:dmm}
with $L=2$ hidden layers, each with 5 neurons and an offset (so $m_1=m_2=6$),
and ReLU activation functions. 
If $\epsilon_i\sim N(0,1)$, the
 output $y_i=\mathds{1}(y^*_i>0)$ is determined by the latent variable
\[
y^*_i =(\betavec + \alphavec_k)'\hvec^{(2)}_{i}+\epsilon_i  \,,
\]
which corresponds to adopting a Bernoulli distribution for ${\cal D}$ 
with $f_{L+1}(x)=\Phi(x)$. Therefore, this example features model mis-specification because the DMM  does not encompass
the DGP.

To implement HVI, the latent variable vector is  
$\zvec=(\alphavec^\top,\yvec^*)$ with $\yvec^*=(y^*_1,\ldots,y^*_n)^\top$.
At step~(b) of Algorithm~\ref{alg:nghvi}, an MCMC scheme with five sweeps is used that
draws alternately from the conditional posteriors
of $\alphavec$ and $\yvec^*$, initialized at their values from the last step of the optimization. This is fast to implement, with details given in Part D
of the Web Appendix. The NAGVAC algorithm
is used to estimate the same DMM, but with a logistic link and $\Omega_\alpha$ a diagonal matrix, which matches the DGP to advantage this approach. 

Comparison of the
different methods is based on their predictive accuracy for both
the training and test data.
Table~\ref{resultstrandata} reports the $F1$ score
and the predictive cross entropy (PCE) defined by
\cite{tranBayesianDeepNet2020}, where low PCE values and high F1 values correspond to greater predictive accuracy. 
Results for na\"ive predictions using group-specific
proportions are provided as a benchmark. Results
are provided for NAGVAC using the stopping rule given by~\cite{tranBayesianDeepNet2020}, and also without the rule and 3000 optimization steps. 
The computational time using a standard desktop is also provided for all methods.
We make three observations about the Bernoulli DMMs learned using the different methods.
First, using NAGVAC with stopping rule has poor
accuracy. Second, while using NAGVAC without the stopping rule
improved accuracy, the computing time required is excessive.
Third, both HVI methods are faster and more accurate. 
 
\begin{figure}[thb!]
	\caption{Simulation results for the Bernoulli DMM in Example~3.}
	\begin{subfigure}{\linewidth}
		\centering
		\includegraphics[width=0.7\linewidth]{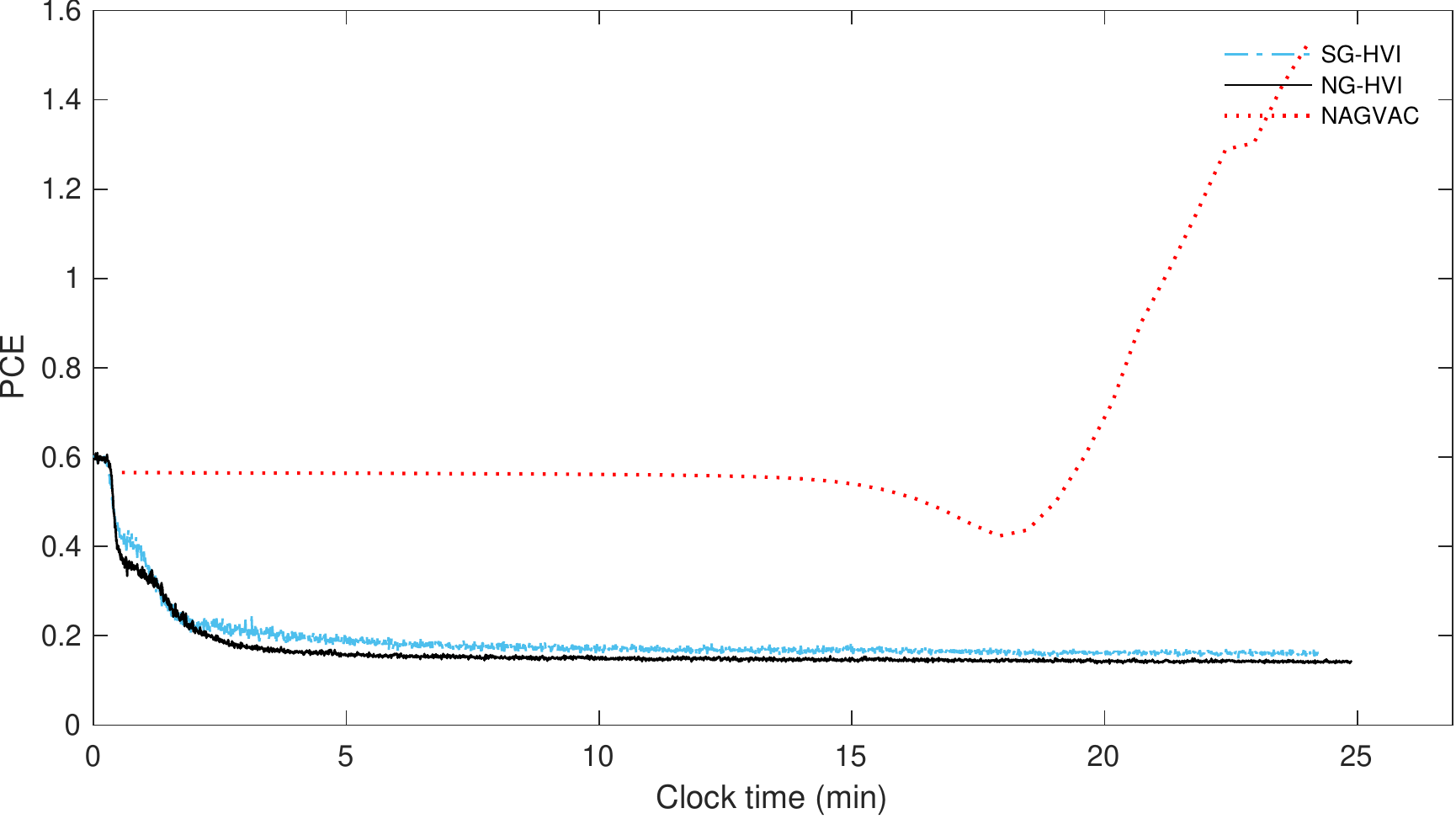}		
		\subcaption{Convergence of predictive cross entropy for the test data using
			the three methods.}
	\end{subfigure}
	
	\begin{subfigure}{\linewidth}
		\centering
		\vspace{15pt}
		\includegraphics[width=0.45\linewidth]{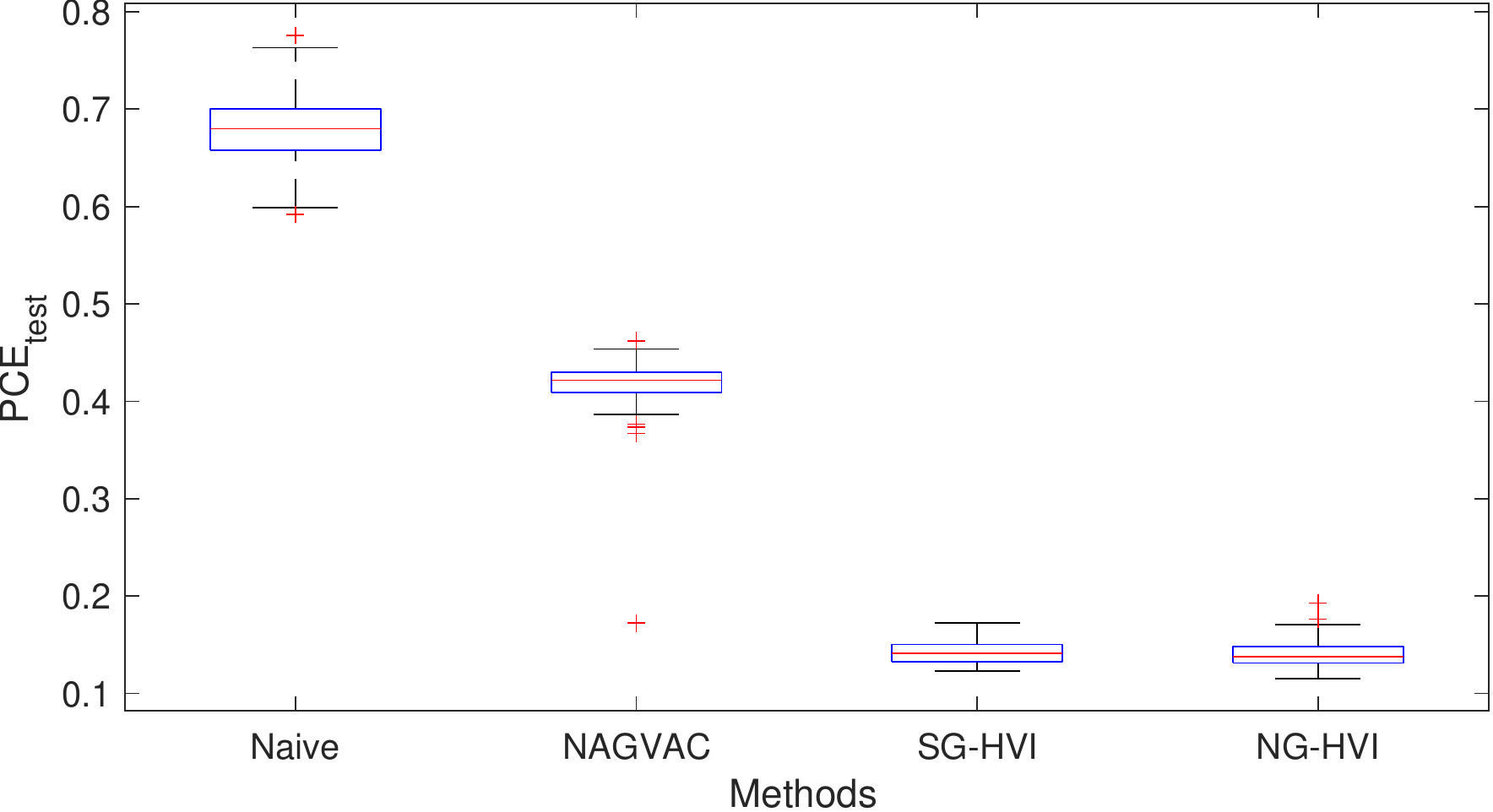}
		\includegraphics[width=0.45\linewidth]{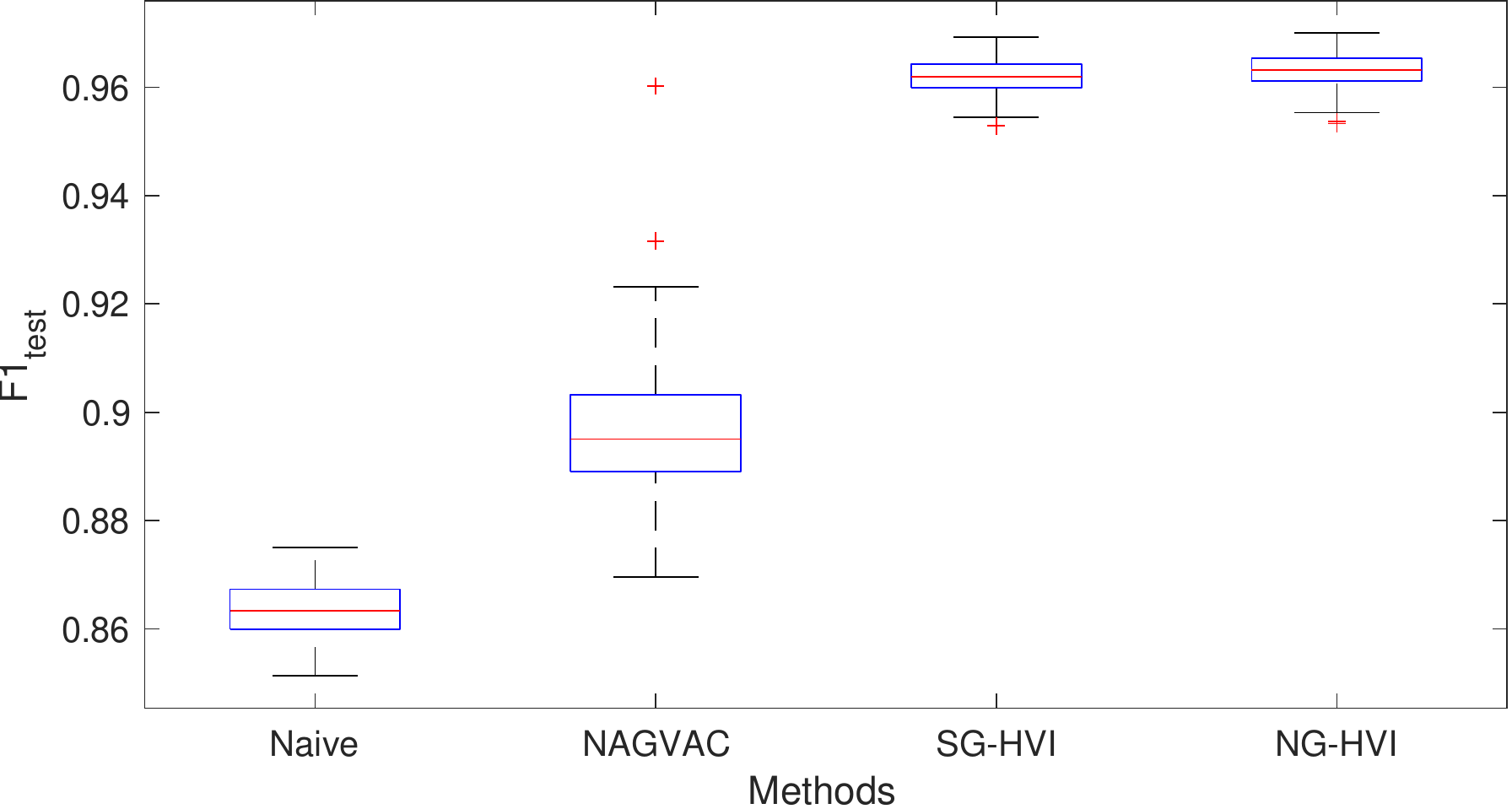}
		\subcaption{Boxplots of test data posterior predictive metrics for Example~3 from 100 simulated datasets. Lower PCE values and higher F1 values indicate increased prediction accuracy. }	\end{subfigure}
	\label{fig:eg3}
\end{figure}

To further illustrate, Figure~\ref{fig:eg3}(a) plots the PCE metric
against clock time for the methods. The poor convergence of NAGVAC
can be seen, as can the fast convergence of NG-HVI. 
To show the robustness of these results, 100 replicate datasets are generated and the methods used to learn the Bernoulli DMM for each.
Figure~\ref{fig:eg3}(b) gives boxplots of the PCE and F1 metrics for the 
test data. Due to the excessive computation time, only the NAGVAC results with stopping rule are given.

\section{DMM  for Financial Asset Pricing}\label{sec:finance}
There is growing interest in the use of machine learning methods
for asset pricing in the financial literature
\citep{fangMachineLearningBased2021,feng2022deep}. Compared to traditional linear models, 
there is evidence that neural networks can increase the predictive accuracy for asset returns~\citep{guEmpiricalAssetPricing2020}, as well as there being heterogeneity across industries \citep{dialloMachineLearningApproach2019,guEmpiricalAssetPricing2020}. 

To explore both features, we
estimate Gaussian DMMs with the output layer given at~\eqref{eq:GaussDMM}, 
and either the three factors of~\cite{famaCommonRiskFactors1993} (hereafter ``FF3''), or the five factors of~\cite{famaFivefactorAssetPricing2015} (hereafter ``FF5''), as inputs. 
The three factors in FF3 are 
the excess market return (ERM), a firm size factor (SMB) and a value premium (HML). 
The five factors in FF5 extends these to include a firm profitability factor (RMW) and an  firm investment strategy factor (CMA). The random effect groups are defined by the 4-digit Security Industry Code (SIC), 
so that the DMM captures industry variation in the factor risk premia.
Two architectures are used. The first has one hidden layer of 8 neurons,
and the second has 3 hidden layers of 32, 16 and 8 neurons as used by~\cite{guEmpiricalAssetPricing2020}. In this example we use NG-HVI with a factor covariance structure for $q^0$ with $p=3$ for a DMM with one hidden layer, and $p=5$ for a DMM with three hidden layers.
We do not identify an optimal architecture, and are unaware of other studies that do so for asset pricing models.


\begin{table}[thb!]
	\centering
	\captionsetup{skip=0pt}
	\caption{The ten most populated 4-digit SIC groups between 2005 and 2019}
	\begin{tabular}{llr}
		& & \\
		\hline \hline
		SIC &Industry Description & No. of Companies\\
		\hline
		6726 &  Unit Investment Trusts, Face-Amount Certificate Offices,  &  \\
		&   and Closed-End Management Investment Offices & $280$ \\
		6722 & Management Investment Offices, Open-End & $116$ \\
		
		6798 & Real Estate Investment Trusts & $100$ \\
		
		6020 &Commercial Banks, Not Elsewhere Classified & $68$ \\
		2834 &Pharmaceutical Preparations & $58$ \\
		6022 &State Commercial Banks & $52$ \\
		6021 & National Commercial Banks & $51$\\
		1311 &Crude Petroleum and Natural Gas & $41$\\
		7372 &Prepackaged Software & $37$ \\
		6331 &Fire, Marine, and Casualty Insurance & $ 34$ \\ 
		\hline \hline
	\end{tabular}
	\label{sic_industry}
\end{table}

Our output data are monthly returns between 2005 and 2019 (inclusive) for the 2583 U.S. companies in the Centre for Research in Security Prices (CRSP) database that were publicly listed during the 
entire period. We standardized all data before fitting the model.
The first 10 years are used for training,
and the last 5 years for testing.
The monthly factor values were obtained from the Kenneth French's data library.\footnote{Available at the time of writing at {\tt http://mba.tuck.dartmouth.edu}.} 
In total, there are 548 industry groups, of which 260 groups feature only one company in the sample period, and 41 groups feature more than 10 companies.  
Table~\ref{sic_industry} summarizes the 10 most populated 4-digit SIC groups in our sample period.


\begin{table}[h]
	\centering
	\captionsetup{skip=0pt}
	\begin{threeparttable}
		\caption{FF3 and FF5 Probabilistic Asset Pricing Models }
		\label{tab:summary_model}
		\begin{tabular}{p{0.25\linewidth} p{0.7\linewidth}}
			\hline \hline
			Model Label &  Model Description\\
			\hline
			Linear 				& Gaussian linear regression.\\
			Random Intercept	& Gaussian linear regression with random intercept.  \\
			Random Coefficients		& Gaussian linear regression with random intercept and coefficients. \\ 
			FNN					& Deep feed forward neural network model, estimated using MATLAB ``train'' routine. This has a Levenberg-Marquardt training function,  mean squared error performance function, and ReLU activation functions for all layers except the output layer, which uses linear activation. For probabilistic prediction the residuals are assumed independent Gaussian.\\  
			DMM				& Same architecture as FNN, with random coefficients for the output layer, estimated using NG-HVI. \\
			
			\hline \hline
		\end{tabular}
	\end{threeparttable}
\end{table}

\subsection{Predictive accuracy}
Table~\ref{tab:summary_model} outlines five  probabilistic asset
pricing models. Their predictive accuracy for both the training and test data is measured
using the $R^2$ metric for point predictions, and the log-score (LS) for probabilistic 
forecasts;\footnote{The log-score is defined as the mean of
	the logarithm of predictive densities evaluated at the observed test data
	values.} higher values of both correspond to greater accuracy.
Table~\ref{ff_result_table_mdl} reports values for the five predictive models and both
the FF3 and FF5 cases. Focusing on the test data results, we make four observations. First, introducing a random
intercept to a linear model does not improve results, and neither does the LS improve
when introducing random coefficients. Second, the FNN does not improve predictive
accuracy over-and-above the traditional linear models. Third, the three hidden layer
architecture dominates the shallow single layer. Last,  the DMM produces a substantial
improvement in predictive accuracy compared to all alternatives when using the three hidden layer architecture.  

\begin{table}[h]
	\centering
	\captionsetup{skip=0pt}
	\begin{threeparttable}
		\caption{Predictive accuracy of the five competing models.} 
		\label{ff_result_table_mdl}
		\begin{tabular}{lcccc}
			\hline \hline
			&\multicolumn{4}{l}{Fama-French 3 Factors}\\
			\cmidrule{2-5} 
			&  $LS_{train}$  &  $LS_{test}$ &$R^2_{train}$  &  $R^2_{test}$  \\
			\hline
			Linear 				& -1.3297 & -1.2502 & 0.1635 & 0.1073   \\
			Random Intercept	& -1.3297 & -1.2502 & 0.1635 & 0.1073  \\
			Random Coefficients		& -1.3191 & -1.2503 & 0.1953 & \textbf{0.1151}  \\ 
			FNN(LM,[8])						& -1.3277 & -1.2589 & 0.1668 & 0.0879 \\
			DMM(NG-HVI,[8])					& -1.2833 & -1.2527 & \textbf{0.2128} & 0.0583  \\
			
			FNN(LM,[32 16 8])				&		 -1.3319 & -1.2530 & 0.1688 & 0.1011 \\
			DMM(NG-HVI,[32 16 8])					& \textbf{-1.2812} & \textbf{-1.2320} & 0.1934 & \textbf{0.1151}  \\
			\hline
			\hline
			&\multicolumn{4}{l}{Fama-French 5 Factors}\\
			\cmidrule{2-5} 
			&  $LS_{train}$  &  $LS_{test}$  &$R^2_{train}$  &  $R^2_{test}$  \\
			\hline
			Linear 				& -1.3294 & -1.2511 & 0.1639 & 0.1052  \\
			Random intercept	& -1.3294 & -1.2511 & 0.1639 & 0.1052  \\
			Random coefficients		 & -1.3178 & -1.2513 & 0.2018 & 0.1051 \\ 
			FNN(LM,[8])						& -1.3279 & -1.2461 & 0.1665 & 0.1018 \\
			DMM(NG-HVI,[8])					& \textbf{-1.2768} & -1.2477 & \textbf{0.2284} & 0.0533  \\
			FNN(LM,[32 16 8])						& -1.3269 & -1.2447 & 0.1697 & 0.1016 \\
			DMM(NG-HVI,[32 16 8])					& -1.2871 & \textbf{-1.2366} & 0.1924 & \textbf{0.1118}  \\
			\hline \hline
		\end{tabular}
For the FNN and DMM models two architectures are considered. The highest metric value (indicating increased accuracy) is highlighted in bold in each column.
	\end{threeparttable}
\end{table}

\subsection{The implied heterogeneous relationship from DMM}
\begin{figure}[ht!] 
	\begin{subfigure}{0.33\linewidth}
		\subcaption{ERM (low)}
		\includegraphics[width=1\linewidth]{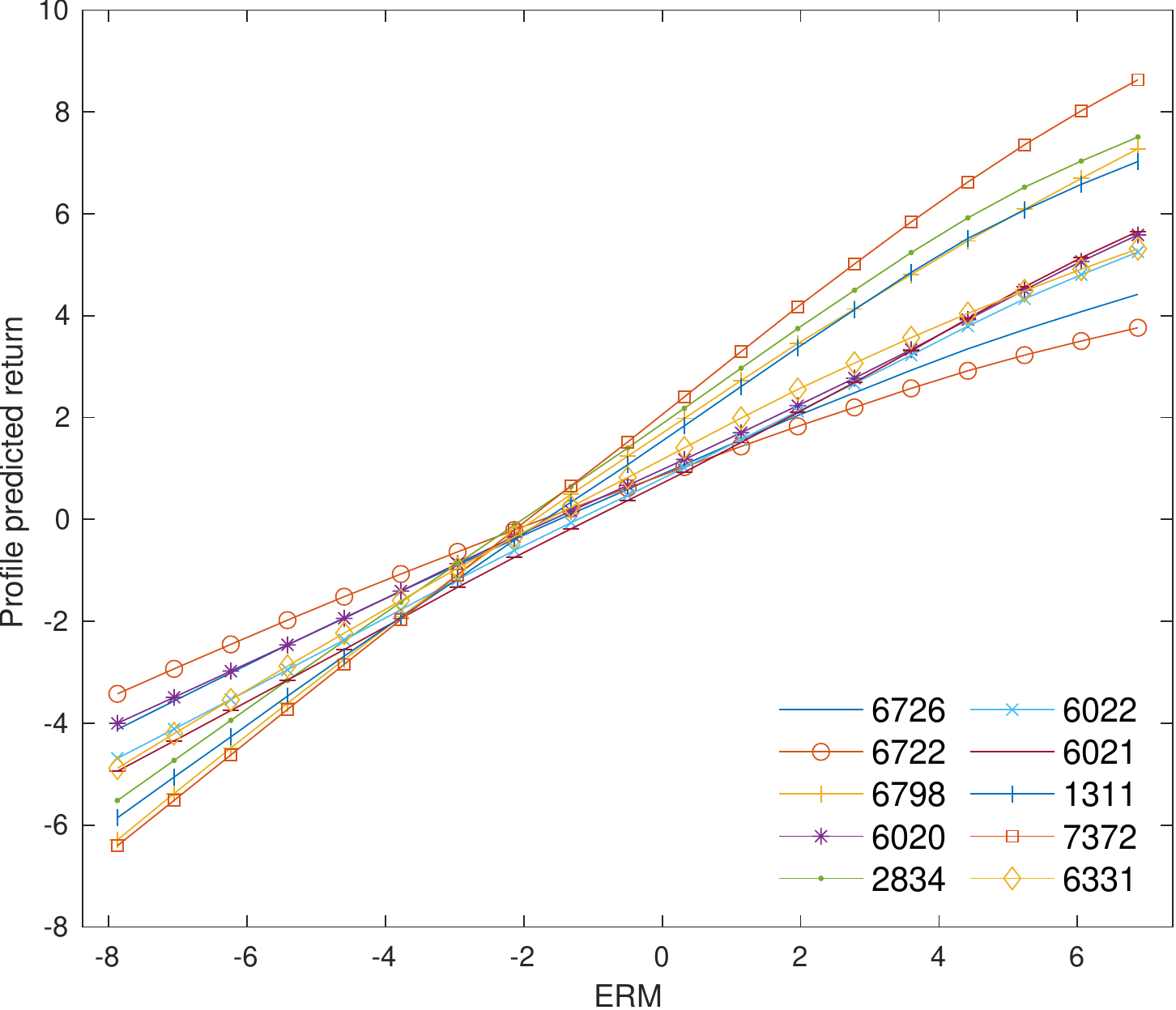}
		\label{ff3_lek_low_excess}
	\end{subfigure}
	\begin{subfigure}{0.33\linewidth}
		\subcaption{SMB (low)}		
		\includegraphics[width=1\linewidth]{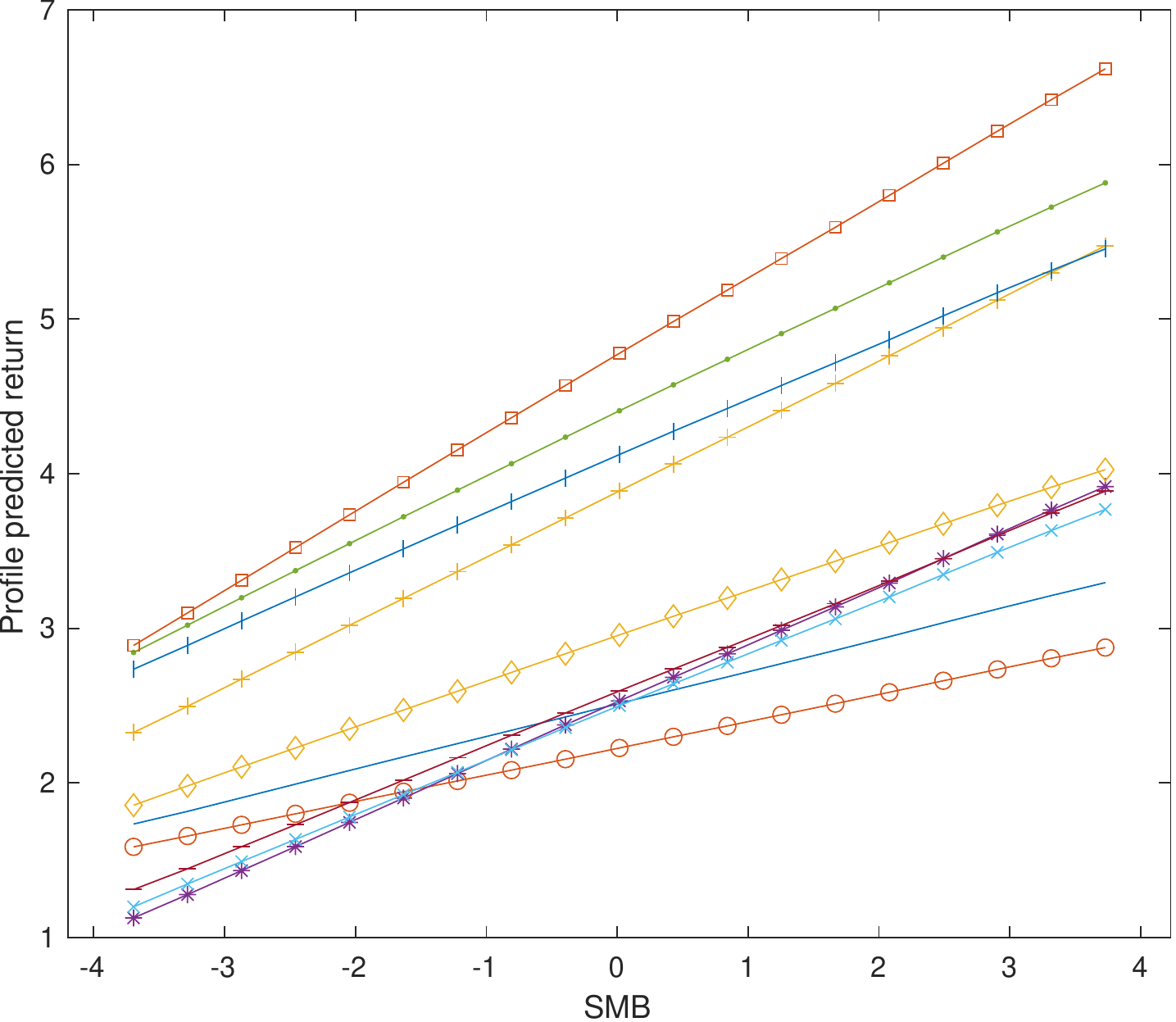}
		\label{ff3_lek_low_SMB}
	\end{subfigure}
	\begin{subfigure}{0.33\linewidth}
		\subcaption{HML (low)}		
		\includegraphics[width=1\linewidth]{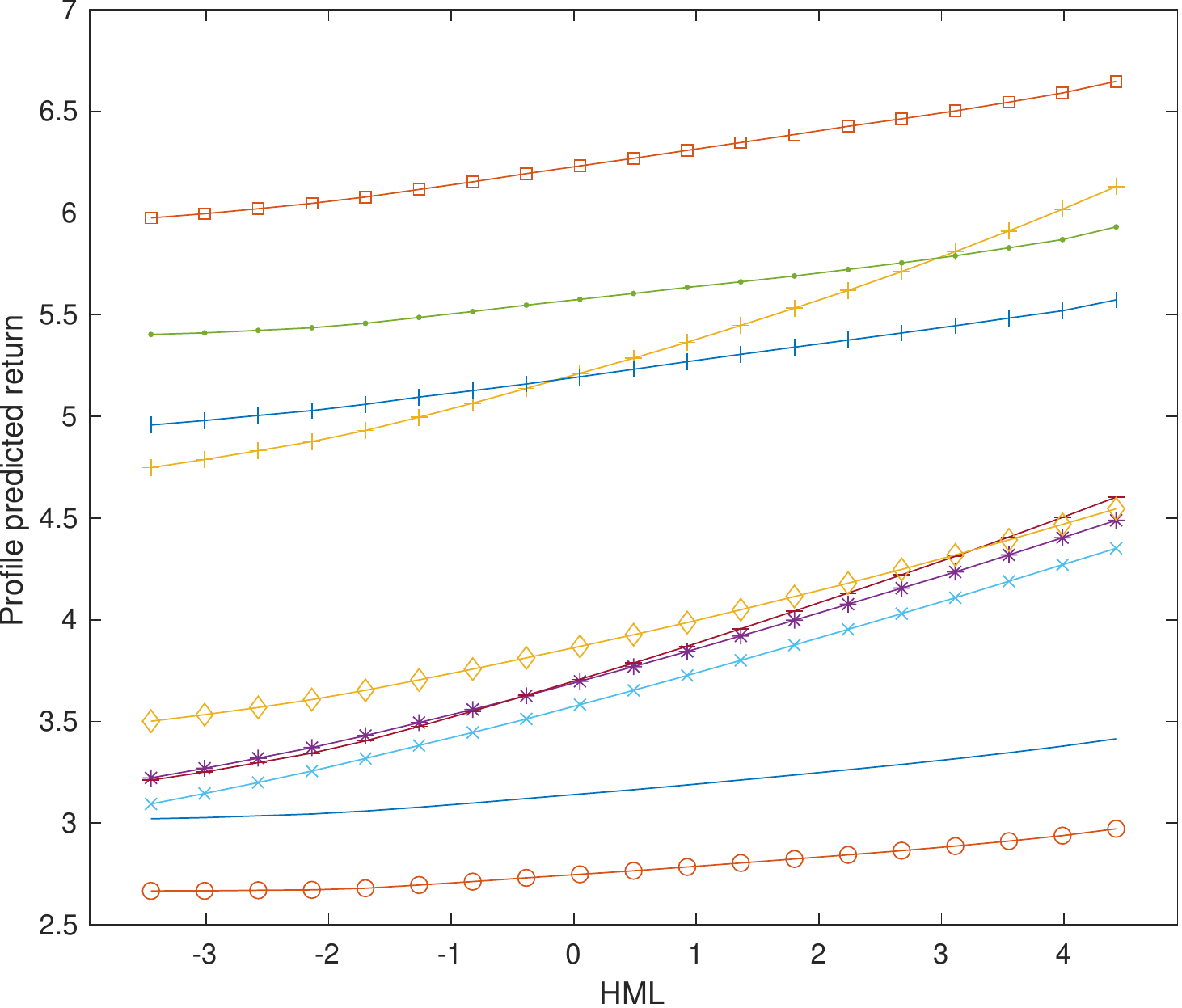}
		\label{ff3_lek_low_HML}
	\end{subfigure}
	\begin{subfigure}{0.33\linewidth}
		\subcaption{ERM (moderate)}
		\includegraphics[width=1\linewidth]{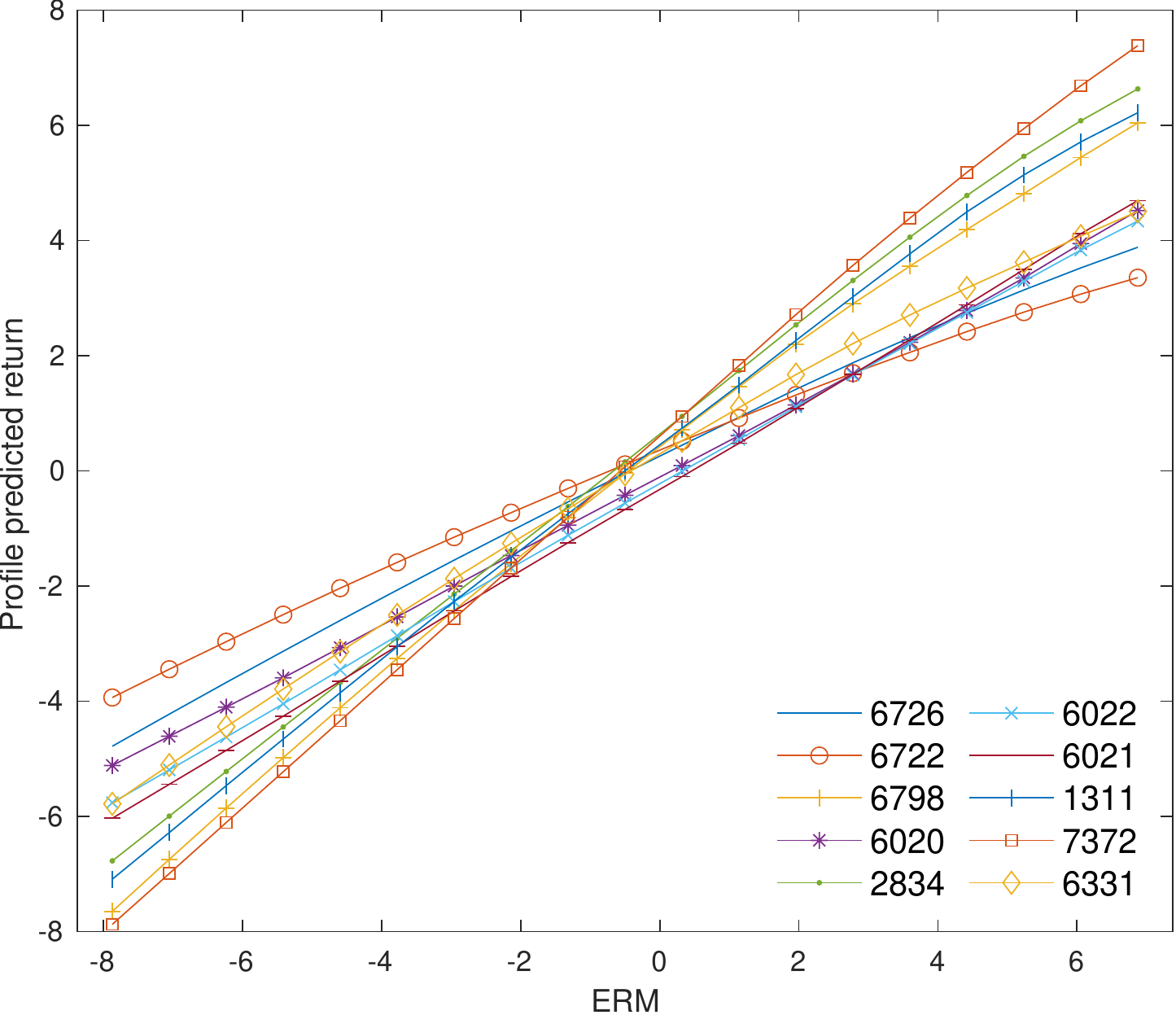}
	\end{subfigure}
	\begin{subfigure}{0.33\linewidth}
		\subcaption{SMB (moderate)}		
		\includegraphics[width=1\linewidth]{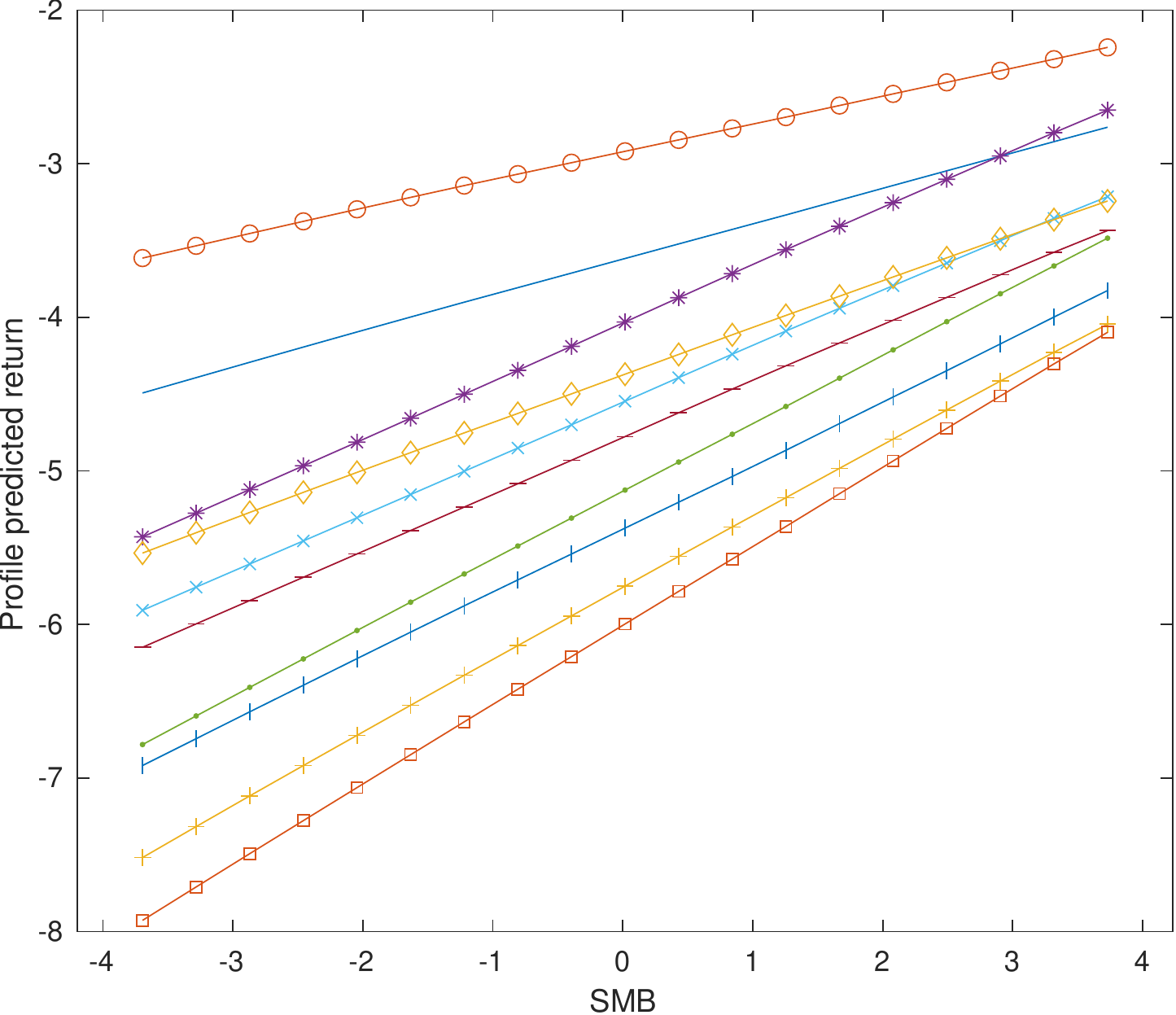}
	\end{subfigure}
	\begin{subfigure}{0.33\linewidth}
		\subcaption{HML (moderate)}		
		\includegraphics[width=1\linewidth]{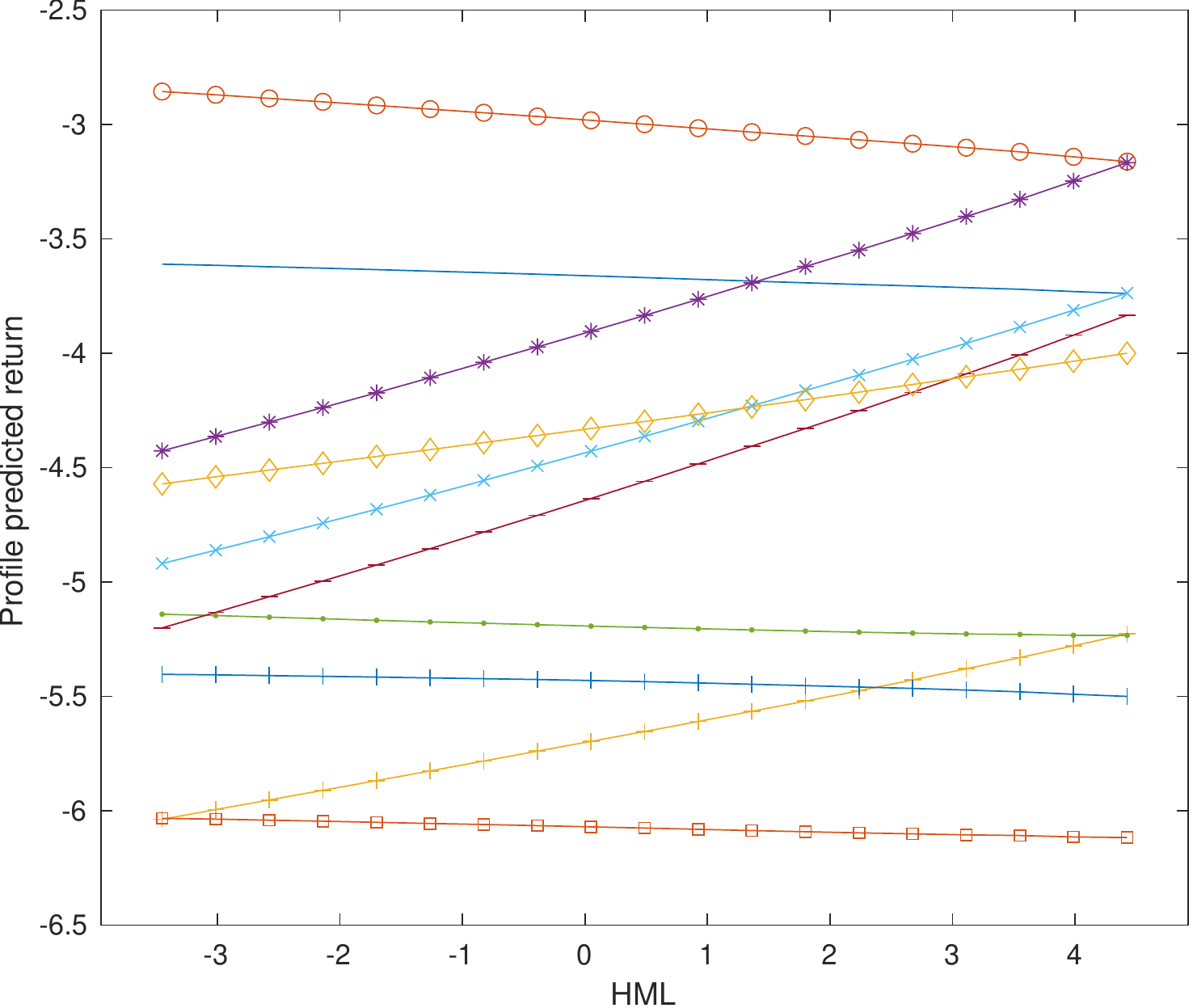}
	\end{subfigure}
	\begin{subfigure}{0.33\linewidth}
		\subcaption{ERM (extreme)}
		\includegraphics[width=1\linewidth]{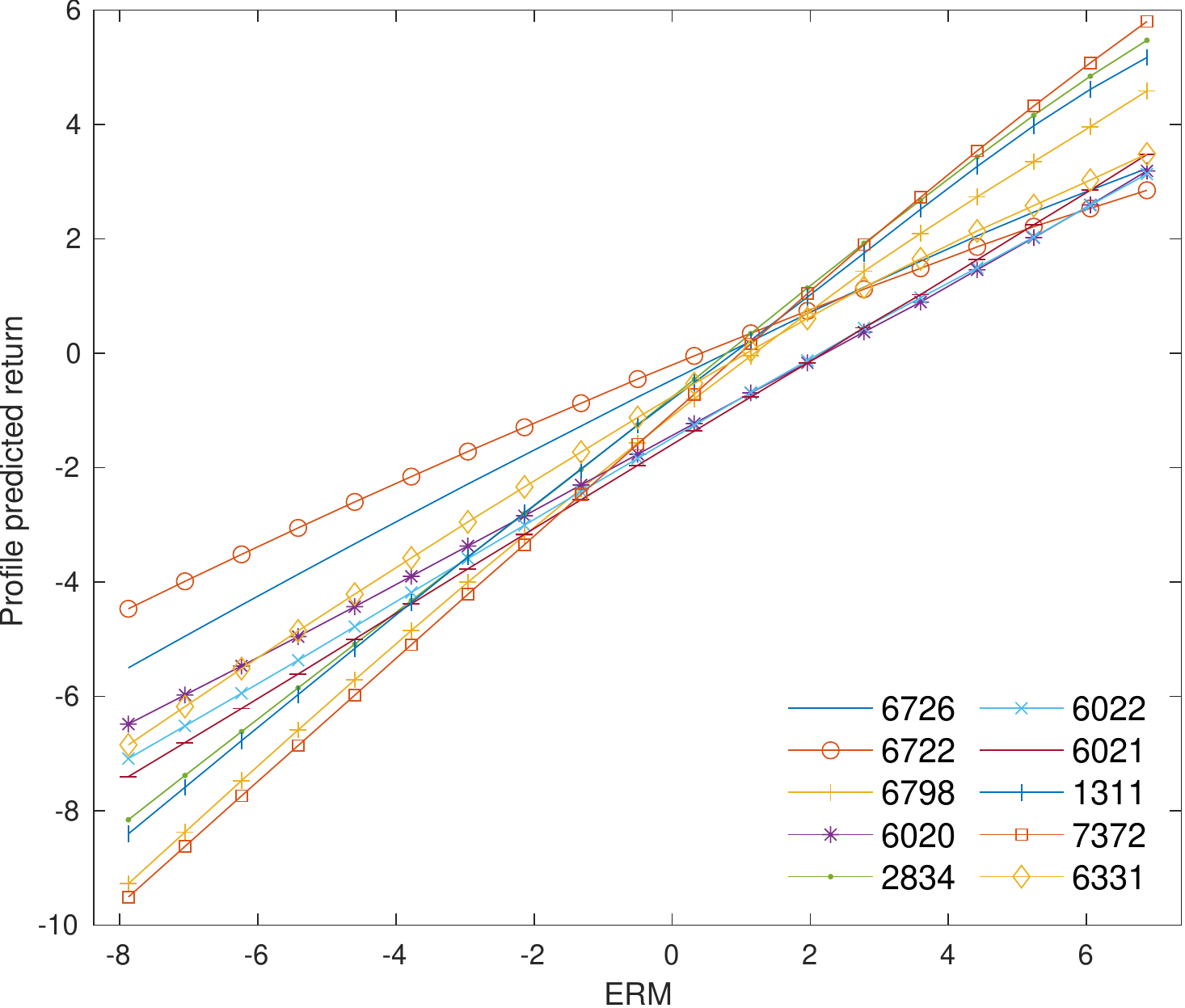}
	\end{subfigure}
	\begin{subfigure}{0.33\linewidth}
		\subcaption{SMB (extreme)}		
		\includegraphics[width=1\linewidth]{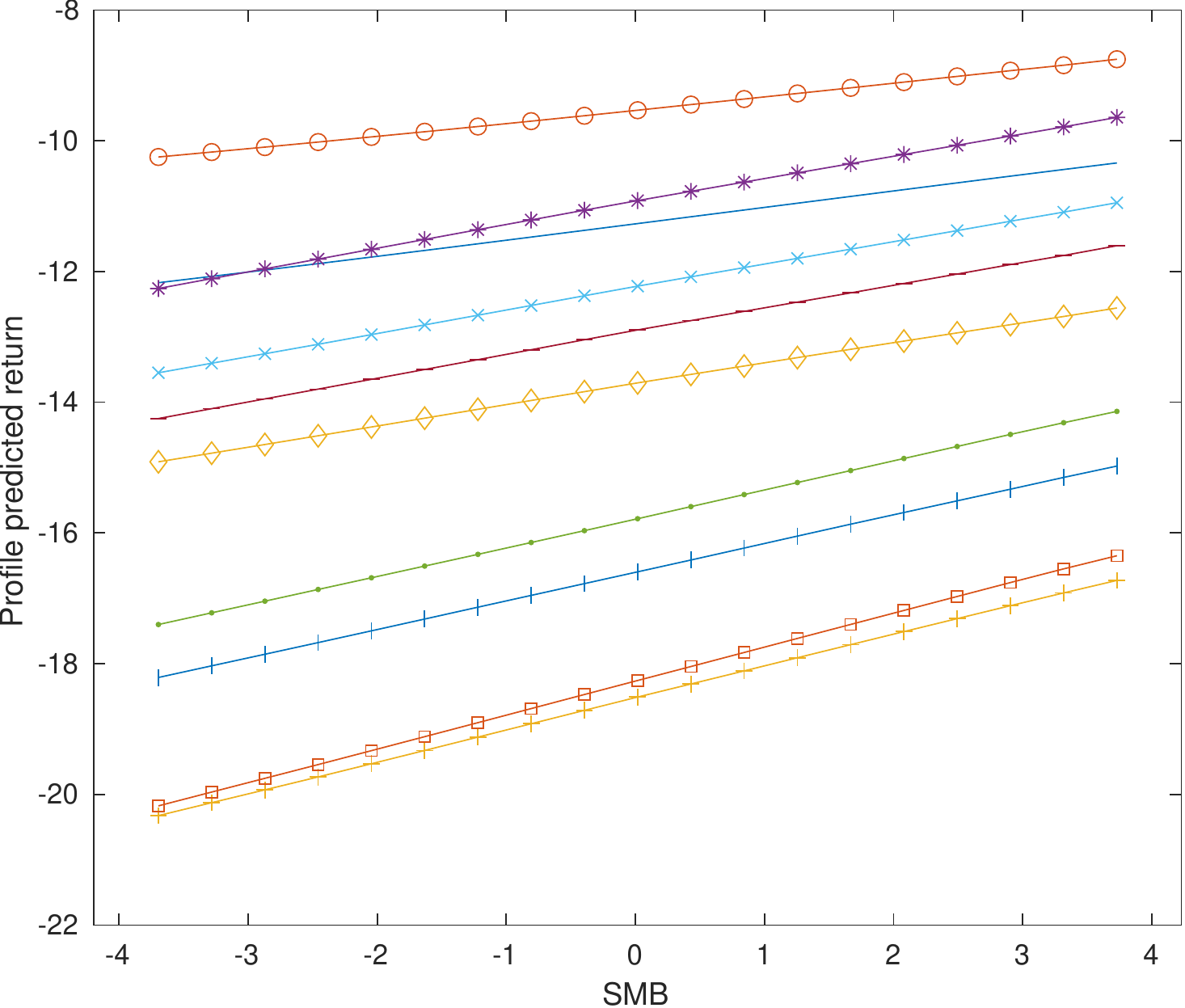}
	\end{subfigure}
	\begin{subfigure}{0.33\linewidth}
		\subcaption{HML (extreme)}		
		\includegraphics[width=1\linewidth]{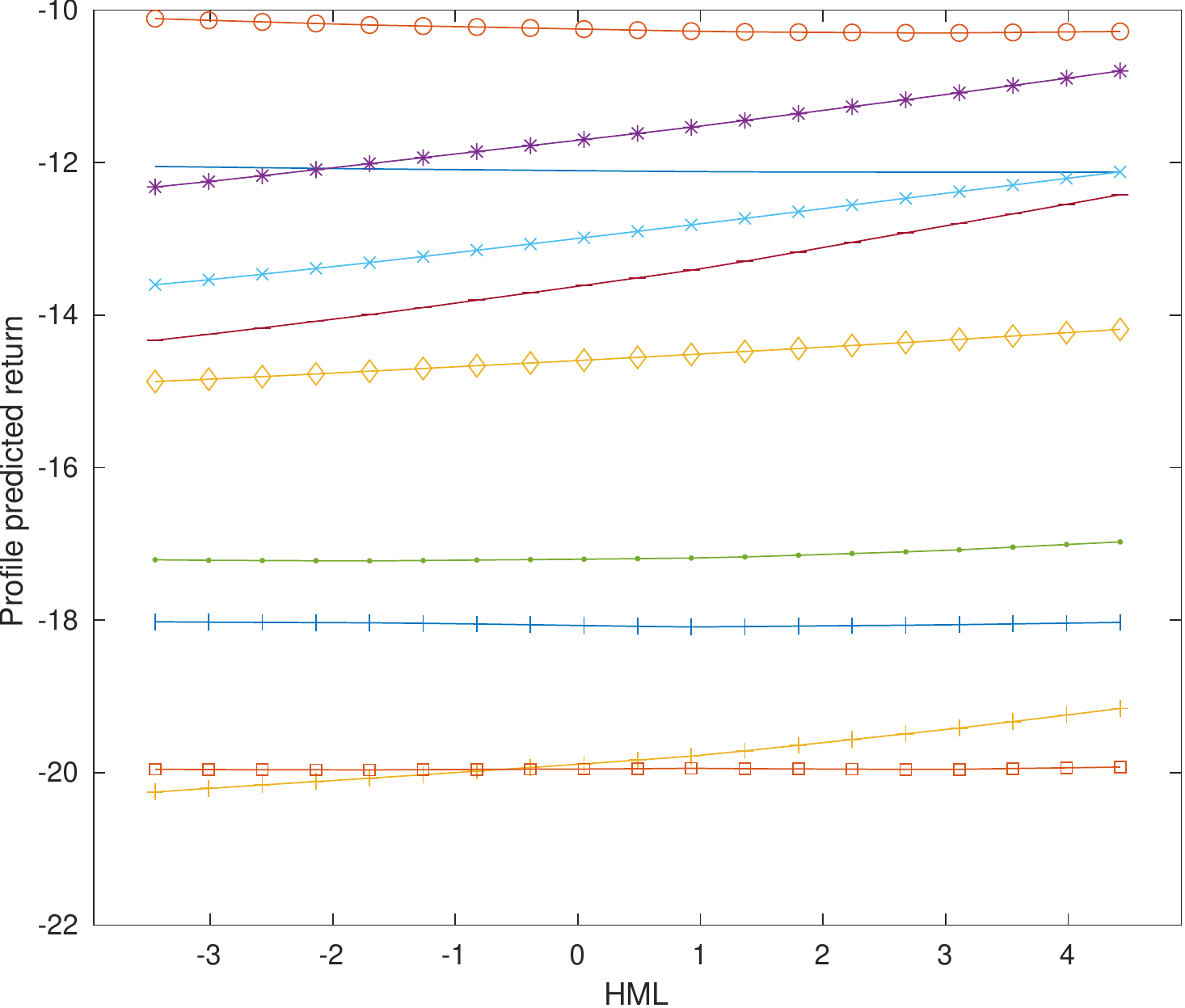}
	\end{subfigure}
	\caption{Lek profile depicting the heterogeneous responses implied by the DMM asset pricing model. The first row uses inputs on July 2005 (low market volatility month), the second row uses inputs on May 2012 (median market volatility month) and the last row inputs on October 2008 (extreme market volatility month).}
	\label{ff3_lek}
\end{figure}

For the best performing DMM model, Figure~\ref{ff3_lek}
plots Lek profiles~\citep{gevreyReviewComparisonMethods2003} of the FF3 factors  to visualize the
heterogeneous relationships for the ten most populated SIC codes given in
Table~\ref{sic_industry}. These depict how each factor affects expected stock
returns, while the remaining factors are fixed at their values on dates that
correspond to different market conditions: July 2005 (low market volatility), May 2012 (medium market volatility); and October 2008 (extreme market volatility).  
 
The profiles suggest non-linearity is not a prominent feature, but complex interactions between the factors are. There is substantial industry-based heterogeneity in the 
slopes. Market volatility does not seem to affect the slopes of ERM meaningfully, but does so for the other two factors. In the low market volatility case, HML has a negative slope for some industries, indicating that investors 
prefer high value rather than high growth stocks in these industries.  
Equivalent Lek profiles for the FF5 model are provided in Part E of the Web Appendix.

\section{Discussion}\label{sec:disc}
While natural gradient stochastic optimization methods have strong potential in 
variational inference, their use is limited due to the computational
complexity of implementation for choices of VAs outside the exponential family.  
In this paper, we propose a natural gradient method (NG-HVI) that employs the VA at~\eqref{eq:hybrid_q}, which is outside the exponential family.
Doing so has three advantages which we list here.
First,  Theorem~\ref{thm:FIM} and Corollary~\ref{cor:ng} show that the computational complexity of evaluating the natural gradient grows only with 
$\mbox{dim}(\thetavec)$, not the dimension of the target 
density $p(\psivec|\yvec)$. When $\mbox{dim}(\thetavec)<<\mbox{dim}(\psivec)$ (which is the case with most 
latent variable models) this reduces the computational complexity of evaluating 
the natural gradient greatly.  Second, the proposed VA may be far from Gaussian 
because the conditional posterior 
$p(\zvec|\thetavec,\yvec)$ may be also. This density, nor
its derivatives, are not required to implement 
Algorithm~\ref{alg:nghvi}. Instead, it is only necessary to draw from this conditional posterior,
which can be achieved using a wide array of existing methods. 
Last, that approximation at~\eqref{eq:hybrid_q} is necessarily more
accurate than any other VA for $\psivec$ that shares the same marginal $q^0(\thetavec)$.
In practice, the importance of this feature grows with the dimension of $\zvec$, because
the cumulative error of even well-calibrated fixed form VAs for $\zvec$ grows as well.
We demonstrate these advantages in our empirical work. The results suggest
that learning the
VA using SNGA proves  more efficient and stable than using SGA as originally 
suggested by~\cite{loaiza-mayaFastAccurateVariational2022}. The proposed
method also out-performed the two natural gradient benchmark methods for 
our examples. While these examples are all latent variable models, NG-HVI can
also be used with 
other stochastic models where $\psivec$ is partitioned in other ways, as long as generation of $\zvec$
from its conditional posterior is
viable. For example, $\zvec$ may contain discrete-valued 
parameters, which is a case that is difficult to deal with using standard VI methods~\citep{tran2019discrete,jiMarginalizedStochasticNatural2021}.

\begin{sidewaystable}[htbp]
	\begin{center}
		\begin{threeparttable}
			\caption{Summary of the mixed effect examples}\label{tab:egsummary}
			{\small
				\begin{tabular}{ccccccccccccc}
					\toprule
					& &\multicolumn{5}{c}{Example Features} &&\multicolumn{4}{c}{NG-HVI Details} &\\ \cline{3-7} \cline{9-12}
					& \makecell{Model\\Specification} & $\mbox{dim}(\alphavec_k)$& $\mbox{dim}(\thetavec)$& $\mbox{dim}(\zvec)$&$K$&\makecell{$n$} &&$p$& $\mbox{dim}(\lambdavec)$& \makecell{Time/\\step (s)}&\makecell{Time to\\fit (min)} &Benchmark\\
					\hline
					\multicolumn{10}{l}{Simulated Examples}\\ \cline{1-2}
					E.g. 1	  & \makecell{Linear} & 1&8&1000&1000&5000&&0 to 3&16 to 40&0.01&0.3&\makecell{DAVI,\\ SG-HVI}\\
					E.g. 2(a)& \makecell{Gaussian\\DMM [5,5]} & 6&88&6000&1000&6000&&3&440&0.05&8.3 &\makecell{DAVI,\\SG-HVI}\\
					E.g. 2(b)&  \makecell{Gaussian\\DMM [32,16]}  & 17&2779&17000&1000&30000&&3&13895&16.5&133.2 &SG-HVI\\
					E.g. 3	  &  \makecell{Bernoulli\\DMM [5,5]}  & 6&72&20000&1000&14000&&1&216&0.48&24.9 &\makecell{NAGVAC,\\SG-HVI}\\
					\hline
					\multicolumn{10}{l}{Financial Application}\\ \cline{1-2}
					FF3&  \makecell{Gaussian\\DMM [8]}       & 9&87&4932&548&309960  &&3&435& 0.11 & 5.6 &Various\\
					FF3&  \makecell{Gaussian\\DMM [32,16,8]} & 9  &  847&4932&548 &309960&&5&5929&0.89&44.7 &Various\\
					FF5&  \makecell{Gaussian\\DMM [8]} 		 & 9&103&4932&548&309960  && 3&515&0.12 & 6.1 &Various\\
					FF5&  \makecell{Gaussian\\DMM [32,16,8]} & 9  &  911&4932&548 &309960&&5&6377&1.34&67.3 &Various\\
					\bottomrule
				\end{tabular}
			}	
		\end{threeparttable}	
	\end{center}
	Note: the example features include the dimension of the random coefficient vector $\alphavec_k$, global parameters $\thetavec$, latent variables $\zvec$, number of groups $K$, and training data sample size $n$. Details of the NG-HVI method include the number of factors $p$ for the covariance matrix of the VA, dimension of the variational parameters $\lambdavec$, and computation times based on 3000 steps using a 2022 HP desktop with Intel i9 processor. The final column indicates the benchmark or comparisons considered.  
\end{sidewaystable}

Our examples provide extensive empirical evidence that NG-HVI works well. Table~\ref{tab:egsummary} provides
a summary of these examples, along with details on the application of NG-HVI for their estimation. We focus on DMMs
that employ 
random coefficients on the output layer of a probabilistic Bayesian 
neural network because they have strong applied potential. 
They allow for heterogeneity, which is important in many studies, and capturing it
using random coefficients  in other stochastic models is a well-established approach~\citep{gelman2006}.
However, DMMs are hard to train because
the posterior is complex and
the dimension of $\zvec$ is often high. Our empirical work demonstrates
that NG-HVI is an effective approach for doing so, and that it improves substantially on SG-HVI.
One alternative is to use importance sampling to integrate out the random 
coefficients, as suggested in the 
original study of~\cite{tranBayesianDeepNet2020}. However, this is more computationally 
demanding than NG-HVI, and the method scales poorly with the number of
output layer nodes. The potential of DMMs is demonstrated in
the financial study, where
the combination of a feed forward neural network and industry-based
heterogeneity produces a significant
increase in predictive accuracy using the Fama and French factors as inputs. 

We finish by noting two directions for further work. First, while we employ
the Gaussian factor covariance VA for $q^0$, 
other VAs for which fast natural gradient updates can be computed 
may also be used. This includes a Gaussian VA
with fast Cholesky updates as suggested by \cite{tanAnalyticNaturalGradient2022},
the 
implicit copula VAs of~\cite{han2016,smithHighDimensionalCopulaVariational2020}, the 
mixture VAs of~\cite{linFastSimpleNaturalGradient2019} or the tractable
fixed forms 
suggested by~\cite{linTractableStructuredNaturalgradient2021}. 
Second, the promising results of the financial study encourage the further exploration of the 
use of DMMs in asset pricing. 

\appendix
\setcounter{table}{0}
\setcounter{figure}{0}
\setcounter{algorithm}{0}
\renewcommand{\thetable}{\Alph{section}\arabic{table}}
\renewcommand{\thefigure}{\Alph{section}\arabic{figure}}
\renewcommand{\thealgorithm}{\Alph{section}\arabic{algorithm}}
\section{Proof of Theorem~\ref{thm:FIM}}\label{App:NatGrad}
Observe that because $q_\lambda(\bm\psi)=
q^0_\lambda(\bm\theta)p(\bm{z}|\bm{\theta},\bm{y})$, where $p(\bm{z}|\bm{\theta},\bm{y})$ is not a function of $\lambdavec$, then
\begin{align*}
	F(\bm\lambda) &= E_{q_\lambda}\left[\nabla_\lambda\log q_\lambda(\bm\psi)\nabla_\lambda\log q_\lambda(\bm\psi)^\top\right]\\
	 &= E_{q_\lambda}\left[\left(\nabla_\lambda\log q^0_\lambda(\bm\theta)+\nabla_\lambda\log p(\bm{z}|\bm{\theta},\bm{y})\right)\left(\nabla_\lambda\log q^0_\lambda(\bm\theta)+\nabla_\lambda\log p(\bm{z}|\bm{\theta},\bm{y})\right)^\top\right]\\
	 &= E_{q_\lambda}\left[\nabla_\lambda\log q^0_\lambda(\bm\theta)\nabla_\lambda\log q^0_\lambda(\bm\theta)^\top\right]\\
	&= E_{q_\lambda^0}\left[E_{p(\bm z|\bm{\theta},\bm{y})}\left[\nabla_\lambda\log q^0_\lambda(\bm\theta)\nabla_\lambda\log q^0_\lambda(\bm\theta)^\top\right]\right]\\
		&= E_{q_\lambda^0}\left[\nabla_\lambda\log q^0_\lambda(\bm\theta)\nabla_\lambda\log q^0_\lambda(\bm\theta)^\top\right]= F^0(\bm\lambda)
\end{align*}
which proves the result.

\section{Matrix inverse}\label{app:fim}
In Section~\ref{sec:ffapprox}, application of the Woodbury formula and standard matrix identities provides the expression
\begin{eqnarray*}
	\widetilde{F}_{11}(\lambdavec)^{-1} & = & -H^{-1} + H^{-1}G(I+G^\top H^{-1}G)^{-1}G^\top H ^{-1}
\end{eqnarray*}
where $H  = \widetilde{E} - (1 + \delta) D^{-2}$, 
$G  =  D^{-2}B \widetilde{C}$, 
$\widetilde{E}  =  -D^{-2}\mbox{diag}(\sum_{i = 1}^{p}\evec_i\circ \bvec_i)D^{-2}$,
$E = B C$, 
$C  =  \widetilde{C}\widetilde{C}^\top$, and
$\widetilde{C} = (I + B^\top D^{-2}B)^{-1}$. The vectors
$\evec_i$ and $\bvec_i$ denote the $i^{th}$ column of matrices $E$ and $B$, respectively. Because $H$ is a diagonal matrix, no $m\times m$ matrices need
to be stored to compute $\widetilde{F}_{11}(\lambdavec)^{-1}\widehat{\nabla_\mu{\cal L}(\lambdavec)}$.

\section{Prior and gradient for $\lvec$}\label{app:omega}
In Section~\ref{sec:nghvidmm}, the precision matrix $\Omega_\alpha^{-1}=LL^\top$ is re-parameterized to $\lvec=\mbox{vech}^\star(L)$, where the operator `$\mbox{vech}^\star$' is 
the half-vectorization of the Cholesky factor $L$, but where the logarithm
is taken of the diagonal elements. The Jacobian of this transformation 
is easy to calculate (e.g. see Theorem 4 in~\cite{deemer1951jacobians}), 
resulting in the prior density
$p(\lvec)\propto p(\Omega_\alpha^{-1})\prod_{i=1}^{m_L}l_{ii}^{(m_L-i+2)}$,
where $p(\Omega_\alpha^{-1})$ is the Wishart prior for the precision matrix. 

The derivative  $\nabla_l \log g(\thetavec,\zvec) = \sum_{k=1}^K \nabla_l \log p(\yvec_k,\alphavec_k|\thetavec) +\nabla_l \log p(\thetavec)$. Using results from \cite{tanUseModelReparametrization2021},
\begin{align*}
	\nabla_{l} \log p(\yvec_k, \alphavec_k|\thetavec) & =D^L\mbox{vech}(L^{-T} - \alphavec_k\alphavec_k^\top L)\\
	\nabla_{l}\log p(\thetavec)& = D^L\mbox{vech}\left((\nu - m_L- 1)L^{-T} - S^{-1}L\right) + \mbox{vech}(\mbox{diag}(\uvec))\,, 
\end{align*}
where $D^L$ is a diagonal matrix of order $m_L(m_L+1)/2$ with diagonal given by $vech(J^L)$, and $J^L$ is an $ m_L\times m_L$ matrix with the
same leading diagonal as $L$, but unity off-diagonal elements.
The vector
$\uvec$ is of length $m_L$ with $ith$ element equal to $m_L-i+2$.

\singlespacing
\newpage
\bibliography{lit_list}
\FloatBarrier

\newpage
\onehalfspacing
\newpage
\noindent
\setcounter{page}{1}
\begin{center}
	{\bf \Large{Online Appendix for ``Natural Gradient Hybrid Varitaional Inference with Application to Deep Mixed Models''}}
\end{center}

\vspace{10pt}

\setcounter{figure}{0}
\setcounter{table}{0}
\setcounter{section}{0}
\setcounter{algorithm}{1}
\renewcommand{\thetable}{A\arabic{table}}
\renewcommand{\thefigure}{A\arabic{figure}}
\renewcommand{\thealgorithm}{\Alph{section}\arabic{algorithm}}
\noindent
This Online Appendix has five parts:

\begin{itemize}
	\item[] {\bf Part~A}: Notational conventions and matrix differentiation rules used.
	\item[] {\bf Part~B}: Additional details on the efficient evaluation of the natural gradient.
	\item[] {\bf Part~C}: Additional details and empirical results for Section~3, including an additional illustrative example of a probit model.
	\item[] {\bf Part~D}: Additional details for the examples in Section~4.
	\item[] {\bf Part~E}: Additional results for the financial example in Section~5.
\end{itemize}
\newpage

\noindent {\bf \large{Part~A: Notational conventions and matrix differentiation rules used}}\\
\ \\
\noindent 
We outline the notational conventions that we adopt in computing
derivatives throughout the paper, which are the same as adopted in~\cite{loaiza-mayaFastAccurateVariational2022}. For a $d$-dimensional vector valued function $g(\bm x)$ of an $n$-dimensional
argument $\bm x$, $\frac{\partial g}{\partial \bm x}$ is the $d\times n$ matrix with element $(i,j)$ $\frac{\partial g_i}{\partial x_j}$.  This means for a scalar $g(\bm x)$, $\frac{\partial g}{\partial \bm x}$ is
a row vector.  When discussing the SGA algorithm we also sometimes write $\nabla_x g(\bm x)=\frac{\partial g}{\partial \bm x}^\top$, which is a column vector.
When the function $g(\bm x)$ or the argument $\bm x$ are matrix valued, then $\frac{\partial g}{\partial \bm x}$ is taken to 
mean $\frac{\partial \text{vec}(g(\bm x))}{\partial \text{vec}(\bm x)}$, where $\text{vec}(A)$ denotes the vectorization of a matrix $A$ obtained by stacking its columns one
underneath another.  If $g(x)$ and $h(x)$ are matrix valued functions, say $g(x)$ takes values which are $d\times r$ and $h(x)$ takes values which are $r\times n$, 
then a matrix valued product rule is
\begin{align*}
	\frac{\partial g(x)h(x)}{\partial x} & = (h(x)^\top\otimes I_d)\frac{\partial g(x)}{\partial x}+(I_n\otimes g(x))\frac{ \partial h(x)}{\partial x}
\end{align*}
where $\otimes$ denotes the Kronecker product and $I_a$ denotes the $a\times a$ identity matrix for a positive integer $a$.  

Some other useful results used repeatedly throughout the derivations below are
$$\text{vec}(ABC)=(C^\top\otimes A)\text{vec}(B),$$
for conformable matrices $A$, $B$ and $C$
the derivative 
\begin{align*}
	\frac{\partial A^{-1}}{\partial A} & = -(A^{-\top}\otimes A^{-1}).
\end{align*}
We also write $K_{m,n}$ for the commutation matrix (see, for example, Magnus and Neudecker, 1999).

Last, for scalar function $g(x)$ of scalar-valued argument $x$, we 
sometimes write $g^{\prime}(x)=\frac{d}{d x}g(x)$ and $g^{\prime\prime}(x)
=\frac{d^2}{d x^2}g(x)$ for the first and second derivatives with respect
to $x$ whenever it appears clearer to do so.
\newpage

\noindent {\bf \large{Part~B: Additional details on the efficient evaluation of the damped natural gradient}}\\
\ \\

\noindent
Here we provide further details on the efficient evaluation of
the damped natural gradient
\[\widetilde{\nabla}_\lambda\mathcal{L}(\bm{\lambda}) \equiv \widetilde{F}^0(\bm\lambda)^{-1}\widehat{\nabla_\lambda\mathcal{L}(\bm{\lambda})}\,,
\]
for the VA at~(1) with the Gaussian VA with factor covariance matrix for $q^0_\lambda$ given in Section 3.4. 

First, restating the closed form expressions given in~\cite{ongGaussianVariationalApproximation2018} for the re-parameterization gradient at~(7) gives
\begin{eqnarray*}
	\nabla_\mu{\cal L}(\lambdavec) &= &E_{f_\varepsilon}\left[\nabla_\theta \log g(\thetavec,\zvec)
	+(BB^\top + D^2)^{-1}(B\varepsilonvec_1^0+\dvec\circ \varepsilonvec_2^0)\right]\\
	\nabla_{\rm{vech}(B)}{\cal L}(\lambdavec) &= &E_{f_\varepsilon}\left[\nabla_\theta \log g(\thetavec,\zvec)(\varepsilonvec_1^0)^\top
	+(BB^\top + D^2)^{-1}(B\varepsilonvec_1^0+\dvec\circ \varepsilonvec_2^0)(\varepsilonvec_1^0)^\top\right]\\
		\nabla_{d}{\cal L}(\lambdavec) &= &E_{f_\varepsilon}\left[\mbox{diag}\left(\nabla_\theta \log g(\thetavec,\zvec)(\varepsilonvec_2^0)^\top
	+(BB^\top + D^2)^{-1}(B\varepsilonvec_1^0+\dvec\circ \varepsilonvec_2^0)(\varepsilonvec_2^0)^\top\right)\right]\,,
\end{eqnarray*}
where ${\rm diag}(A)$ is the vector of diagonal elements of the square matrix $A$. 
The expectations are evaluated at a single draw from the density
\[\varepsilonvec=\left((\varepsilonvec^0_1)^\top,(\varepsilonvec^0_2)^\top,\zvec^\top\right)^\top\sim 
f_\varepsilonvec(\varepsilonvec^0,\zvec)=f_{\varepsilonvec^0}(\varepsilonvec^0)p(\zvec|h(\varepsilonvec^0,\lambdavec),\yvec)\,,
\]
with $\varepsilonvec^0=((\varepsilonvec_1^0)^\top,(\varepsilonvec_2^0)^\top)^\top$,
the transformation $h(\varepsilonvec^0,\lambdavec)=\muvec+B\varepsilonvec_1^0+\dvec\circ \varepsilonvec_2^0$ and $f_{\varepsilon_0}$ is the density of a $N(\bm{0},I_{m+p})$ distribution. 
Generating from $p(\zvec|h(\varepsilonvec^0,\lambdavec),\yvec)$ is undertaken
either directly using a Monte Carlo method, or approximately using a few steps 
of an MCMC scheme initialized at the draw of $\zvec$ obtained at the previous step of the NGA algorithm.
In the computations above, the $m\times m$ matrix $(BB^\top+D^2)^{-1}$ is not evaluated or stored directly, but is replaced by
the Woodbury formula $(BB^\top+D^2)^{-1}=D^{-2}-D^{-2}B(I+B^\top D^{-2}B)^{-1}B^\top D^{-2}$ in the equations. Multiplying through enables efficient evaluation.

Second, to evaluate the damped FIM, we first note that for the Gaussian factor
covariance VA for $q^0_\lambda$, the FIM $F^0$ is given by the patterned matrix at~(8).  \cite{ongLikelihoodfreeInferenceHigh2018} give the following closed form expressions for the block
matrices:
\begin{eqnarray*}
	F_{11}(\lambdavec) & = & (BB^\top+D^2)^{-1}\\
	F_{22}(\lambdavec) & = & 2(B^\top \Sigma^{-1} B \otimes \Sigma^{-1})\\
	F_{33}(\lambdavec) & = & 2(D\Sigma^{-1}\circ \Sigma^{-1}D)\\
	F_{32}(\lambdavec) & = & 2(B^\top \Sigma^{-1}D \otimes \Sigma^{-1})E_m^\top
\end{eqnarray*}
where $E_m$ is an $m \times m^2$ permutation matrix such that for an $m \times m$ matrix $A$, $E_m \mbox{vec}(A) = \mbox{diag}(A)$.

The damped FIM is equal to~(8), but with the leading diagonal blocks replaced by 
$\widetilde{F}_{jj}(\lambdavec) = F_{jj}(\lambdavec) + \delta \mbox{diag}\left(F_{jj}(\lambdavec)\right)$ for $j=1,2,3$. The 
elements
$\widetilde{F}_{11}(\lambdavec)^{-1}\widehat{\nabla_\mu{\cal L}(\lambdavec)}$ are computed using the analytical
expression for $\widetilde{F}_{11}(\lambdavec)^{-1}$ given in Appendix~B
of the paper.
The remaining elements
of the damped natural gradient are
obtained by using a pre-conditioned conjugate gradient solver (such as ``{\tt pcg}'' in MATLAB) to solve the system
\[
\left[\begin{array}{cc}
\widetilde{F}_{22}(\lambdavec) &F_{32}(\lambdavec)^\top \\
F_{23}(\lambdavec) &\widetilde{F}_{33}(\lambdavec) 
\end{array}
\right]^{-1}
\left(\begin{array}{c}
\widehat{\nabla_{\rm{vech}(B)}{\cal L}(\lambdavec)}\\
\widehat{\nabla_{d}{\cal L}(\lambdavec)} 
\end{array}
\right)\,.
\]
\newpage

\noindent {\bf \large{Part~C: Additional Details for the Simulation in Section~3}}\\
\ \\
\noindent  {\bf {C.1: Linear random effects model}}\\
In the random effects model we set the fixed effect vector $\betavec=(0.8292,-1.3250,0.9909,1.6823,-1.7564,0.0580)$, and the 
covariate vector $\xvec_i$ consists of an intercept and
five values generated from a
correlated Gaussian. The error variance $\sigma^2_\epsilon=1$ throughout,
and the random effect variance $\sigma^2_\alpha$ is varied as outlined
in Section~3.

To generate the latent variables in the HVI methods, 
it is straightforward to show that $p(\alphavec|\thetavec,\yvec)=\prod_{k=1}^K p(\alpha_k|\thetavec,\yvec_k)$,
where $\alpha_k|\thetavec,\yvec_k\sim N(\mu_{\alpha_k},\sigma^2_{\alpha_k})$ with
\begin{align*}
\mu_{\alpha_k} & = \sigma^2_{\alpha_k} \frac{\iotavec_k^\top(\yvec_k - X_k\betavec)}{\sigma^2_e}\\
\sigma^2_{\alpha_k}  & = \left(\frac{1}{\sigma^2_\alpha} + \frac{\iotavec_k^\top\iotavec_k}{\sigma^2_e}\right)^{-1}
\end{align*}
Here $X_k$ and $\yvec_k$ are the covariate matrix and response variable corresponding to group $k$, respectively. The vector $\iotavec_k$ comprises
unity elements and is the same length as $\yvec_k$. 

\begin{figure}[H]
	\centering
	\begin{subfigure}{0.327\linewidth}
		\subcaption{posterior of $\beta_1$}
		\includegraphics[width=1\linewidth]{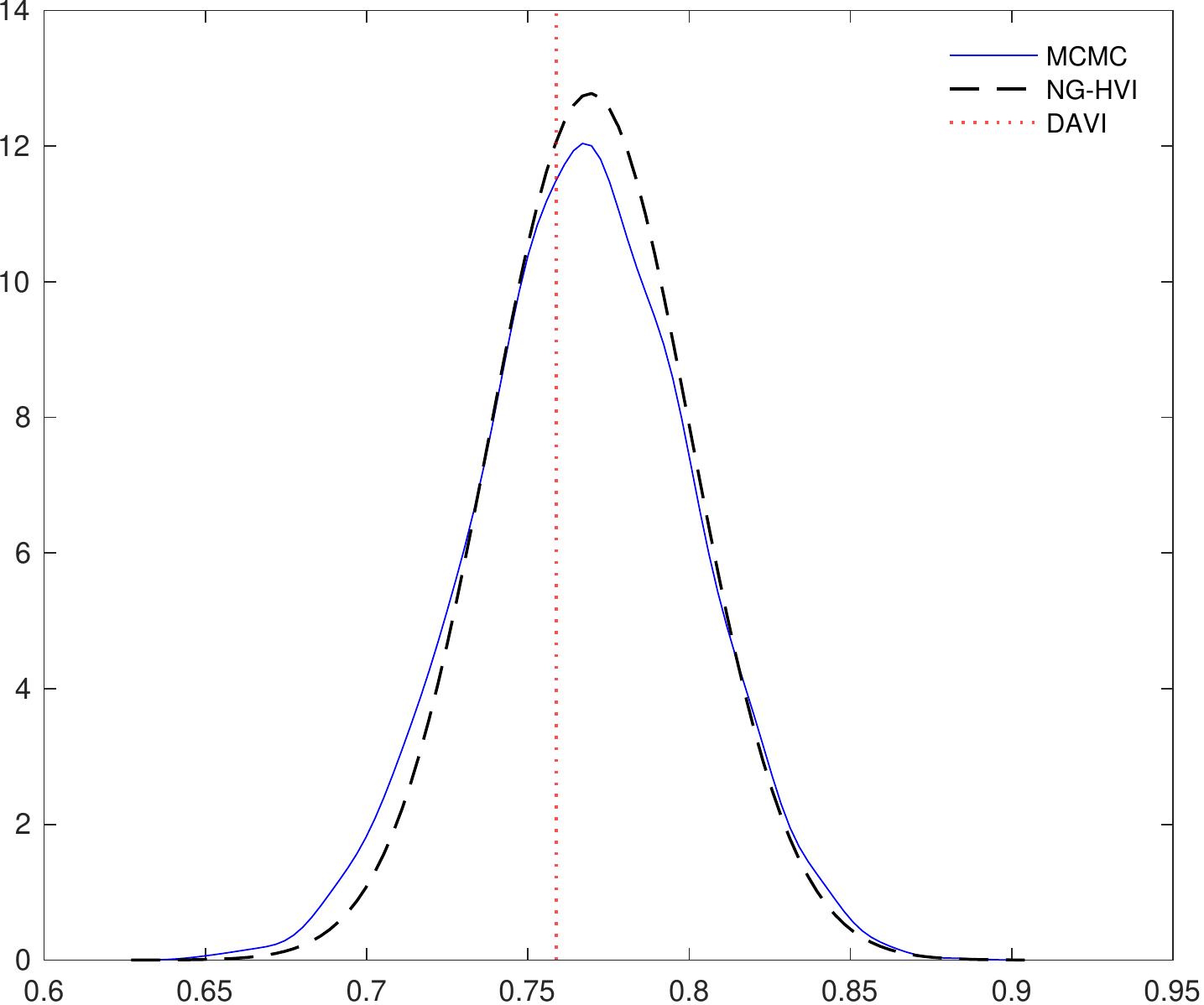}
	\end{subfigure}
	\begin{subfigure}{0.33\linewidth}
		\subcaption{posterior of $\beta_2$}
		\includegraphics[width=1\linewidth]{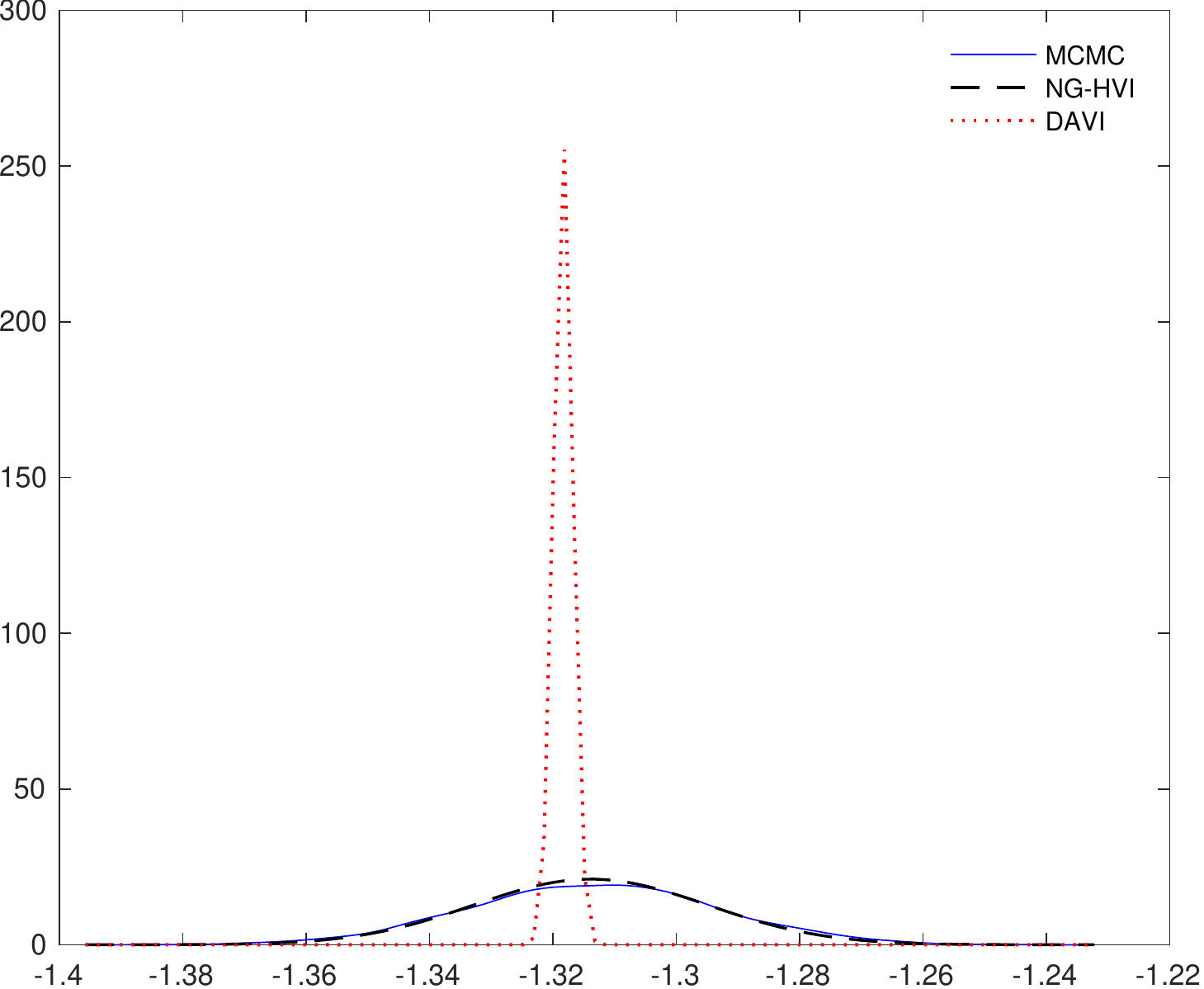}
	\end{subfigure}
	\begin{subfigure}{0.33\linewidth}
		\subcaption{posterior of $\beta_3$}
		\includegraphics[width=1\linewidth]{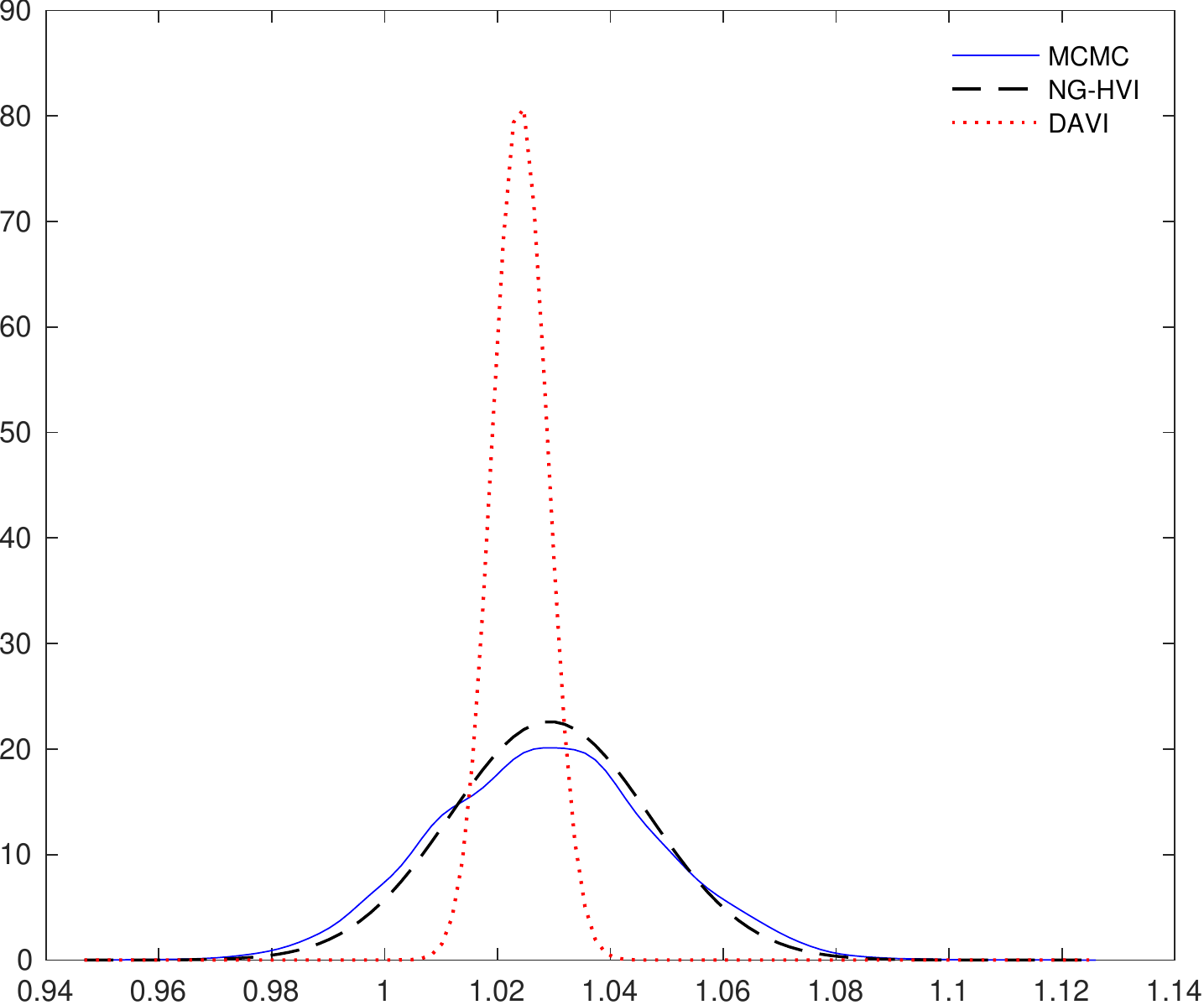}
	\end{subfigure}
	\begin{subfigure}{0.33\linewidth}
		\subcaption{posterior of $\beta_4$}
		\includegraphics[width=1\linewidth]{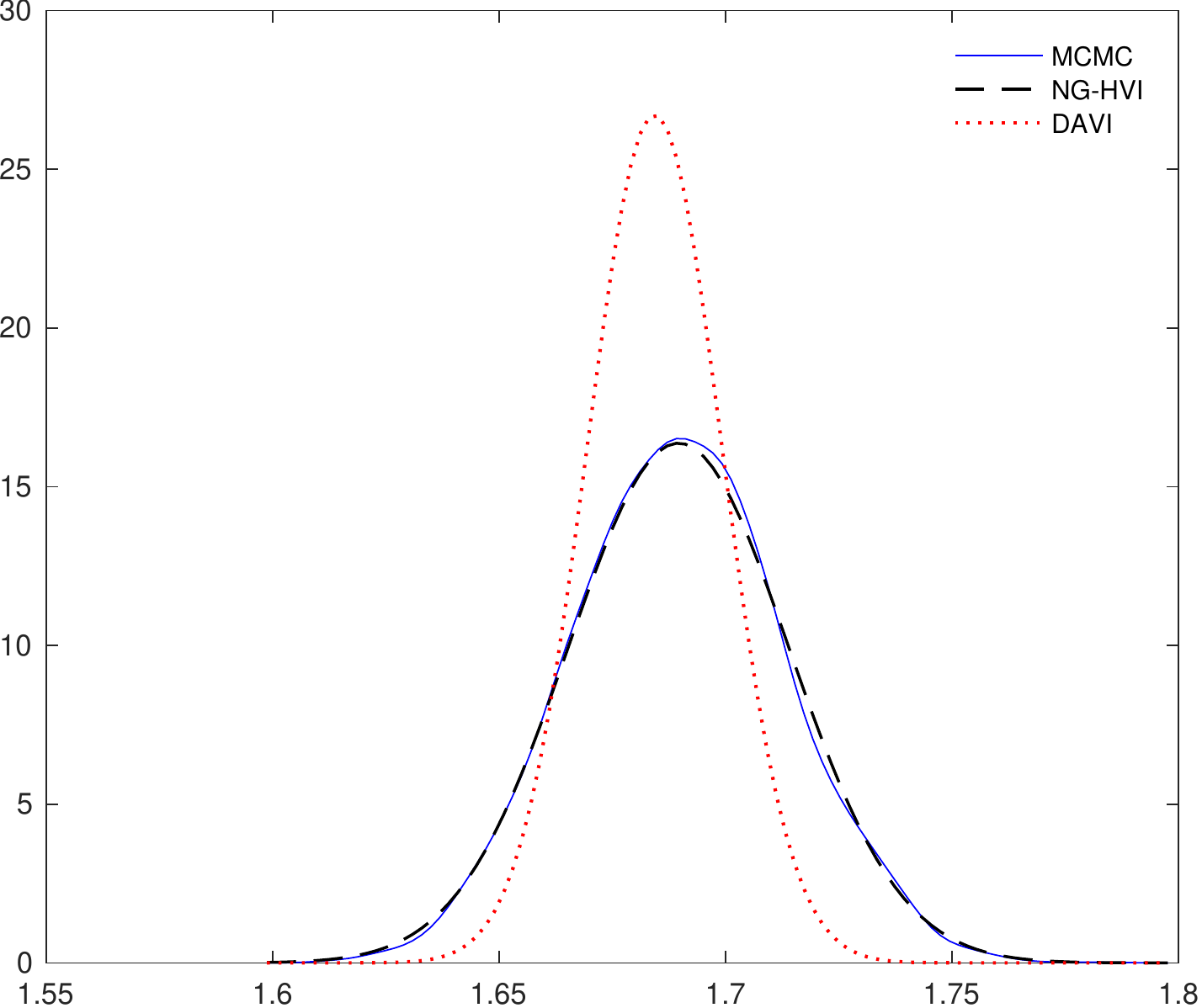}
	\end{subfigure}
	\begin{subfigure}{0.33\linewidth}
		\subcaption{posterior of $\beta_5$}
		\includegraphics[width=1\linewidth]{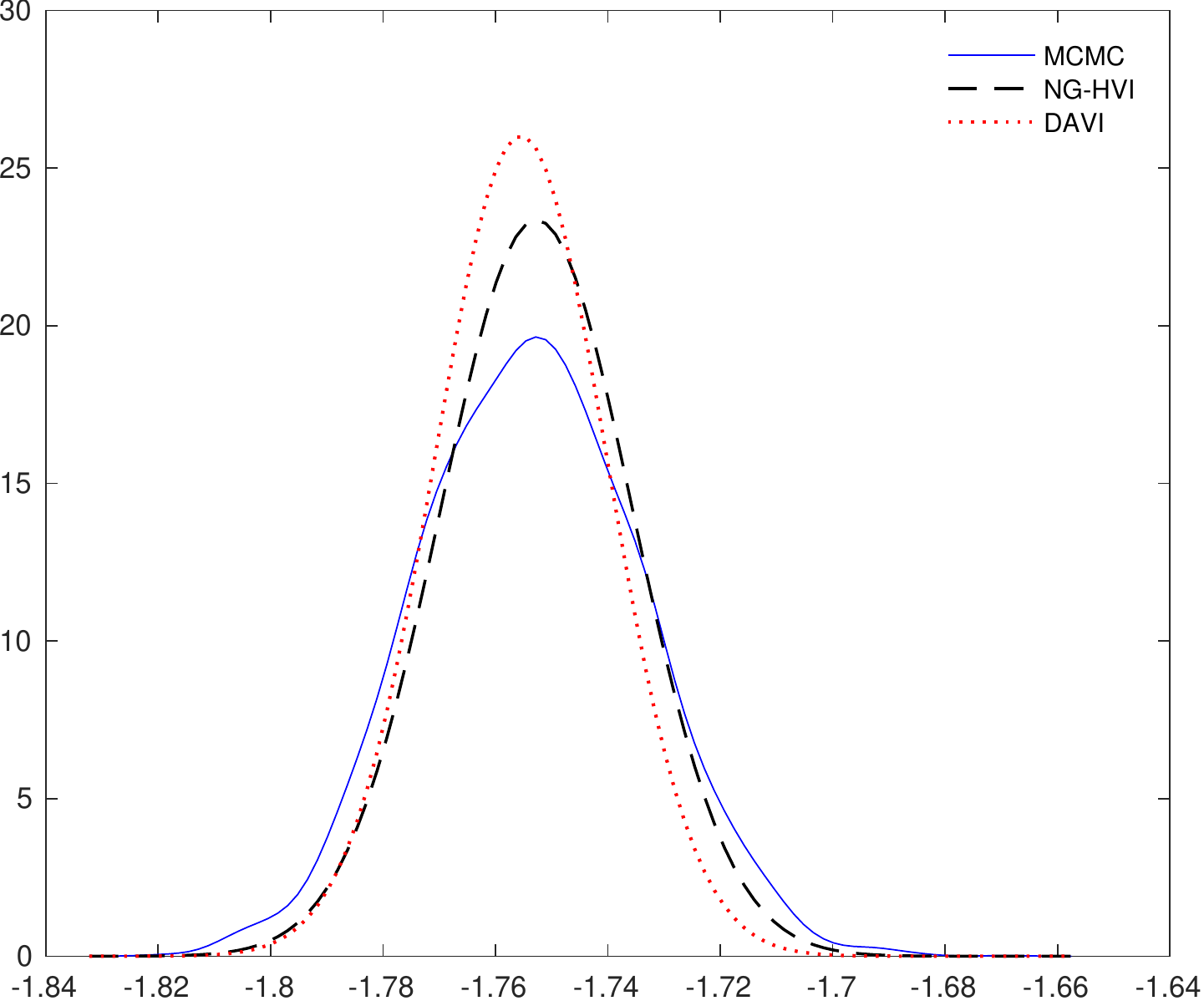}
	\end{subfigure}
	\begin{subfigure}{0.33\linewidth}
		\subcaption{posterior of $\beta_6$}
		\includegraphics[width=1\linewidth]{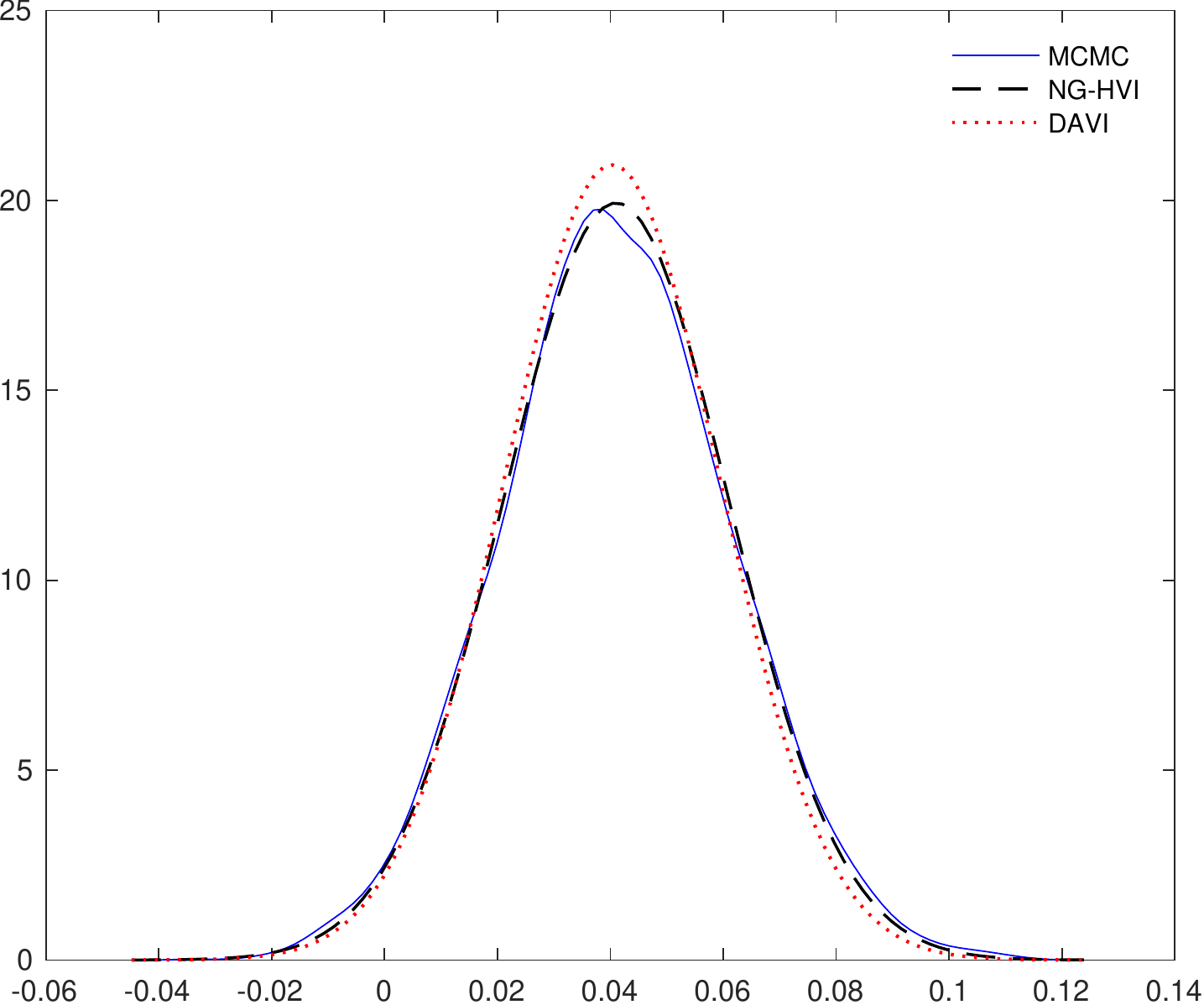}
	\end{subfigure}
	
	\begin{subfigure}{0.33\linewidth}
		\subcaption{Posterior density of $\log(\sigma^2_\epsilon)$}
		\includegraphics[width=1\linewidth]{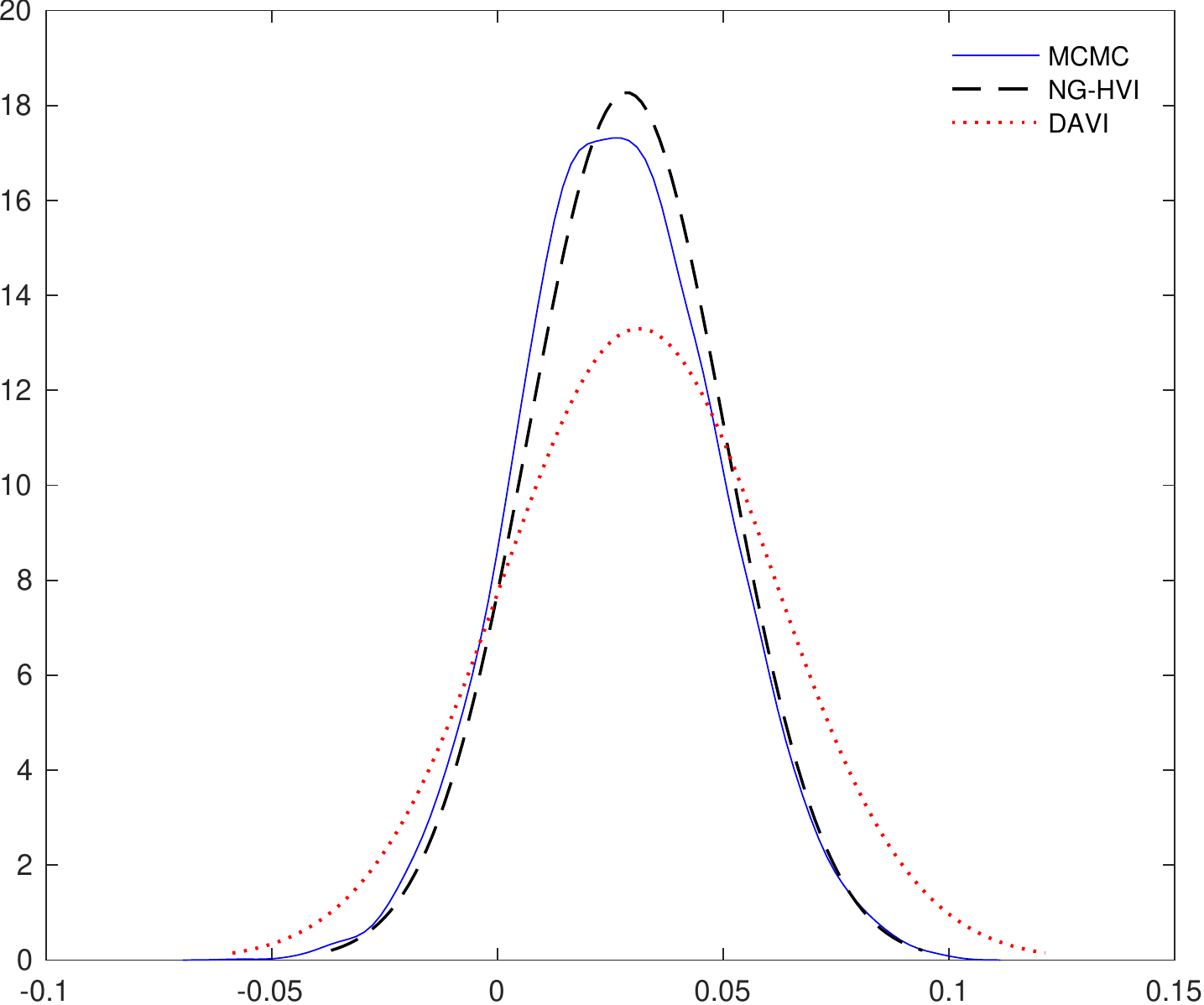}
	\end{subfigure}
	\begin{subfigure}{0.32\linewidth}
		\subcaption{Posterior density of $\log(\sigma^2_\alpha)$}
		\includegraphics[width=1\linewidth]{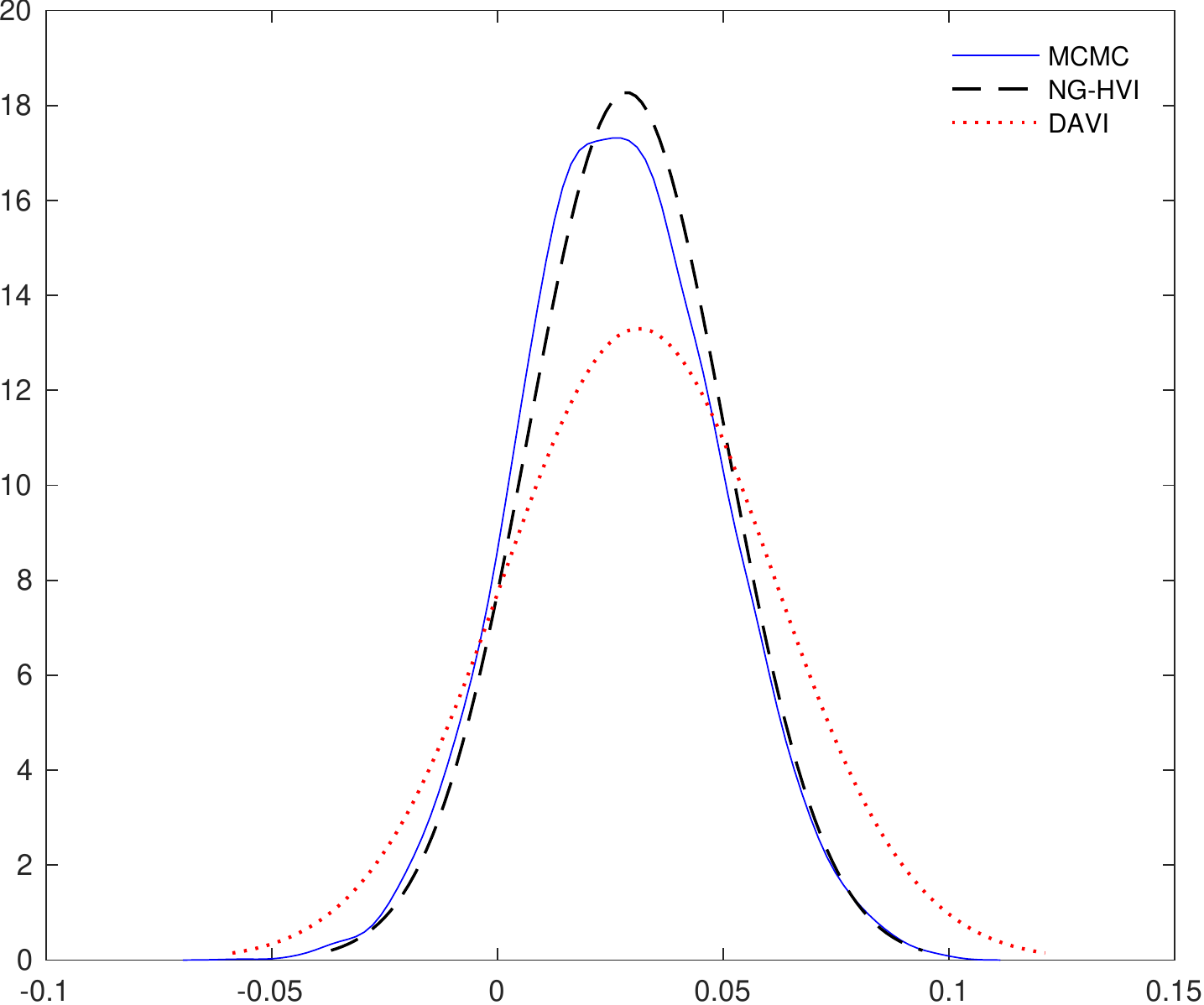}
	\end{subfigure}
	\caption{Comparison of the approximate posterior densities of the elements of $\thetavec$ from the VI methods and the exact posterior obtained using MCMC. These results are for the linear random effects regression when of $\sigma^2_\epsilon = \sigma^2_\alpha = 1$ and $K = 1000$. Results are based on 10000 steps to ensure convergence.}
	\label{fig:MCMC_LMM_tan}
\end{figure}

Computation of the exact posterior using MCMC is standard, and the exact
marginal posteriors are presented in Figure~\ref{fig:MCMC_LMM_tan}
for the case where $\sigma^2_\epsilon = \sigma^2_\alpha = 1$ and $K = 1000$. 
Variational posterior densities
are also given for the NG-HVI.
The variational posteriors from NG-HVI are more
accurate than those from DAVI, with the latter poorly 
calibrating the approximations for $\sigma^2_\alpha$ and $\sigma^2_\epsilon$ and under-estimating the variance for $\betavec$. In particular, the variational posterior for the intercept coefficient $\beta_1$ produced by DAVI in panel~(a) has very small variance. 

\newpage
\noindent  {\bf {C.2: Probit random effects model}}\\
We outline an additional example that extends
the linear random effects model in Section~3.4 (Example~1) 
to a probit model with random effect. This further illustrates
the increase in optimization efficiency using SNGA over SGA. 
Consider a probit model with a scalar random effect:
\begin{align*}
	y_{i}^* &= \betavec^\top \xvec_{i} + \alpha_k + \epsilon_{i} \\
	y_{i} & = \mathds{1}(y_{i}^* > 0)\,,
\end{align*}
where the indicator function
$\mathds{1}(X)=1$ if $X$ is true, and zero otherwise, and 
$\epsilon_i\sim N(0,1)$. Following much of the random effects literature, index $k$ denotes the group membership of observation $i$. 

We generate $n=2000$ observations from a DGP
that has $K=100$ groups and $3$ covariates as follows:
\begin{enumerate}
	\item[(i)] Generate $\xvec^0_i \sim N_3(\muvec_x, \Sigma_x)$ with $\muvec_x = \mathbf{0}_3, \Sigma_x = \L\L' + diag(\dvec)$, and set  $\xvec_i = (1,\xvec^{0,\top}_i)^\top$.
	\item[(ii)] Generate $\betavec \sim N(\mathbf{0}_4, 10 I_4)$. 
	\item[(iii)] Generate $\alpha_k \sim N (0,1)$ for $k = 1, \dots, 100$. 
	\item[(iv)] Generate $y_{i}^* \sim N(\betavec^\top \xvec_{i} + \alpha_k,1)$ and set $y_{i} = \mathds{1}(y_{i}^* > 0)$ for $i=1,\ldots,2000$.
\end{enumerate}

In this example $y_i^*$ and $\alpha_k$ are both latent variables, so for implementing the HVI methods we set
$\zvec = (\yvec^{*\top},\alphavec^\top)^\top$ where 
is $\yvec^{*} = (y_1^*,y_2^*,\dots, y_n^*)^\top,\alphavec = (\alpha_{1}, \alpha_{2}, \dots, \alpha_{K})^\top $. To generate from the distribution $\zvec|\thetavec,\yvec$ we use MCMC to generate alternately from $\alphavec|\yvec^\star,\thetavec$ and then from $\yvec^\star|\alphavec,\yvec$. 
Generation from each is very fast because their densities are products of
independent univariate Gaussian densities. We find that just five sweeps of this 
MCMC scheme, initialized at the values of $\zvec$ from the last step of 
the stochastic optimization algorithm, provides strong results. 
The approach of using MCMC within SGA was explored
by~\cite{loaiza-mayaFastAccurateVariational2022} for more complex models 
and found to work well. 

\begin{figure}[ht!]
	\begin{subfigure}{0.5\linewidth}
		\subcaption{Convergence of RMSE against steps}
		\includegraphics[width=1\linewidth]{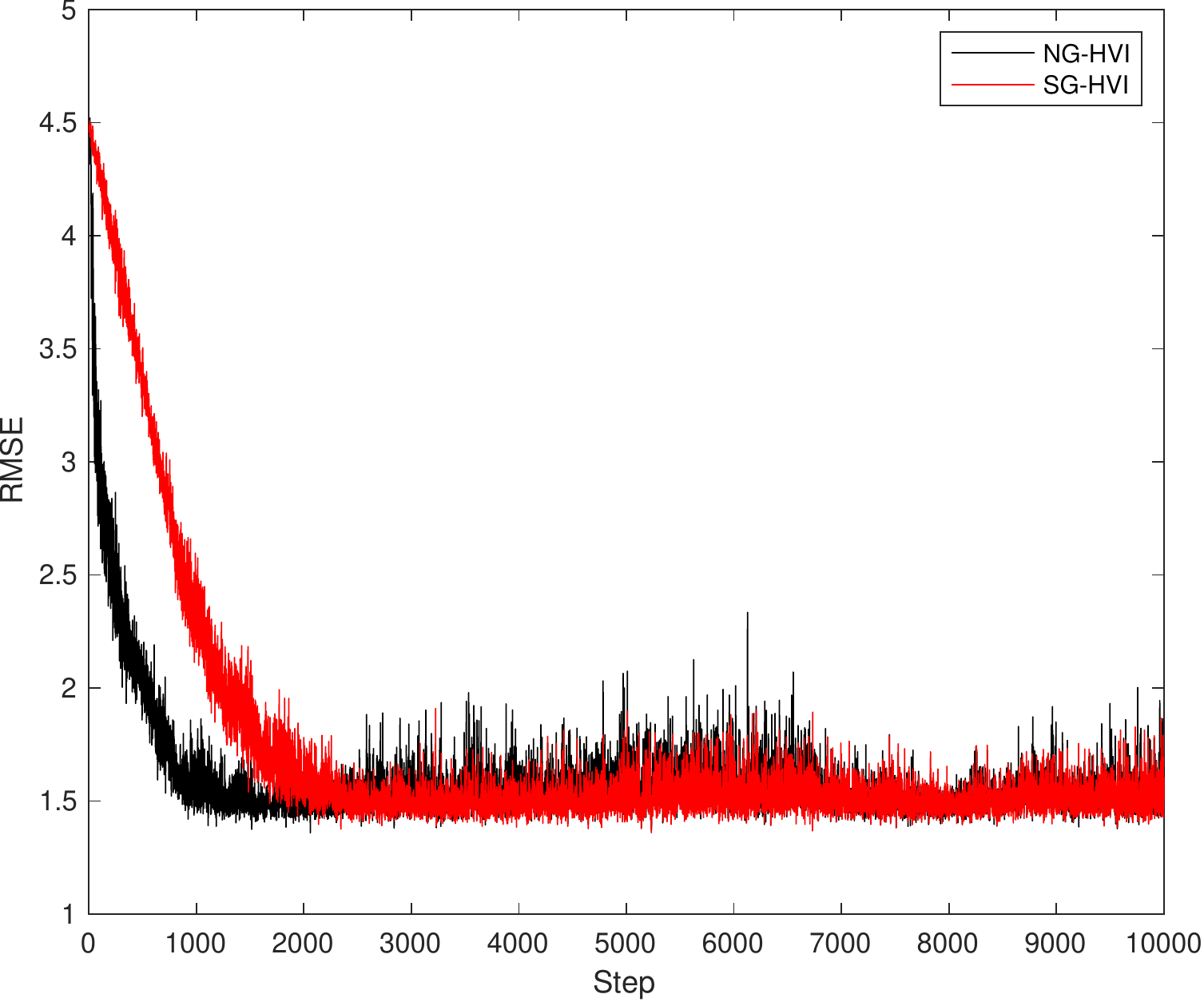}
	\end{subfigure}
	\begin{subfigure}{0.485\linewidth}
		\subcaption{Convergence of RMSE against wall clock}
		\includegraphics[width=1\linewidth]{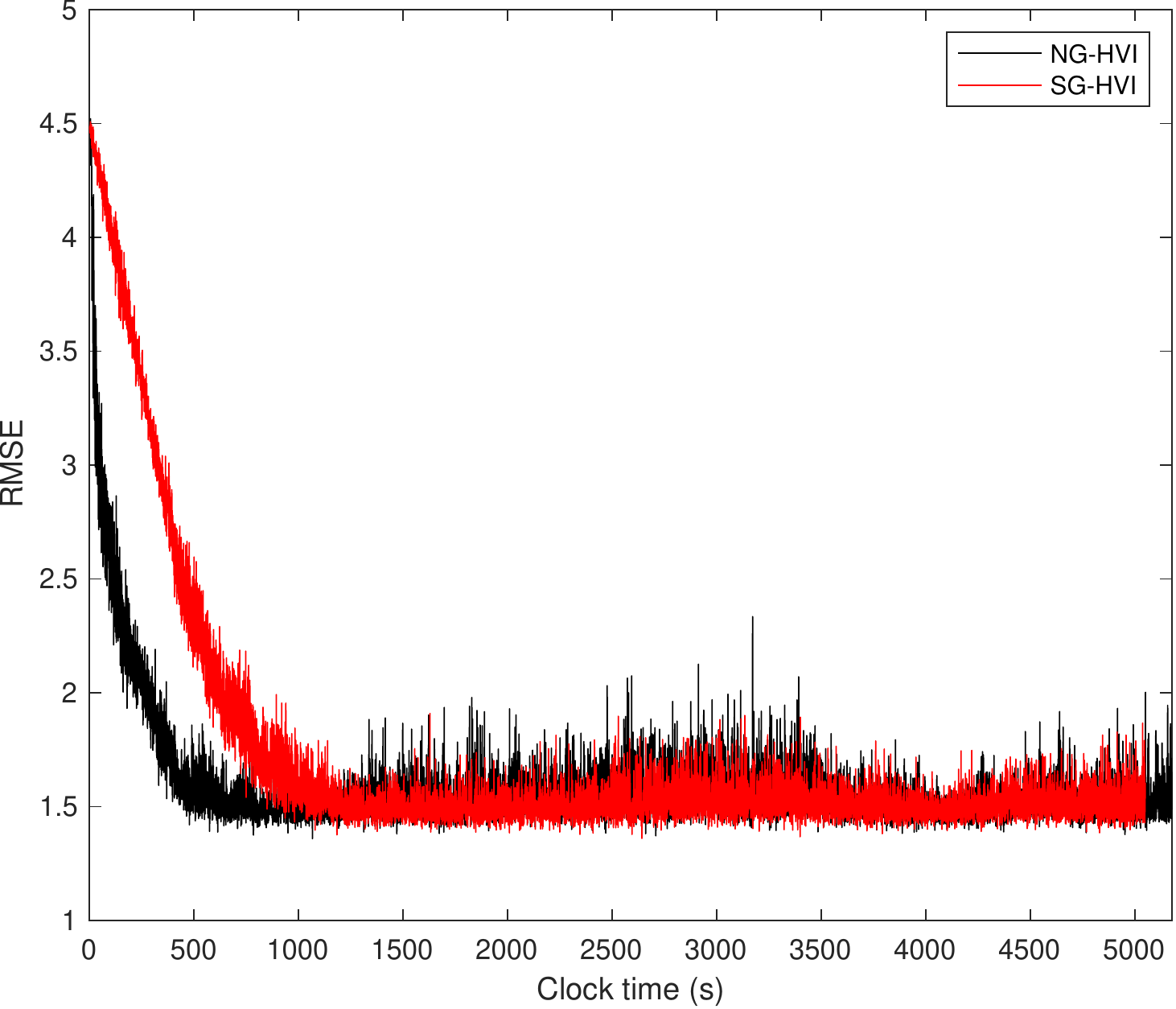}
	\end{subfigure}
	\caption{Convergence of RMSE.}
	\label{converge_rmse}
\end{figure}

In this simulation, the values $\yvec^*$ are known. Therefore, to monitor convergence, the
root mean squared error
\begin{align*}
	RMSE = \sqrt{\frac{\sum_{i = 1}^{2000}(\hat{y}_{i}^* - y_{i}^*)^2}{2000}}
\end{align*}
is computed,
where $\hat{y}_{i}^*$ is the variational mean of the latent variable $y_i^*$.
Figure~\ref{converge_rmse} plots this against (a)~step number of the HVI algorithm, and~(b) wall clock time. 
By this measure, NG-HVI convergences in less than
half the time required by SG-HVI. 

\clearpage
\noindent {\bf \large{Part~D: Additional Details for the Examples in Section~4}}\\
\ \\
\noindent  {\bf {D.1: Parameterization, priors and gradients}}\\
This section provides further details on the parameterization and choice of priors, then provides the required gradients and derivatives to implement
the HVI method.

The model parameters of the Gaussian DMM in Section~4 are $\thetavec = (\wvec^\top,\betavec^\top,\sigma^2_\epsilon,\lvec^\top)^\top$. For variational inference, we need to transform $\sigma^2_\epsilon$ to the real line. Thus we introduce $\theta_\epsilon = \log(\sigma^2_\epsilon)$ and the updated model parameters are $\thetavec = (\wvec^\top,\betavec^\top,\theta_\epsilon,\lvec^\top)^\top$. Then we have $g(\thetavec,\zvec) = \prod_{k=1}^K \bigl(p(\yvec_k|\alphavec_k,\thetavec) p(\alphavec_k|\thetavec)\bigr) p(\thetavec)$ with 
\begin{align*}
	p(\yvec_k|\alphavec_k,\thetavec) & =  \phi(\yvec_k;H_k(\betavec + \alphavec_k), \sigma^2_\epsilon)\\
	P(\alphavec_k|\thetavec) & =  \phi(\alphavec_k;\boldmath{0},\Omega_\alpha)\\
	p(\wvec) & =  \phi(\wvec;\boldmath{0},\sigma^2_w I)\\
	p(\betavec) & =  \phi(\betavec;\boldmath{0},\sigma^2_\beta I)\\
	p(\theta_\epsilon) & = \frac{b^a}{\Gamma(a)}\exp(\theta_\epsilon)^{-a}\exp\bigl(-b\,\exp(-\theta_\epsilon) \bigr)\\
	p(\lvec) & = 2^{m_L} p(\Omega_\alpha^{-1})\prod_{i=1}^{m_L}l_{ii}^{(m_L-i+2)}
\end{align*}
Here, $p(\theta_\epsilon)$ was constructed by using the prior $p(\sigma^2_\epsilon)= p_{IG}(\sigma^2_\epsilon;a,b)$ and deriving the corresponding prior on $\theta_\epsilon$. The density of a $N(\muvec,\Sigma)$ distribution is denoted as $\phi(\xvec; \muvec,\Sigma)$, 
and $p(\Omega_\alpha^{-1})$ has is the density 
of a Wishart distribution with degrees of freedom $\nu$ and scale matrix $0.01I$. 
For this example we set $\sigma^2_w = 100$ and $\sigma^2_\beta = 100$. $H_k$ is the $n_k \times (m_L+1)$ matrix with each row containing $\hvec^{(L)}_i$ for $i \in k$. \\

Given above definitions, the function $\log g(\thetavec,\zvec)$ with $\zvec = (\alphavec_1^\top, \dots, \alphavec_K^\top)^\top$ can be written as:
\begin{align*}
	 \log g(\thetavec,\zvec) = \sum_{k=1}^K\bigl( \log p(\yvec_k|\alphavec_k,\thetavec) +\log p(\alphavec_k|\thetavec) \bigr) + \log p(\wvec) + \log p(\betavec)+ \log p(\theta_\epsilon) + \log p(\lvec)
\end{align*}
The gradient of $\log p(\yvec_k|\alphavec_k,\thetavec)$ with respect to $\wvec$ and $\betavec$ can be evaluated jointly. Denote $\cvec = (\wvec^\top, \betavec^\top)$, then
\begin{align*}
	\nabla_c \log p(\yvec_k|\alphavec_k,\thetavec) &= \sigma^{-2}_\epsilon \nabla_c \etavec_k\\
	\etavec_k & = H_k (\betavec + \alphavec_k)
\end{align*}
$\nabla_c \etavec_k$ can be evaluated using standard back-propagation algorithm. The gradient of $\log p(\yvec_k|\alphavec_k,\thetavec)$ with respect to $\theta_\epsilon$ can be computed as:
\begin{align*}
	\nabla_{\theta_\epsilon} \log p(\yvec_k|\alphavec_k,\thetavec) & = -\frac{1}{2}n_k + \frac{1}{2}\exp(-\theta_\epsilon)(\yvec_k - \etavec_k)^\top (\yvec_k - \etavec_k)
\end{align*}
See Appendix~C in the manuscript for the gradient of $\log g(\thetavec,\zvec)$ with respect to $\lvec$. It is straight forward to show the following:
\begin{align*}
	\nabla_\wvec \log p(\wvec) & = -1/\sigma^2_w\wvec\\
	\nabla_\betavec \log p(\betavec) & = -1/\sigma^2_\beta\betavec\\
	\nabla_{\theta_\epsilon} \log p(\theta_\epsilon) & = - a + b \exp(-\theta_\epsilon)
\end{align*}

The model parameters of the Bernoulli DMM in Section~4 are $\thetavec = (\wvec^\top,\betavec^\top,\omegavec^\top)^\top$, where $\omegavec$ is the non-zero elements of diagonal matrix $\Omega_\alpha$ with each element follows an $IG(0.1,0.1)$ distribution. Again, for variational inference we use log-transformation to transform $\omegavec$ to real line. The prior for $\wvec$ and $\betavec$ are normal distribution with mean $\mathbf{0}$ and $\sigma^2_w = 50$, $\sigma^2_\beta = 5$. The function $\log g(\thetavec,\zvec)$ with $\zvec = (y_1^*, \dots, y_n^*, \alphavec_1^\top, \dots, \alphavec_K^\top)^\top$ is now:
\begin{align*}
	g(\thetavec,\zvec) & =  \sum_{k = 1}^{K}\sum_{j = 1}^{n_k} \bigl( (\mathds{1}(y^*_{j} \leq 0, y_{j}= 0) + \mathds{1}(y^*_{j} > 0, y_{j}= 1)) \log p(y^*_{j}|\alphavec_k,\thetavec) + \log p(\alphavec_k|\thetavec) \bigr) \nonumber \\
	& \quad + \log p(\wvec) + \log p(\betavec) + \log p(\omegavec) \nonumber\\
	& =  \sum_{k = 1}^{K}\sum_{j = 1}^{n_k} \bigl( \log p(y^*_{j}|\alphavec_k,\thetavec) + \log p(\alphavec_k|\thetavec) \bigr) + \log p(\wvec) + \log p(\betavec) + \log p(\omegavec)
\end{align*}
$\mathds{1}(y^*_{j} \leq 0, y_{j}= 0) + \mathds{1}(y^*_{j} > 0, y_{j}= 1)$ always sum to $1$ as $y_i^*$ is drawn from truncated normal using HVI methods, thus the second equation follows. The gradient of $\log p(y^*_j|\alphavec_k,\thetavec)$ with respect to $\wvec$ and $\betavec$ again can be evaluate using back-propagation algorithm in a similar way as above, with $\sigma^2_\epsilon=1$.

\noindent  {\bf {D.2: Additional details on Example~2}}\\
In the data generating process (DGP) in Section~4.3, we simulated data from a DMM with two hidden layers, each with 5 neurons. The inputs $\xvec_i\sim N(\bm{0},V_x)$ with
\begin{align*}
	V_x = \begin{pmatrix}
		1  &  0 &-0.5  & 0.2  & 0\\
		0   & 1  &  0  & -0.5 &0.2\\
		-0.5 &   0  &  1  &  0 &-0.5\\
		0.2 &-0.5  &  0  &  1  &  0\\
		0 & 0.2 &-0.5  &  0  &  1
	\end{pmatrix}\,,
\end{align*}
and $\alphavec_k\sim N(0,\Omega_\alpha)$ with $\Omega_\alpha^{-1} = diag( 1.0443,    9.0498,    0.4569,    0.5190,    0.2857,    2.7548)$. 
The output layer fixed effect coefficient vector $\betavec=(0.8292,-1.3250,3.9909,1.6823,-1.7564,0.5580)^\top$, where the first element is the offset. 
The hidden layer weights are fixed across simulations, with their values derived
by fitting NAGVAC to the simulated dataset provided in \cite{tranBayesianDeepNet2020} and use the weighting matrices in the last step of this algorithm. 
We use $6$ observations from each group for training, $2$ observations 
from each group for testing, and another $2$ observations from each group for implementing the NAGVAC stopping rule. 

To implement step~(b) of Algorithm~1, the conditional posterior  $\alphavec_k|\yvec_k,\thetavec\sim N(\muvec_k,\Sigma_k)$ with
\begin{align*}
	\muvec_k & = \Sigma_k H_k^\top(\yvec_k - H_k\betavec)/\sigma^2_\epsilon\\
	\Sigma_k & = (\Omega_\alpha^{-1} +  H_k^\top H_k/\sigma^2_\epsilon)^{-1}
\end{align*}

We also use NAGVAC to fit the data using the code provided by the authors with default values. This approach assumes diagonal $\Omega_\alpha$, which matches the DGP. It also employs a stopping rule based on improvements in predictive cross entropy (PCE) for validation data, for which we draw a further 2 observations per group from the DGP. However, we find that NAGVAC does not work well with the simulated data in this example in terms of predictive accuracy, producing negative $R^2$ for both in-sample and out-of-sample prediction. Thus we do not include the results of NAGVAC in this example.
\vspace{20pt}

\noindent {\bf {D.3: Additional details on Example~3}}\\ 
The DGP in Example~3 in Section~4 generates $y_i \sim \mbox{Bernoulli}(p_i)$ where
\begin{align*}
	p_i &= \frac{1}{1+exp(-\eta_i)}\\	
		\eta_i &= 2 + 3(x_{i,1} - 2x_{i,2})^2 - 5\left(\frac{x_{i,3}}{(1+x_{i,4})^2}\right) - 5x_{i,5} + a_k\,.
\end{align*}
Following~\cite{tranBayesianDeepNet2020}, the five covariates
$x_{i,j}, j = 1,\dots, 5$ are generated from uniform distributions $\mathcal{U}(-1,1)$ and $a_k \sim N(0,1)$. This DGP is the same as that 
suggested by these authors, except that we employ a logistic function for
determining $p_i$, rather than a probit function. The reason for the difference is to advantage
the NAGVAC algorithm which also uses a logistic link.

Both HVI methods employ an MCMC step within the SNGA optimization. 
For both HVI methods, we draw alternately from the conditional posterior 
of $\alphavec$ and $y^*$. These both factorize into products over the groups
and observations, respectively, with
\begin{align*}
	\alphavec_k|\thetavec, \yvec_k^*  &\sim \mathcal{N}\left((\Omega_\alpha^{-1} + H_k'H_k)^{-1}H_k'(\yvec_k^* - H_k'\betavec),(\Omega_\alpha^{-1} + H_k'H_k)^{-1}\right) \\
	y_i^* | \alphavec_k, \thetavec, y_i &\sim \begin{cases} TN_{(0,\infty)}(\hvec^{(L)\top}_i(\betavec + \alphavec_k),1)\quad \text{ if } y_i =1\\  TN_{(-\infty, 0 )}(\hvec^{(L)\top}_i(\betavec + \alphavec_k),1)\quad \text{ if } y_i = 0 \end{cases} 
\end{align*}
Here $H_k$ has the same definition as above, while
 $TN_{(a, b )}(\mu,\sigma^2)$ represents truncated normal distribution with mean $\mu$, variance $\sigma^2$ and support $[a,b]$. We run this MCMC scheme for 5 sweeps, after initializing $\alphavec,\yvec^*$ at their values from the last step of the optimization
algorithm. The approach of using MCMC within SGA was explored
by~\cite{loaiza-mayaFastAccurateVariational2022} for more complex models 
and found to work well. 

\noindent {\bf {D.4: Algorithms for evaluating DMM predictive distributions}}

\begin{algorithm}[h]
	\begin{algorithmic}[0]
		\State Denote  $\lambdavec^*$ to be the fitted variational parameter, and $X_{train}$, $\yvec_{train}$ and $n_{train}$ to be the design matrix,  output vector and number of observations in the training data set, respectively. For the testing sample denote these quantities as $X_{test}$, $\yvec_{test}$ and $n_{train}$.
		\For{$j = 1 : J $}
		\State  (1) Generate $\thetavec^{j} \sim q_{\lambdavec^*}(\thetavec)$.
		\State  (2) Generate $\alphavec^{j}$ from conditional variational posterior $q_{\lambdavec^*}(\alphavec|\thetavec^{j})$.
		\For{$i = 1:n_{test}$}.
		\State  (3.1) Set $\eta^{j}_{i} = f_{L+1}\big((\betavec^{j} + \alphavec^{j}_k)^\top\hvec^{(L)}_i  \big)\circ f_L\left(W_L^{j}\hvec^{(L-1)}_i\right)\circ \dots \circ f_1\left(W^{j}_1 \xvec_{test,i}\right)$.
		\State (3.2)  Set $p^{j}_i = \phi(y_i; \eta^{j}_{i}, \sigma_\epsilon^{2,j})$.
		\EndFor
		\EndFor
		\For{$i = 1:n_{test}$} 
		\State (4.1) $\hat{y}_i = \frac{1}{J}\sum_{j = 1}^{J}\eta^{j}_{i}$.
		\State (4.2) $\hat{p}_i = \frac{1}{J}\sum_{j = 1}^{J}p^{j}_i$.
		\EndFor
		\State (5.1) $R^2_{test} = 1 - \frac{\sum_{i=1}^{n_{test}}(y_{i}-\hat{y}_i)^2}{\sum_{i=1}^{n_{test}}(y_{i}-\bar{y})^2}$ with $\bar{y} = \frac{1}{n_{test}}\sum_{i=1}^{n_{test}}y_i$
		\State (5.2) $RMSE_{test} = \sqrt{\sum_{i=1}^{n_{test}}(y_{i}-\hat{y}_i)^2/n_{test}}$.
		\State (5.3) $LS_{test} = \frac{1}{n_{test}}\sum_{i=1}^{n_{test}}\log{\hat{p}_i}$.
	\end{algorithmic}
	\caption{Evaluating performance of variational predictive distribution for Gaussian DMM}
	\label{predictive_Gaussian}
\end{algorithm}
For HVI methods, we generate $\alphavec^{j}$ from its variational conditional posterior $p(\alphavec_k |\thetavec^{j},X_{train},\yvec_{train})$. For DAVI we generate $\alphavec^{j}$ from calibrated variational approximation $q_{\lambdavec^*}(\alphavec|\thetavec^{j},X_{train},\yvec_{train})$. As NAGVAC integrate out $\alphavec$, it does not calibrate a variational posterior for the random effects, thus we generate $\alphavec^{j}$ from calibrated normal distribution $N(\mathbf{0}, \Omega_\alpha^j)$. 

\clearpage
\begin{algorithm}[h]
	\begin{algorithmic}[0]
		\State Denote  $\lambdavec^*$ to be the fitted variational parameter, and $X_{train}$, $\yvec_{train}$ and $n_{train}$ to be the design matrix,  output vector and number of observations in the training data set, respectively. For the testing sample denote these quantities as $X_{test}$, $\yvec_{test}$ and $n_{train}$.
		\For{$j = 1 : J $}
		\State  (1) Generate $\thetavec^{j} \sim q_{\lambdavec^*}(\thetavec)$.
		\State  (2) Implement the following Gibbs sampling scheme\footnotemark:
		\For {$r = 1:R$} 
		\State (2.1)  Generate ${\alphavec_k}^{j} \sim p(\alphavec_k| {\thetavec}^{j}, {\yvec^{*,j}}_{train})$ for $k = 1, \dots,K$.
		\State (2.2)  Generate $ \yvec_{train,i}^{*,j} \sim p(\yvec_{train,i}^{*,j}|{\alphavec}^{j}, {\thetavec}^{j},\yvec_{train,i},\xvec_{train,i})$ for $i = 1, \dots, n_{train}$.
		\EndFor
		\For{$i = 1:n_{test}$}
		\State  (3.1) Set $\eta^{j}_{i} = f_{L+1}\big((\betavec^{j} + \alphavec^{j}_k)^\top\hvec^{(L)}_i  \big)\circ f_L\left(W_L^{j}\hvec^{(L-1)}_i\right)\circ \dots \circ f_1\left(W^{j}_1 \xvec_{test,i}\right)$.
		\State (3.2) ${p}^{j}_{i} = \Phi(\eta^{j}_{i})$\footnotemark.
		\EndFor
		\EndFor
				\For{$i = 1:n_{test}$} 
		\State (4.1)  $\widehat{p}_{i} = \frac{1}{J}\sum_{j=1}^{J}{p}^{j}_{i} $.
		\State (4.2)  $\widehat{y}_{i} = I(\widehat{p}_{i}>0.5)$.
		\EndFor
		\State (5) $PCE_{test} = -\sum_{i = 1}^{n_{test}}(y_i log(\widehat{p}_{i})+(1-y_i)log(1-\widehat{p}_{i}))/n_{test}$.
	\end{algorithmic}
	\caption{Evaluating performance of variational predictive distribution for Bernoulli DMM}
	\label{predictive_binary}
\end{algorithm}
At step (2) of the above algorithm, we generate $\alphavec^{j}$ from calibrated normal distribution $N(\mathbf{0}, \Omega_\alpha^j)$ for NAGVAC, due to the same reason described above. As NAGVAC uses logit link,  ${p}^{j}_{i} = 1/\left(1 + exp(-\eta^{j}_{i})\right)$.
\mycomment{
\addtocounter{footnote}{-2}
\stepcounter{footnote}\footnotetext{At this step, we generate $\alphavec^{j}$ from calibrated normal distribution $N(\mathbf{0}, \Omega_\alpha^j)$ for NAGVAC, due to the same reason described above. }
\stepcounter{footnote}\footnotetext{NAGVAC uses logit link, so ${p}^{j}_{i} = 1/\left(1 + exp(-\eta^{j}_{i})\right)$.} 
}
\newpage

\noindent {\bf {E: Additional Results for Section~\ref{sec:finance}.}}\\
\begin{figure}[ht!]
	\begin{subfigure}{0.45\linewidth}
		\subcaption{ERM}
		\includegraphics[width=0.9\linewidth]{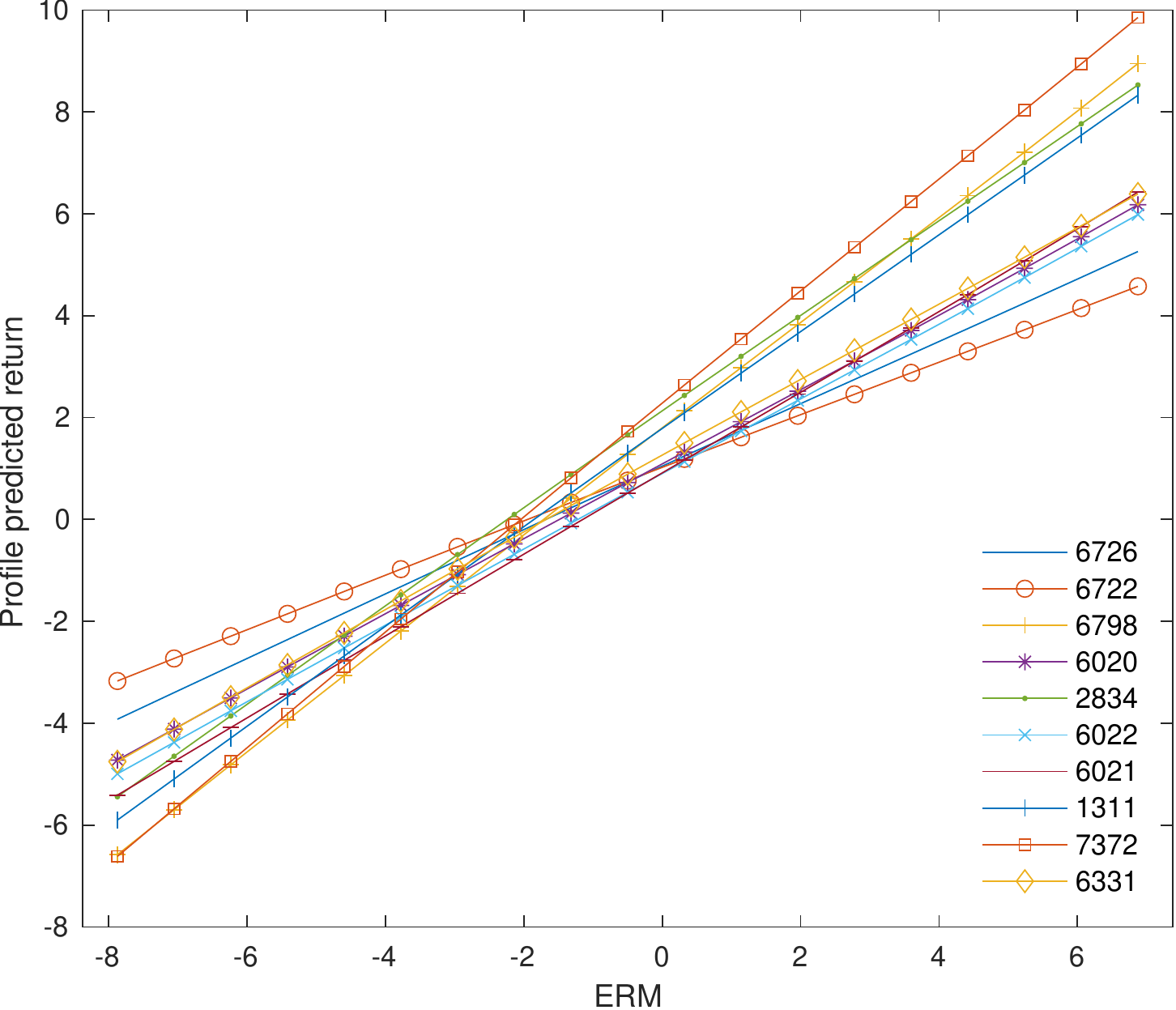}
		\label{ff5_lek_mktrf}
	\end{subfigure}
	\begin{subfigure}{0.45\linewidth}
		\subcaption{SMB}		
		\includegraphics[width=0.9\linewidth]{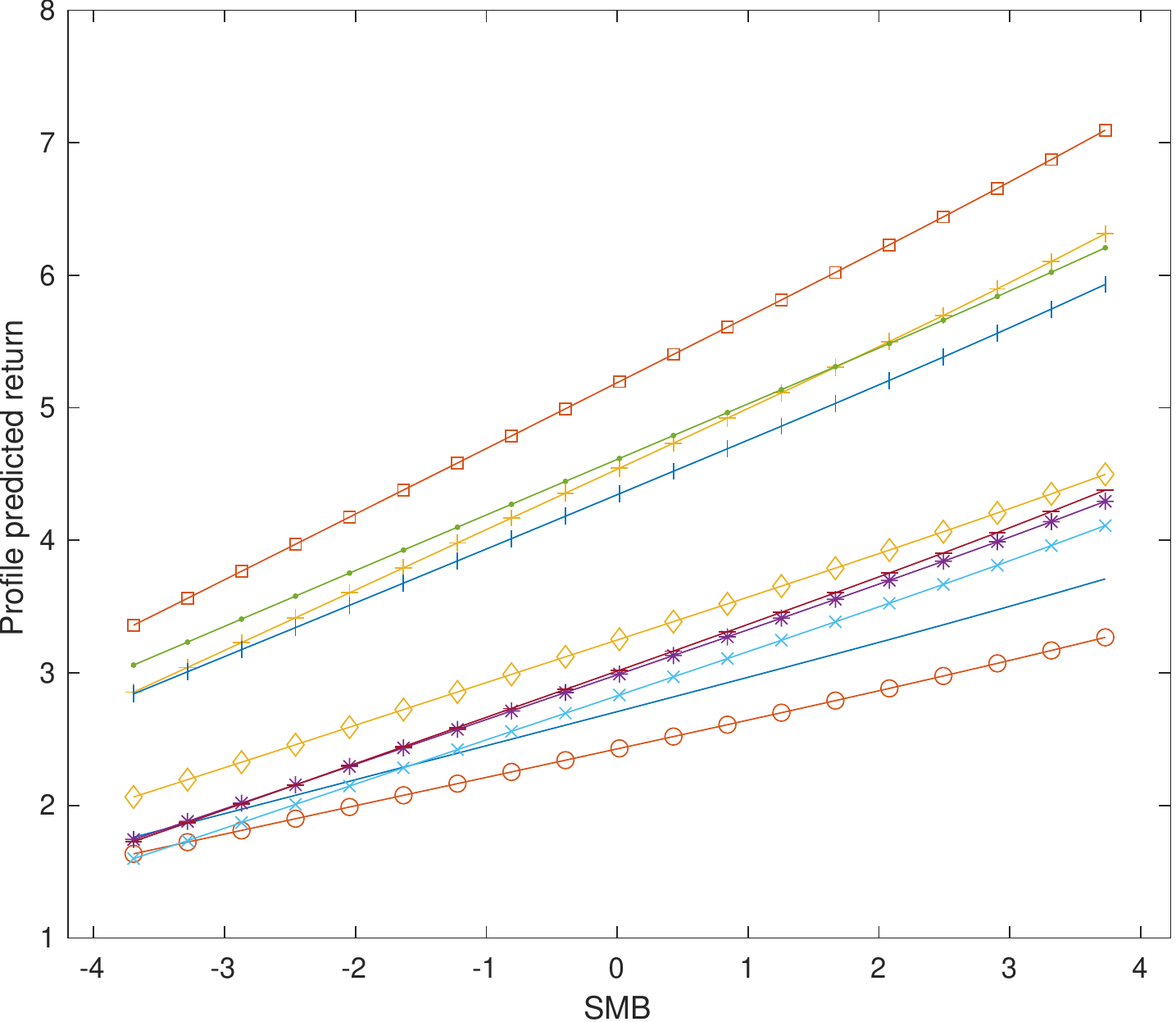}
		\label{ff5_lek_smb}
	\end{subfigure}
	\begin{subfigure}{0.45\linewidth}
		\subcaption{HML}		
		\includegraphics[width=0.9\linewidth]{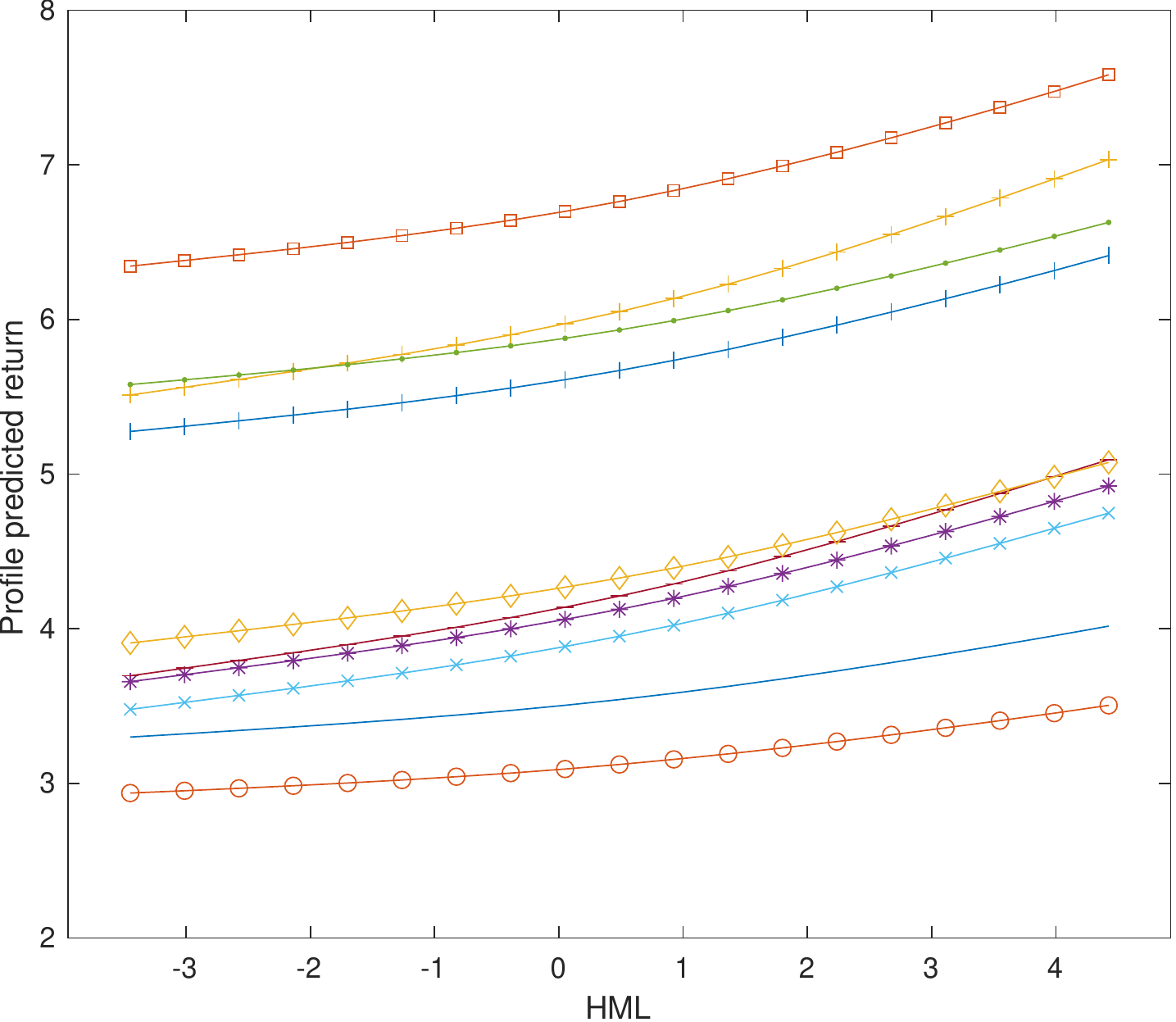}
		\label{ff5_lek_HML}		
	\end{subfigure}
	\begin{subfigure}{0.45\linewidth}
		\subcaption{RMW}		
		\includegraphics[width=0.9\linewidth]{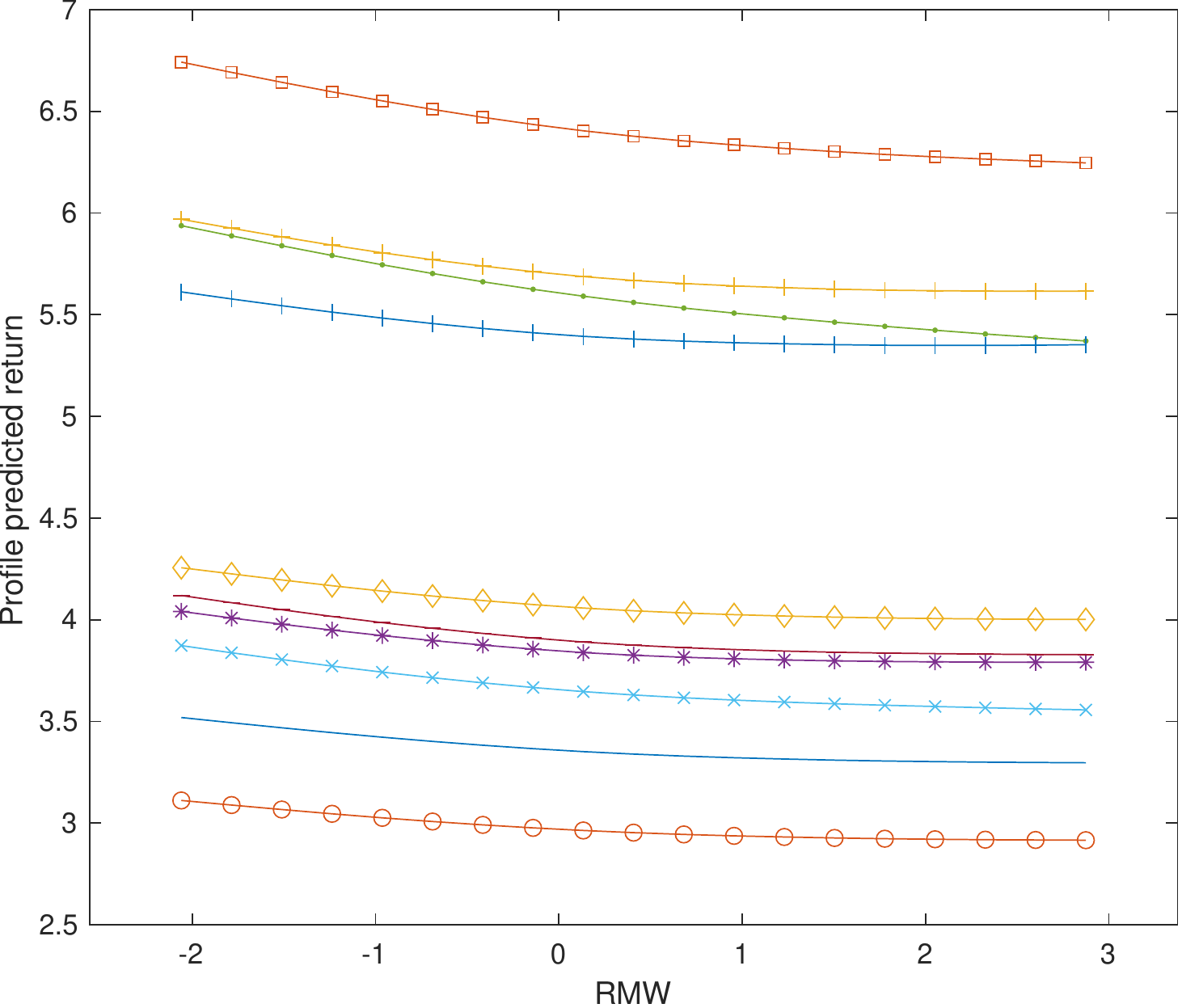}
		\label{ff5_lek_rmw}		
	\end{subfigure}
	\begin{subfigure}{0.45\linewidth}
		\subcaption{CMA}		
		\includegraphics[width=0.9\linewidth]{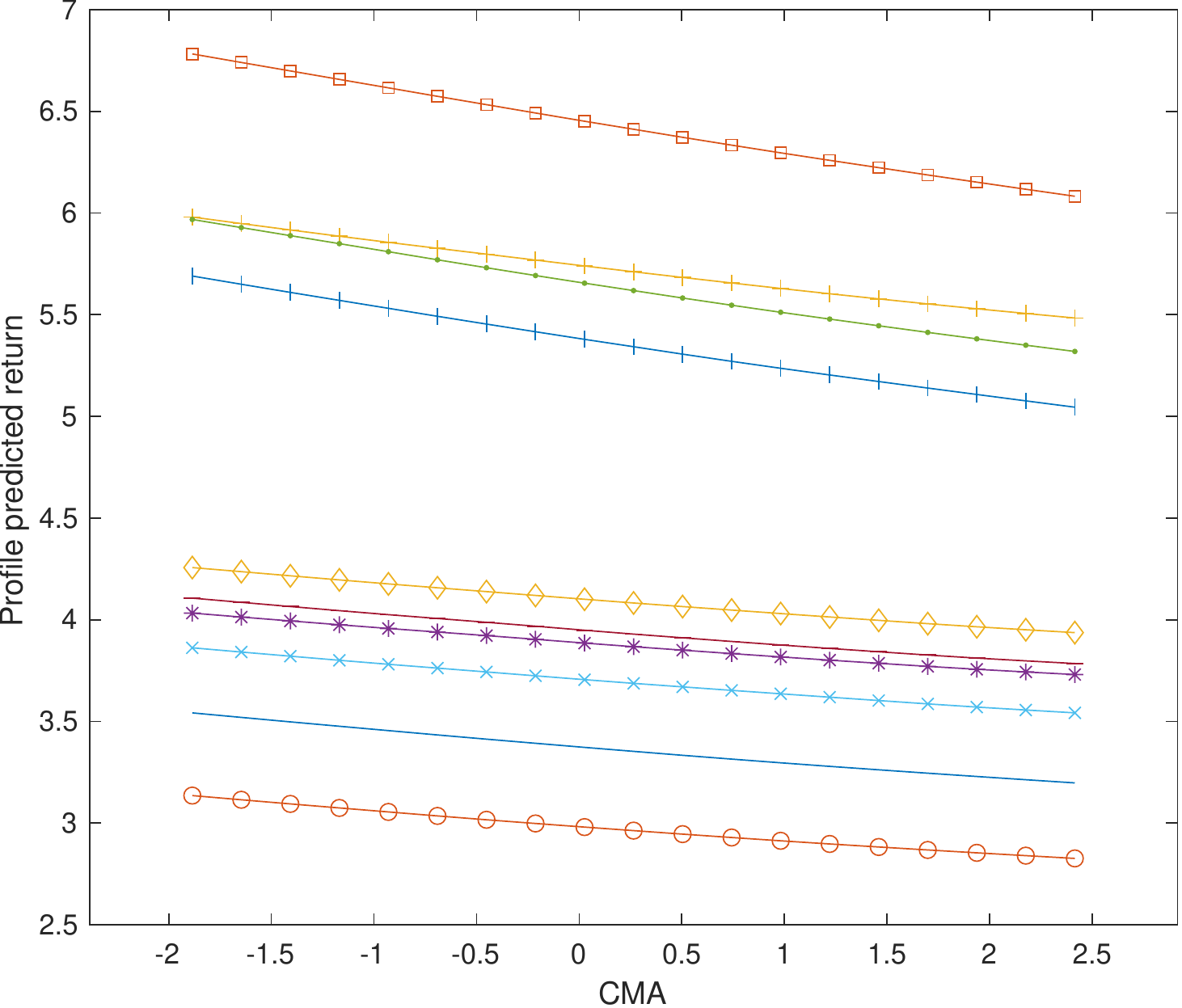}
		\label{ff5_lek_cma}		
	\end{subfigure}
	\caption{Non-linear transformation and heterogeneity of DeepLMM model (5FF). For July 2005 which has low level of stock market volatility (as measured by VIX index).}
	\label{ff5_lek}
\end{figure}
\begin{figure}[ht!]
	\begin{subfigure}{0.45\linewidth}
		\subcaption{ERM}
		\includegraphics[width=0.9\linewidth]{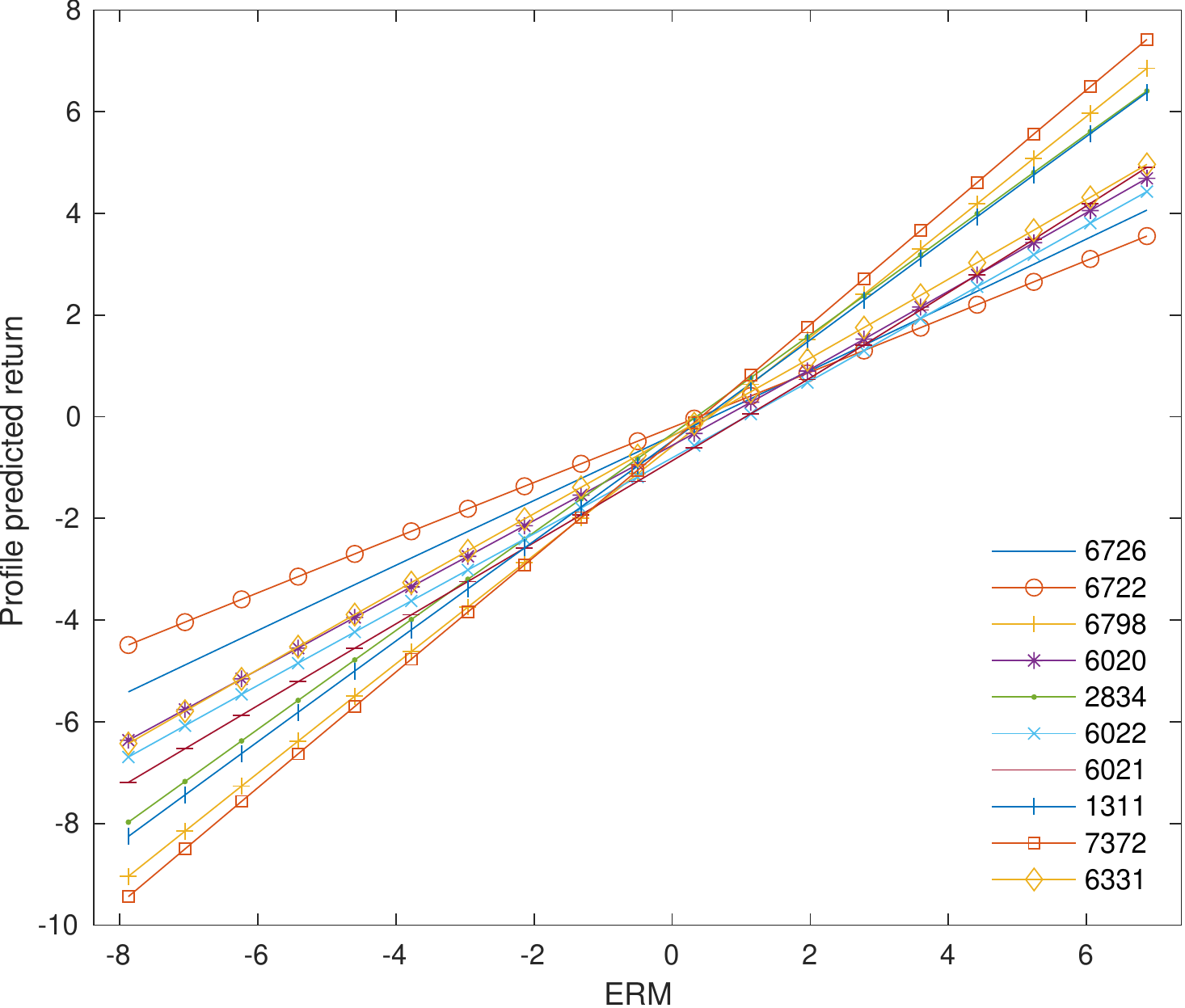}
		\label{ff5_lek2_mktrf}
	\end{subfigure}
	\begin{subfigure}{0.45\linewidth}
		\subcaption{SMB}		
		\includegraphics[width=0.9\linewidth]{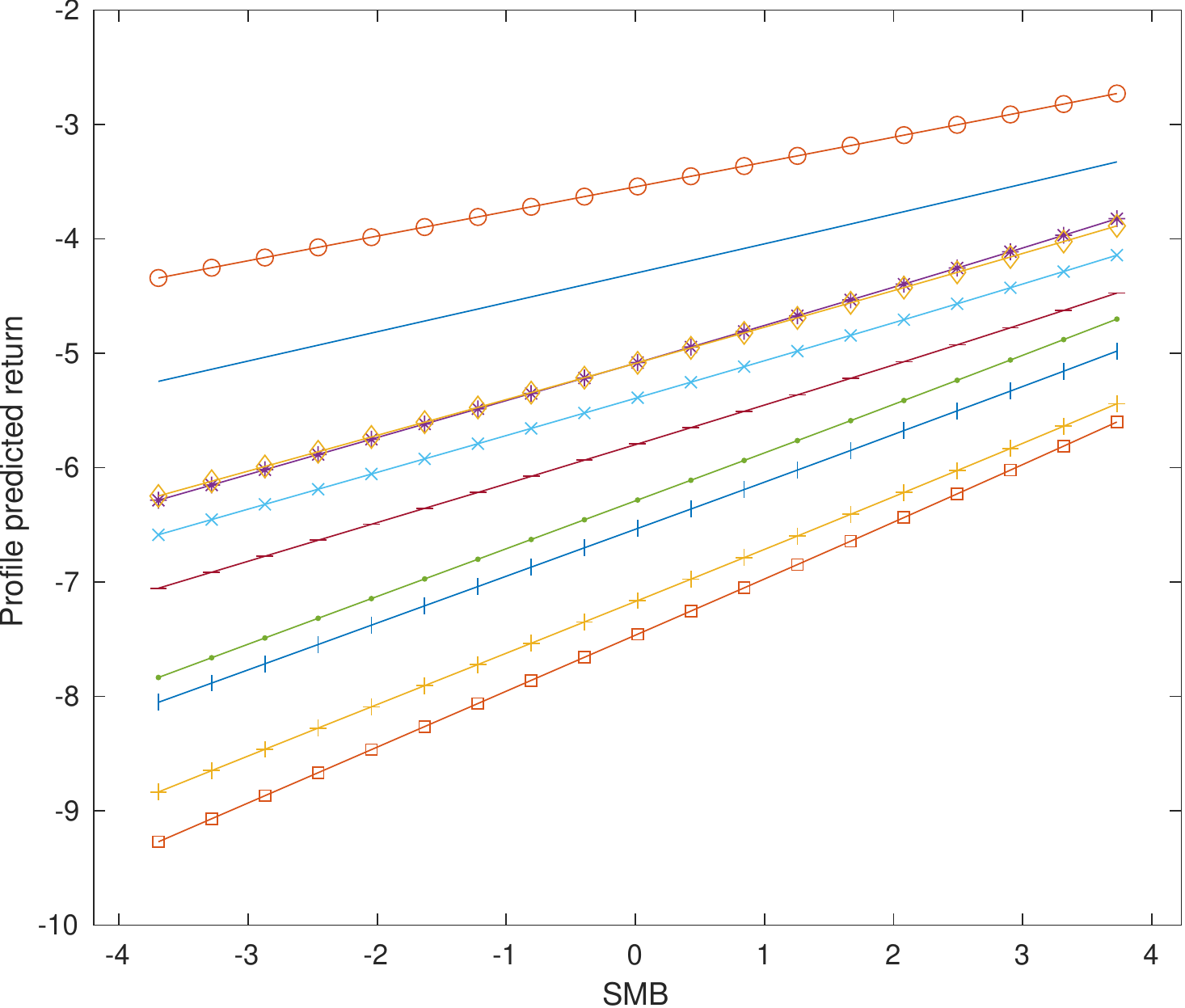}
		\label{ff5_lek2_smb}
	\end{subfigure}
	\begin{subfigure}{0.45\linewidth}
		\subcaption{HML}		
		\includegraphics[width=0.9\linewidth]{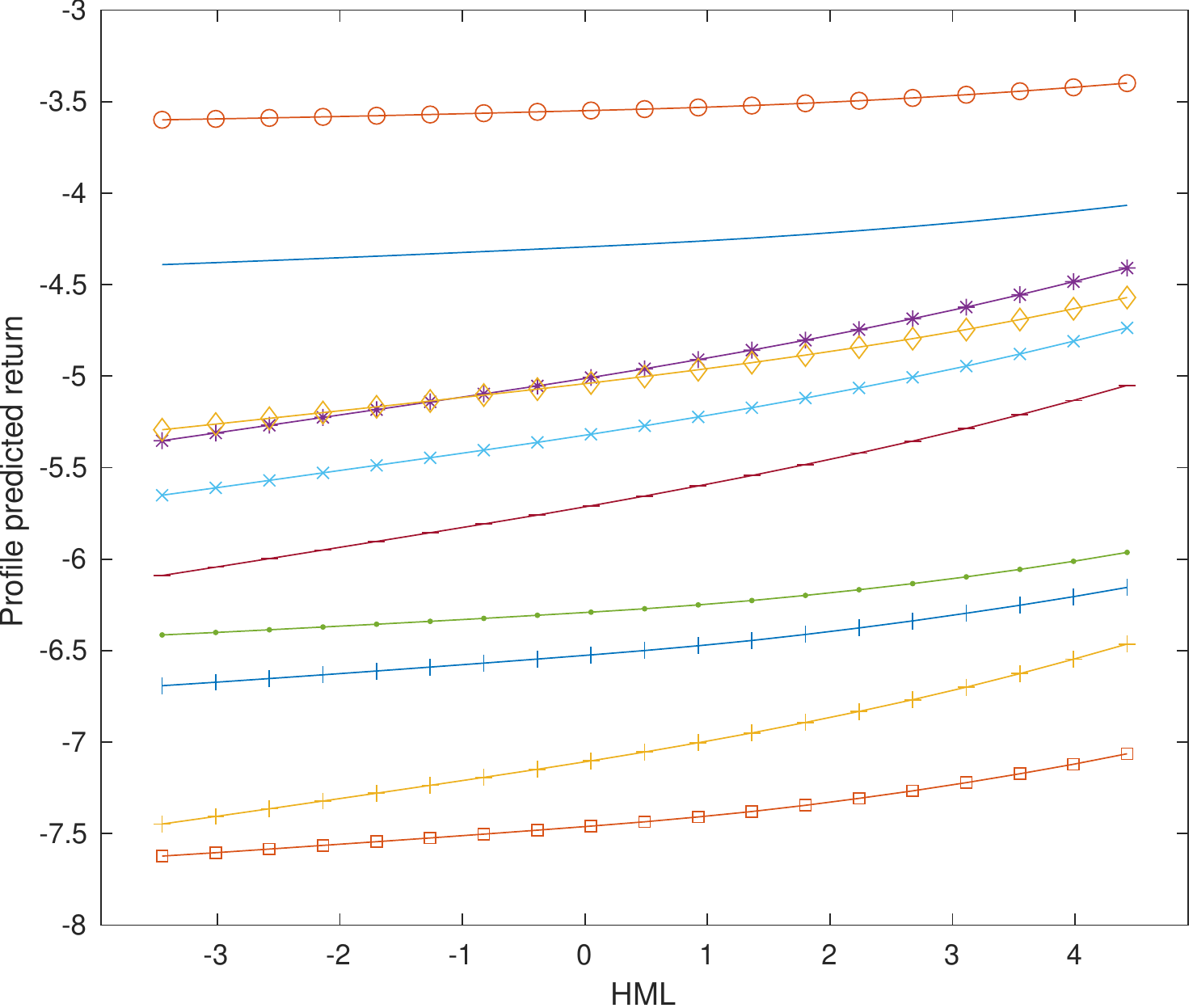}
		\label{ff5_lek2_HML}		
	\end{subfigure}
	\begin{subfigure}{0.45\linewidth}
		\subcaption{RMW}		
		\includegraphics[width=0.9\linewidth]{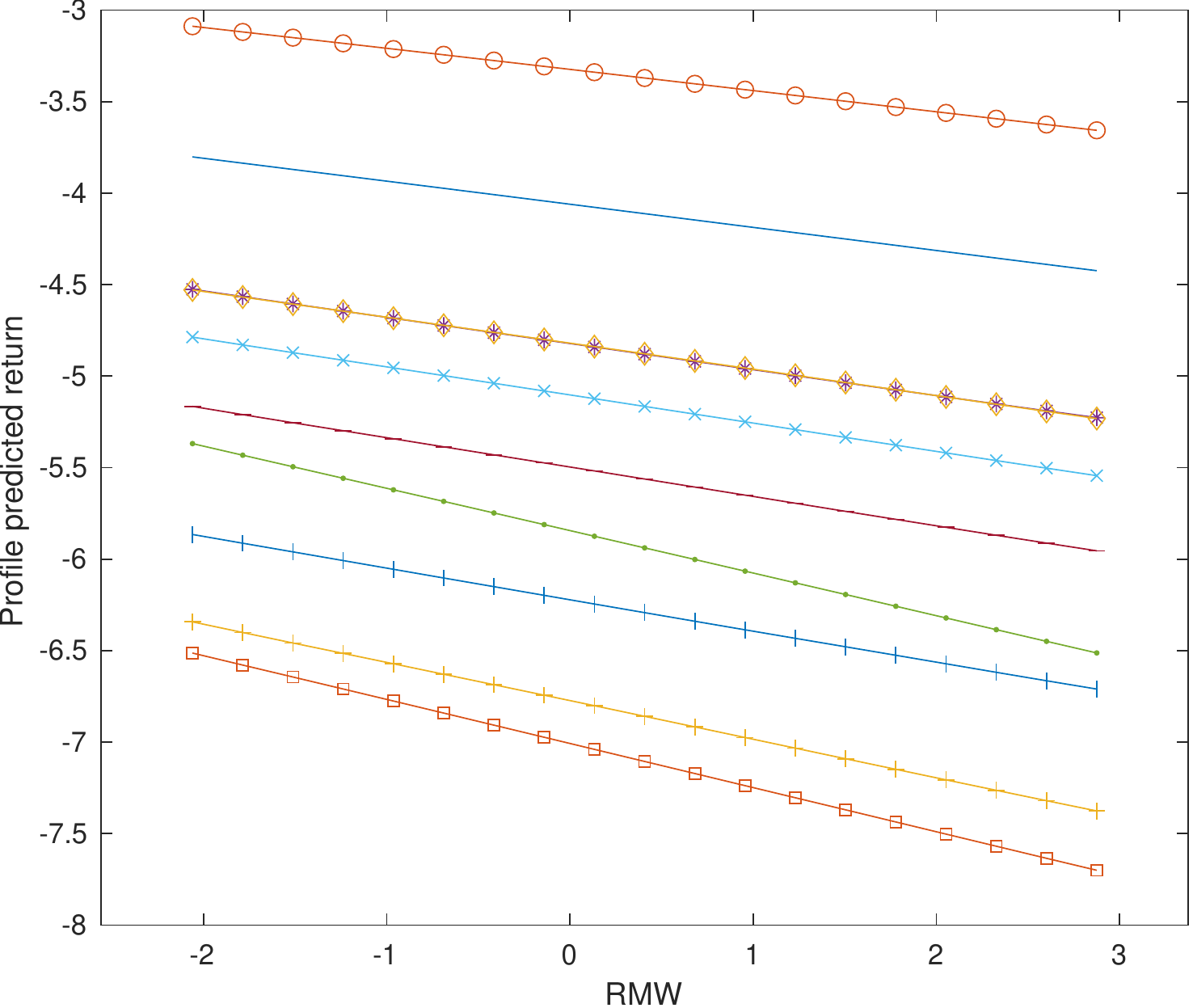}
		\label{ff5_lek2_rmw}		
	\end{subfigure}
	\begin{subfigure}{0.45\linewidth}
		\subcaption{CMA}		
		\includegraphics[width=0.9\linewidth]{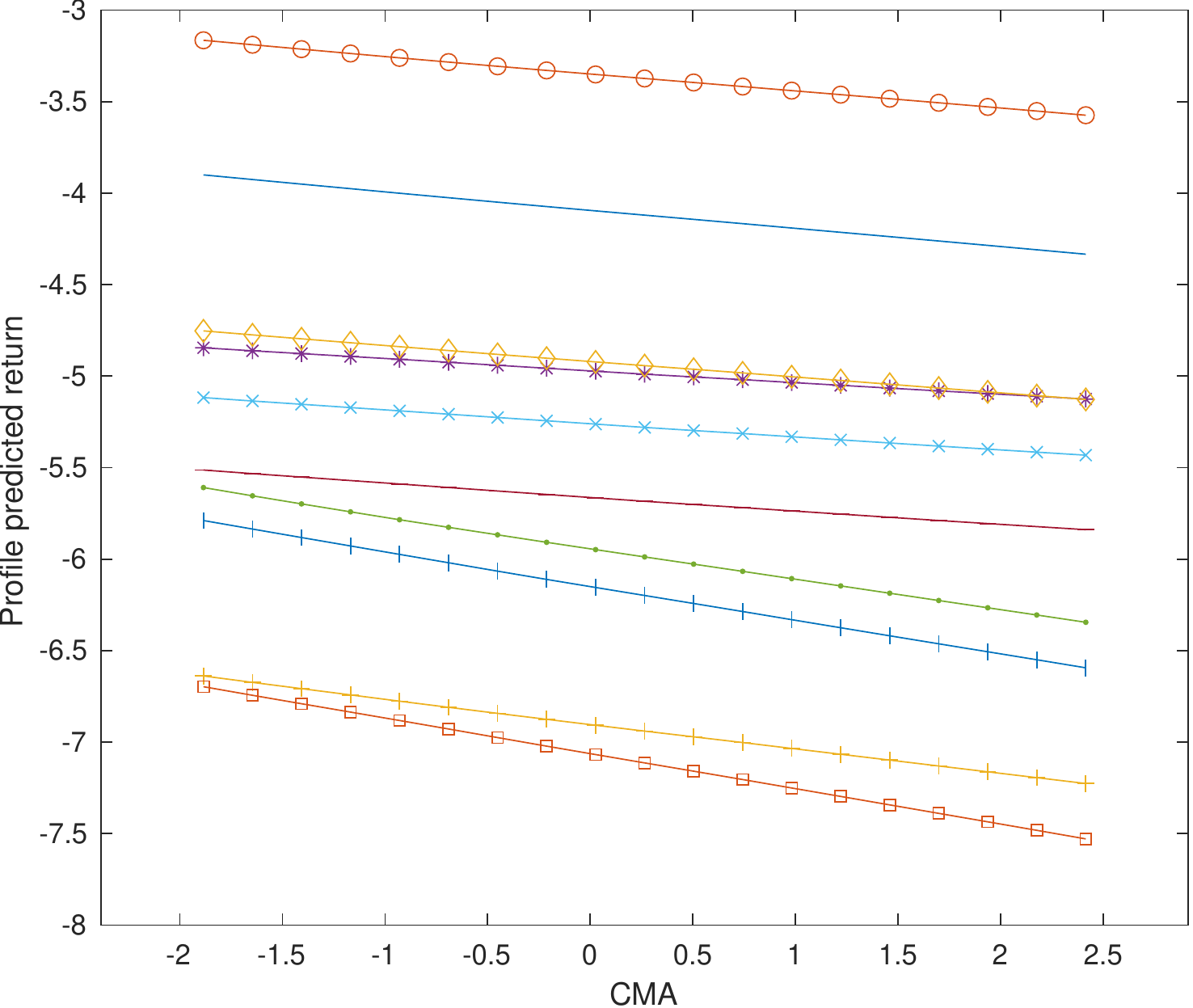}
		\label{ff5_lek2_cma}		
	\end{subfigure}
	\caption{Non-linear transformation and heterogeneity of DeepLMM model (5FF). For May 2012 which has the median level of stock market volatility (as measured by VIX index).}
	\label{ff5_lek2}
\end{figure}
\begin{figure}[ht!]
	\begin{subfigure}{0.45\linewidth}
		\subcaption{ERM}
		\includegraphics[width=0.9\linewidth]{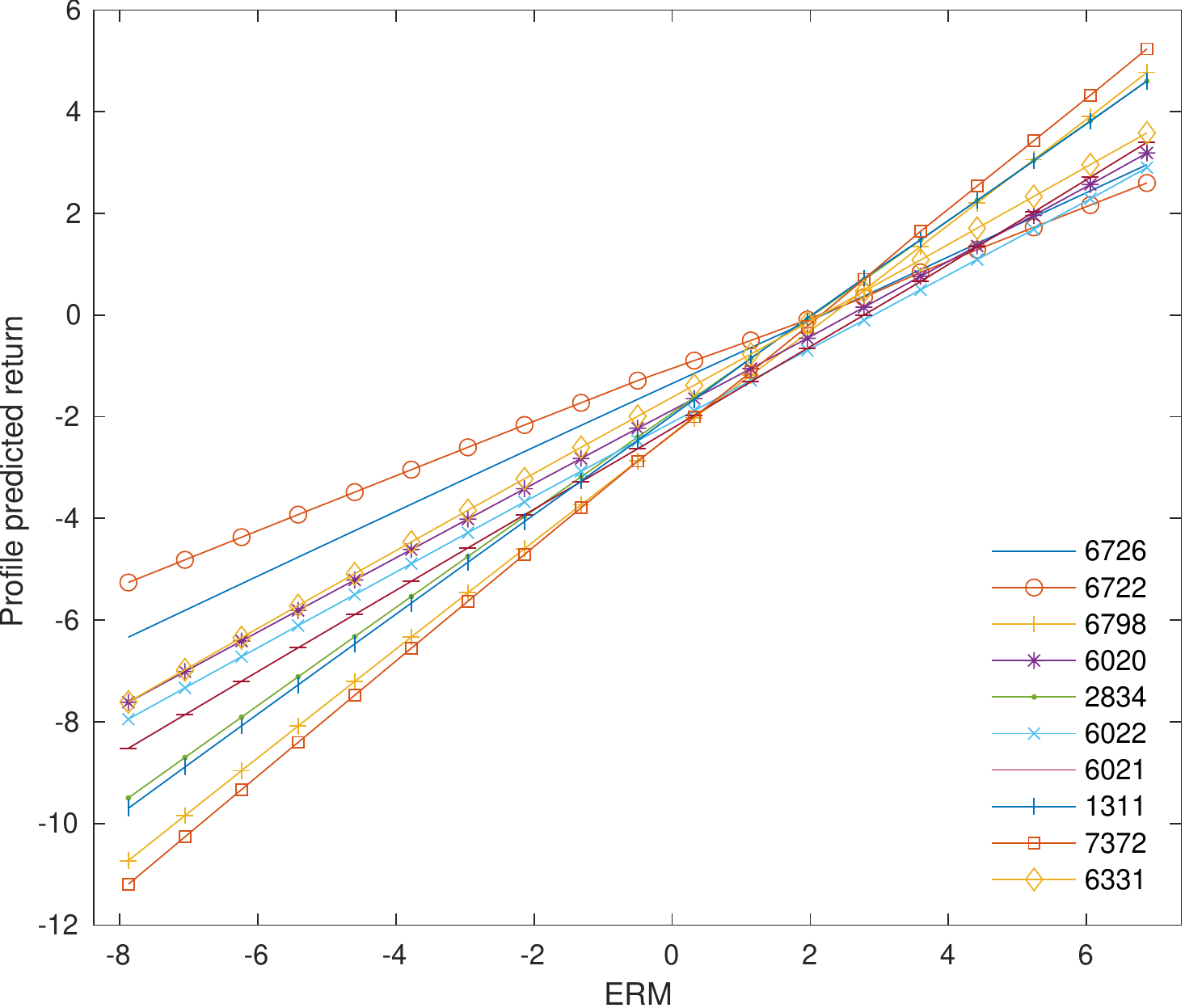}
		\label{ff5_lek3_mktrf}
	\end{subfigure}
	\begin{subfigure}{0.45\linewidth}
		\subcaption{SMB}		
		\includegraphics[width=0.9\linewidth]{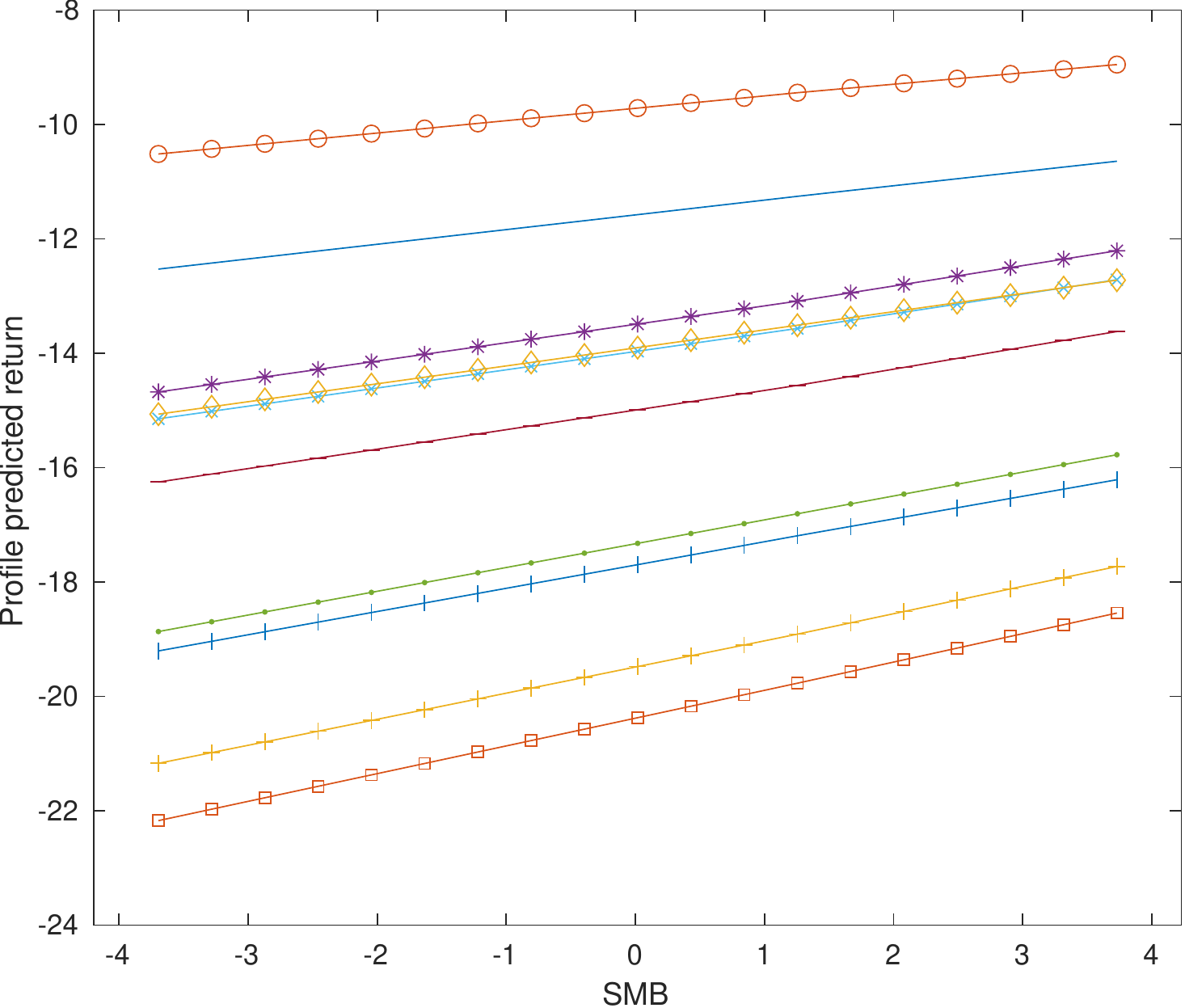}
		\label{ff5_lek3_smb}
	\end{subfigure}
	\begin{subfigure}{0.45\linewidth}
		\subcaption{HML}		
		\includegraphics[width=0.9\linewidth]{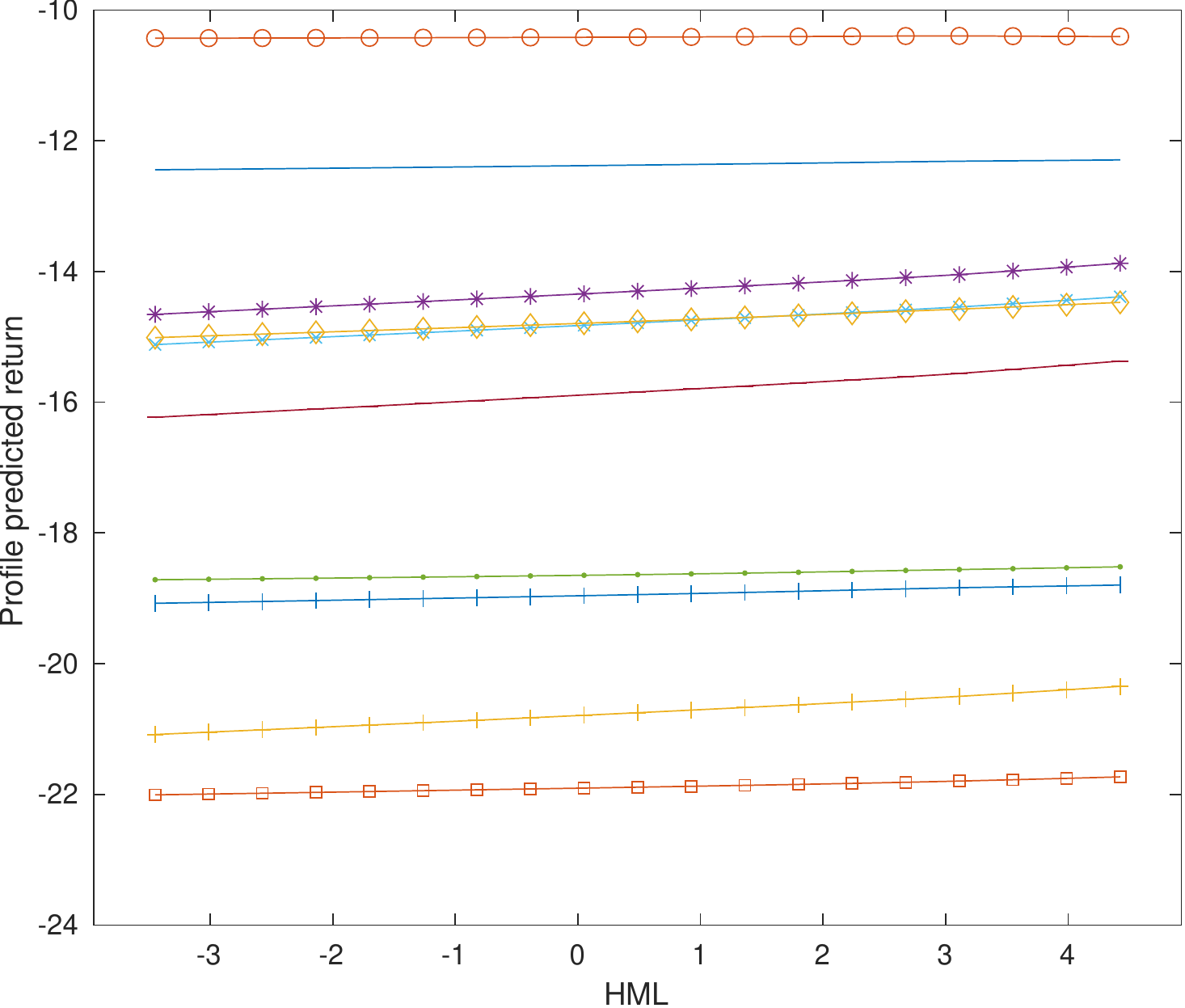}
		\label{ff5_lek3_HML}		
	\end{subfigure}
	\begin{subfigure}{0.45\linewidth}
		\subcaption{RMW}		
		\includegraphics[width=0.9\linewidth]{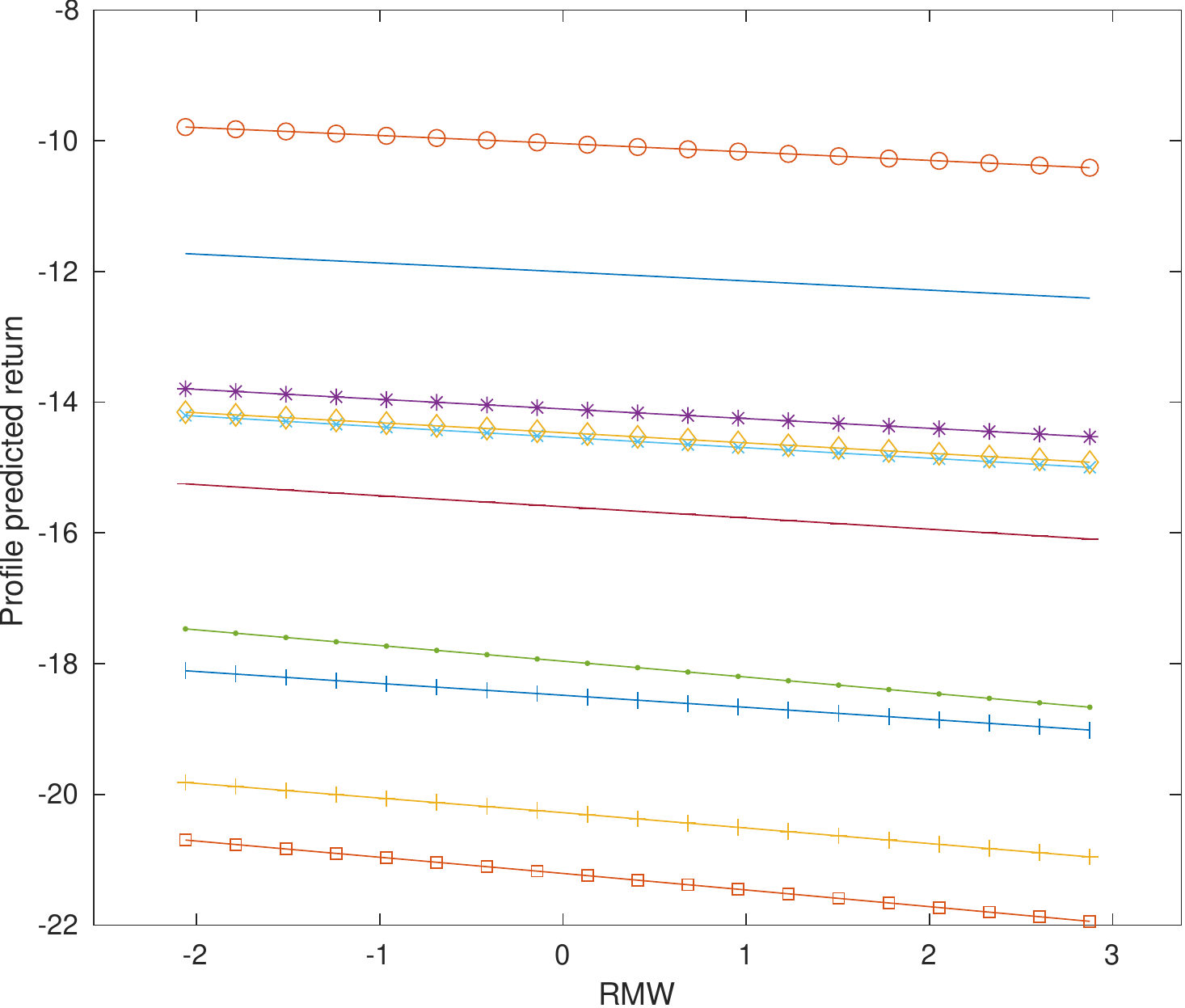}
		\label{ff5_lek3_rmw}		
	\end{subfigure}
	\begin{subfigure}{0.45\linewidth}
		\subcaption{CMA}		
		\includegraphics[width=0.9\linewidth]{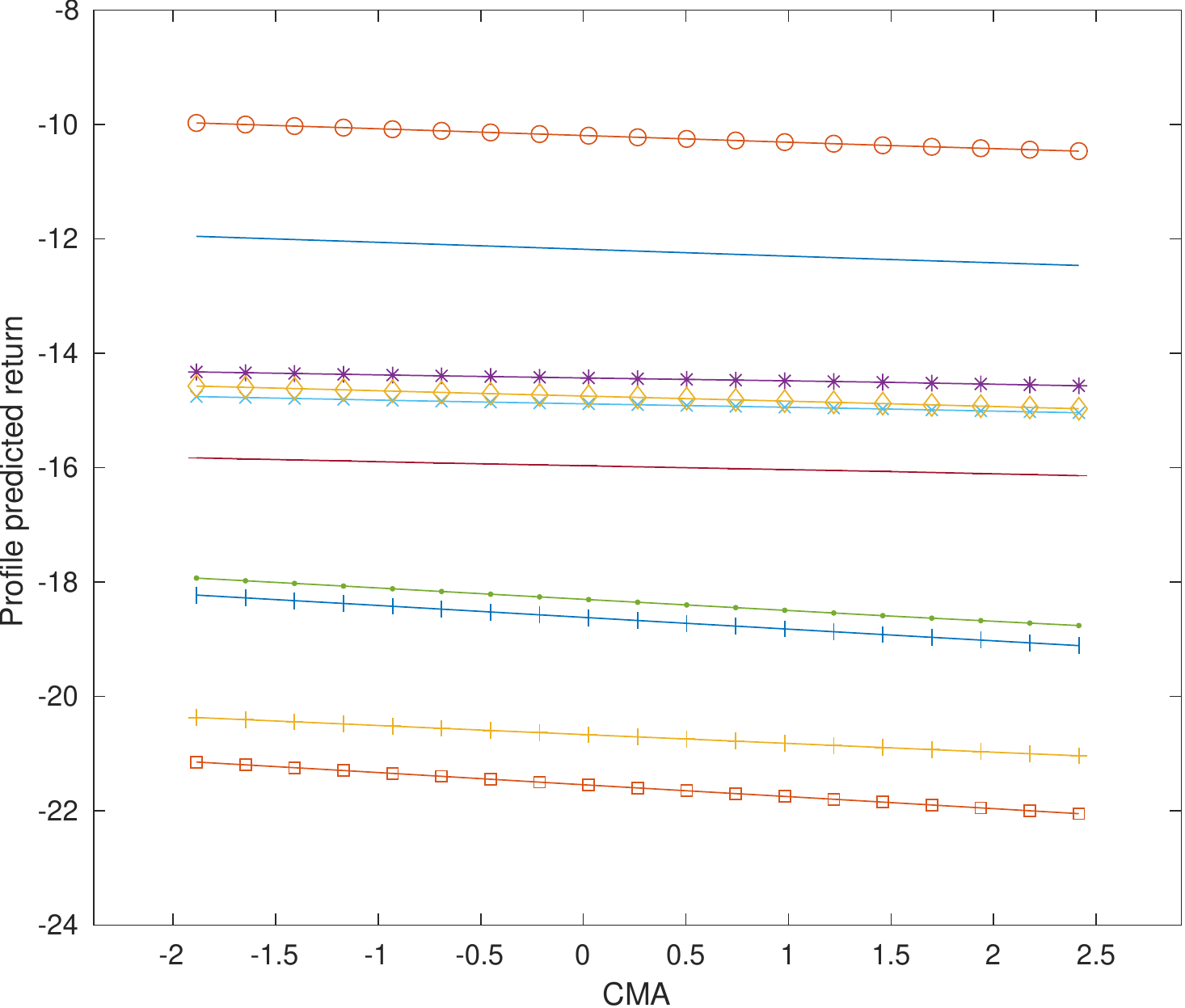}
		\label{ff5_lek3_cma}		
	\end{subfigure}
	\caption{Non-linear transformation and heterogeneity of DeepLMM model (5FF). For October 2008 which has the highest level of stock market volatility (as measured by VIX index).}
	\label{ff5_lek3}
\end{figure}
\end{document}